\pdfoutput=1
\documentclass[11pt]{article}

\usepackage[final]{acl}

\usepackage{times}
\usepackage{latexsym}

\usepackage[T1]{fontenc}
\usepackage[utf8]{inputenc}

\usepackage{microtype}

\usepackage{inconsolata}

\usepackage{graphicx}

\usepackage{kotex}
\usepackage{amsmath,amssymb,amsfonts}
\usepackage{subcaption}
\usepackage{multirow}
\usepackage{graphicx}
\usepackage{pifont}
\usepackage{listings} 
\usepackage{xcolor}
\usepackage{colortbl}
\usepackage{multirow}
\usepackage{booktabs}
\usepackage{xcolor}
\usepackage{newtxtext,newtxmath}
\usepackage{xcolor}
\usepackage{mdframed}
\mdfsetup{%
linecolor=white,
backgroundcolor=gray!20, 
}

\usepackage{authblk} 
\usepackage{hyperref} 

\title{\includegraphics[height=0.9em]{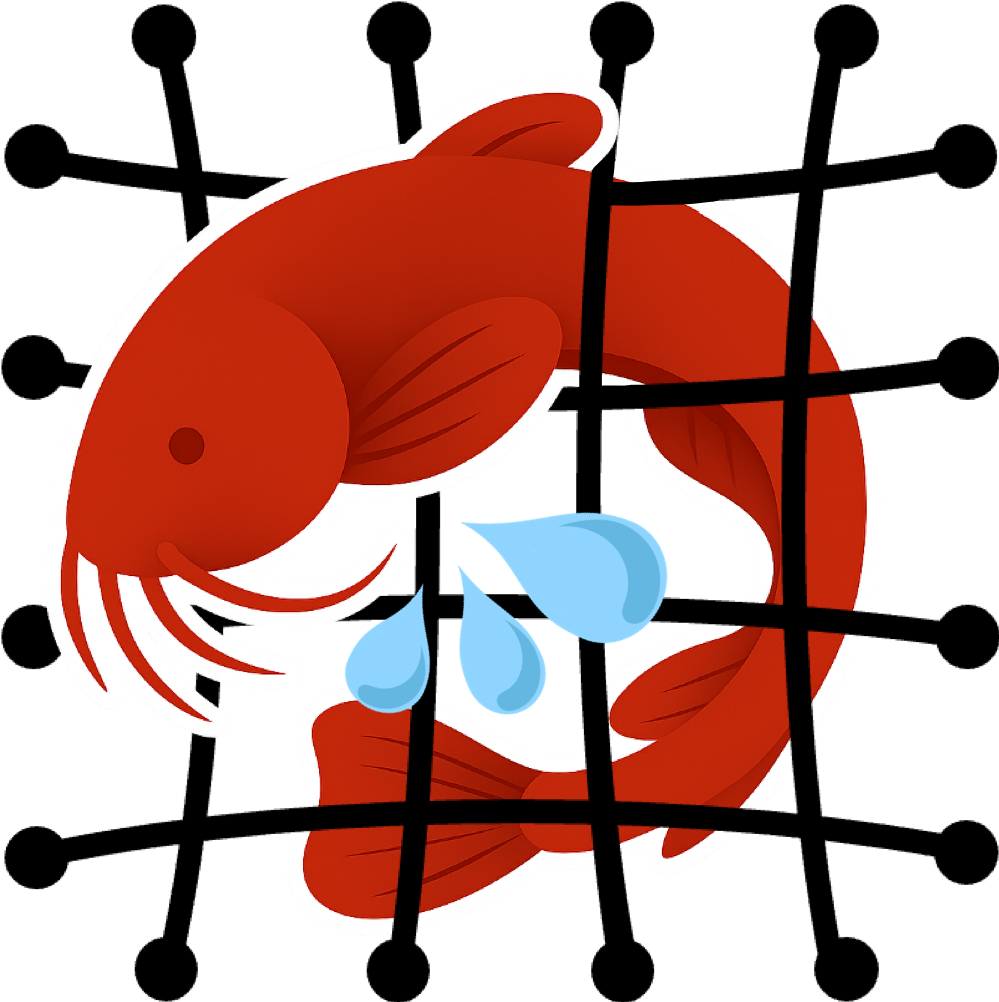}\hspace{0.1em}\texttt{KatFishNet}: Detecting LLM-Generated Korean Text through
Linguistic Feature Analysis}

\author{
\quad Shinwoo Park 
\quad Shubin Kim
\quad Do-Kyung Kim 
\quad Yo-Sub Han$^{\star}$ \\
Yonsei University, Seoul, Republic of Korea \\
\texttt{\small pshkhh@yonsei.ac.kr, shubs919@yonsei.ac.kr, kdky95@yonsei.ac.kr, emmous@yonsei.ac.kr} 
}

\newcommand{\correspondingfootnote}{
    \let\oldthefootnote=\thefootnote
    \renewcommand{\thefootnote}{}
    \footnotetext{$\star$ Corresponding author.}
    \let\thefootnote=\oldthefootnote
}

\begin{document}

\maketitle

\correspondingfootnote

\newcommand{\eg}{{\it e.g.}}%
\newcommand{\ie}{{\it i.e.}}%
\newcommand{\mcal}[1]{{\cal{#1}}}
\newcommand{\calL}{\mbox{$\mathcal{L}$}}
\newcommand{\calB}{\mbox{$\mathcal{B}$}}
\newcommand{\blue}[1]{\textcolor{blue}{\textbf{}#1}} 
\newcommand{\red}[1]{\textcolor{red}{\textbf{}(#1)}}
\newcommand{\green}[1]{\textcolor{green}{\textbf{}#1}}

\begin{abstract}

The rapid advancement of large language models~(LLMs)
increases the difficulty of distinguishing 
between human-written and LLM-generated text. 
Detecting LLM-generated text is crucial for upholding academic integrity, 
preventing plagiarism, protecting copyrights, and ensuring ethical research practices.
Most prior studies on detecting LLM-generated text
focus primarily on English text. 
However, languages with distinct morphological and syntactic characteristics 
require specialized detection approaches. 
Their unique structures and usage patterns hinder 
the direct application of methods primarily designed for English. 
Among such languages, we focus on Korean, 
which has relatively flexible spacing rules, 
a rich morphological system, and less frequent comma usage compared to English.
We introduce \texttt{KatFish}, 
the first benchmark dataset 
for detecting LLM-generated Korean text. 
The dataset consists of text written by humans and 
generated by four LLMs across three genres.
By examining spacing patterns, part-of-speech diversity, and comma usage, 
we illuminate the linguistic differences between human-written and 
LLM-generated Korean text. 
Building on these observations, we propose \texttt{KatFishNet}, 
a detection method specifically designed for the Korean language. 
\texttt{KatFishNet} achieves an average of 19.78\% higher AUC-ROC 
compared to the best-performing existing detection method.
Our code and data are available at \url{https://github.com/Shinwoo-Park/katfishnet}.

\end{abstract}

\section{Introduction} 
\label{sec:introduction}

The rise of LLMs has led to significant advancements in 
various writing tasks~\citep{brown2020language,gomez-rodriguez-williams-2023-confederacy,xiao2024humanaicollaborativeessayscoring}.
However, their ability to generate coherent texts also raises concerns about 
potential misuse, such as spreading misinformation~\citep{pan2023risk,wang2024megafake} and 
facilitating academic dishonesty~\citep{zellers2019defending,perkins2023academic}.
Consequently, detecting LLM-generated text is paramount for safeguarding academic integrity, 
preventing plagiarism, protecting copyrights, 
and upholding research ethics~\citep{guo2023close,su2023hc3,wu2023survey,orenstrakh2024detecting}.

\begin{figure}[t!]
    \centering
    \begin{tabular}{c} 
       \begin{subfigure}[t]{\linewidth}
         \hspace{-5mm}\includegraphics[width=7.5cm]{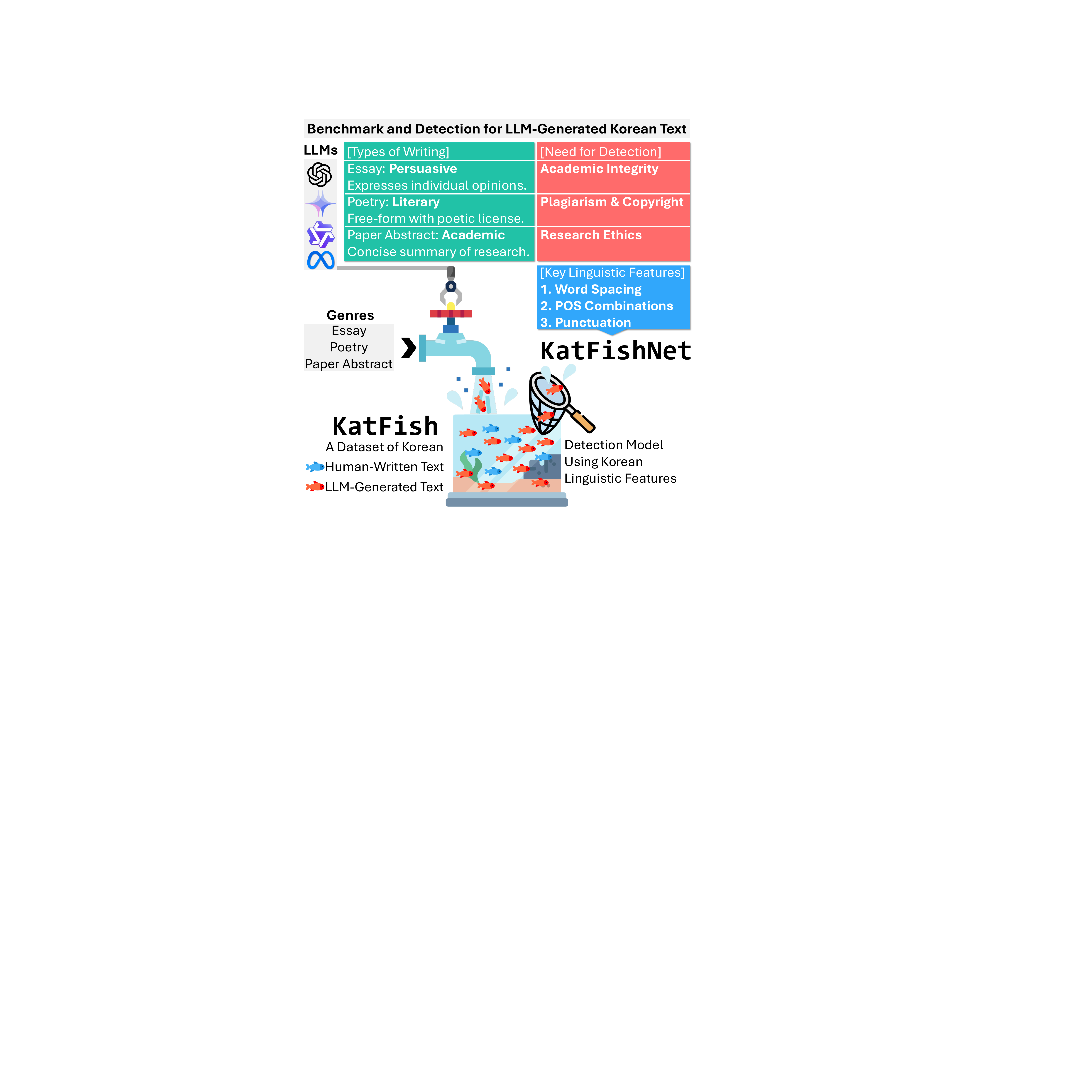}\hfill
            \centering
        \end{subfigure}\qquad
    \end{tabular}
    \caption{An illustration of our \texttt{KatFish} dataset~(Sec.~\ref{sec:dataset_construction}) 
    and the detection method 
    \texttt{KatFishNet}~(Sec.~\ref{sec:linguistic_analysis}).}\label{fig:motivation}
\end{figure}

Despite its significance, research on Korean text has been 
limited.
The limitations in research on detecting LLM-generated
Korean text include the lack of suitable benchmarks
and the challenges arising from the unique linguistic
characteristics of the 
Korean language~\citep{park-etal-2020-empirical,kim-etal-2022-korean,yoon-etal-2023-towards,choi-etal-2024-optimizing,kim2024does}.
We present the 
first dataset for detecting LLM-generated 
Korean text, 
along with a detection method that utilizes 
Korean linguistic features. 
Figure~\ref{fig:motivation} provides an 
illustration of our proposed dataset 
and detection approach.

We present 
\texttt{KatFish}~(\underline{K}ore\underline{a}n LLM-generated \underline{t}ext Benchmark \underline{F}or \underline{i}dentifying Author\underline{sh}ip), 
a dataset developed specifically for detecting Korean text
generated by LLMs.
\texttt{KatFish} includes text samples from
argumentative essays, poetry, and research paper abstracts,
produced by both humans and four LLMs.
Korean has distinct morphological and syntactic features, 
including flexible spacing rules and a rich system of postpositions 
and verb endings. 
These characteristics indicate that detection strategies 
designed for English may be less effective for Korean, 
highlighting the need for a language-specific approach.
We explore the linguistic differences 
between human-written and LLM-generated Korean text 
by analyzing three key aspects: 
1) word spacing; 
2) part-of-speech combinations; and 
3) punctuation. 
Our analysis uncovers distinct feature differences, 
revealing patterns that can be exploited for effective detection.
Based on these findings, we propose \texttt{KatFishNet}, 
a machine learning-based detection method that 
incorporates the linguistic characteristics of 
the Korean language. 

\section{Dataset Construction: 
\texorpdfstring{\includegraphics[height=0.9em]{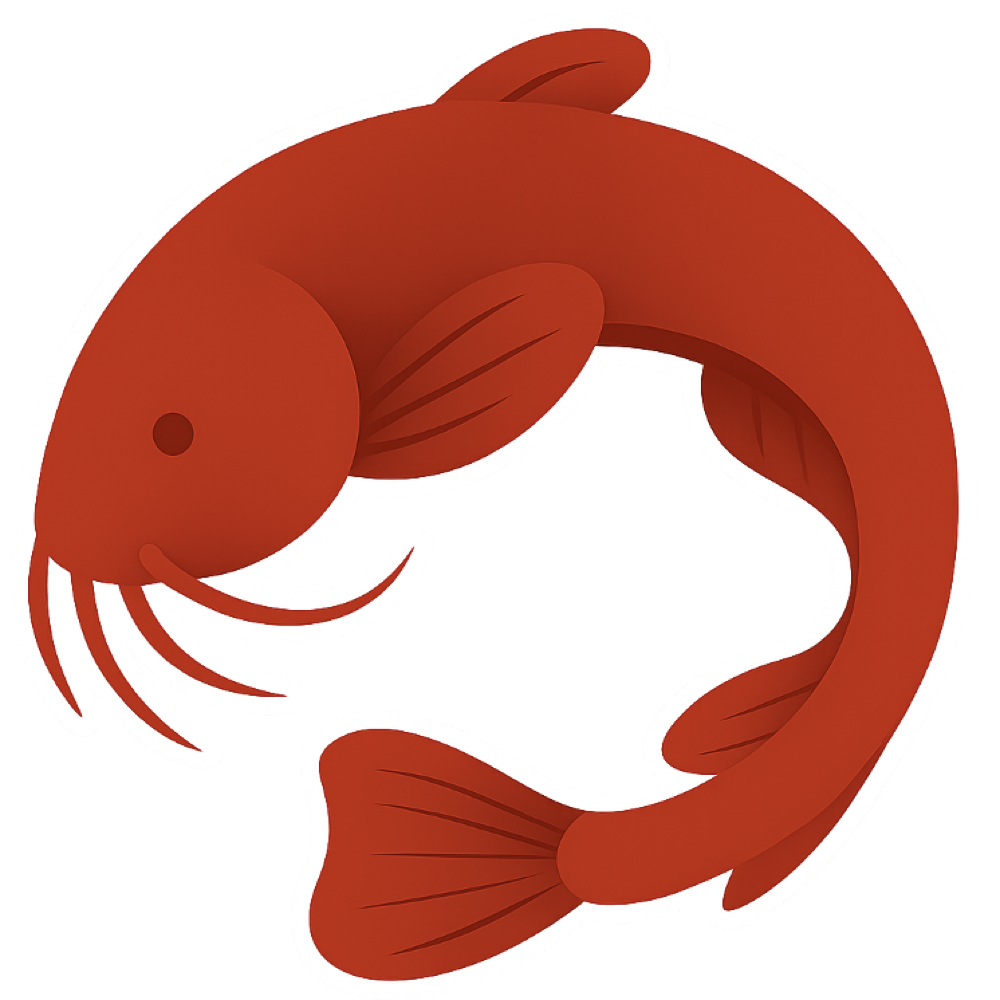}\hspace{0.1em}}{[icon]}\texttt{KatFish}}
\label{sec:dataset_construction}

The \texttt{KatFish} consists of 
three types of text:
\begin{itemize}
    \item \textbf{Essay}: Argumentative essays aim to persuade readers of a specific viewpoint or claim.
    They include a thesis statement, a logically organized structure with 
    supporting evidence and counterarguments, 
    and a concise conclusion. 
    \item \textbf{Poetry}: Poetry is a creative form of writing that 
    focuses on expressing emotions and artistic ideas. 
    It often features metaphor, symbolism, and rhythm, 
    breaking traditional linguistic norms to achieve distinctive artistic effects.
    \item \textbf{Paper Abstract}: Paper abstracts are concise 
    summaries of academic research. 
    They use precise language and technical terminology 
    to clearly communicate the purpose, methodology, 
    and key findings of a study.
\end{itemize}
We select these three genres to 
build a dataset that captures a diverse range of real-life scenarios and linguistic 
features while addressing practical challenges in
detecting LLM-generated Korean text. 
Each genre 
highlights the importance of accurate detection 
in real-world applications.
For a detailed rationale behind our genre selection, see Appendix~\ref{sec:genre_rationale}.

\paragraph{Linguistic and Structural Diversity}
Essays typically follow a logical and coherent structure, 
incorporating argumentation and supporting evidence. 
Poetry stands out with its use of metaphor, rhythm, and 
stylistic innovation, often pushing the boundaries of 
traditional linguistic norms.
In contrast, paper abstracts are compact and information-dense, 
marked by the frequent use of technical terms and discipline-specific language.

\paragraph{Practical Importance of Detection}
Detecting LLM-generated text in these genres is critical due to potential misuse.
Essays generated by LLMs could facilitate academic dishonesty, 
eroding the value of original thinking. 
LLM-generated poetry raises concerns about plagiarism and the loss 
of authenticity in creative expression. 
Similarly, LLM-generated paper abstracts could threaten the 
integrity of academic research by introducing inaccuracies.

\subsection{Human-Written Text Collection}
We collect human-written Korean text from different sources 
depending on the type of writing. 
\paragraph{Essay}
We collect writings from the essay corpus 
provided by AIHub\footnote{\url{https://aihub.or.kr/}}, 
which includes argumentative essays written by elementary, middle, and 
high school students. 
The collected essays cover a total of 11 topics: 
4 topics for elementary school students, 
4 topics for middle school students, 
and 3 topics for high school students.
There are no overlapping topics among the essays 
written by elementary, middle, and high school students. 
The statistics of the essays written by humans included in our final dataset are as follows:
1) Elementary school: 69 essays; 
2) Middle school: 78 essays; and
3) High school: 34 essays.
Descriptions of each essay topic are provided in Table~\ref{tab:essay_topic}.

\paragraph{Poetry}
We collect free verse poems from the poetry corpus provided 
by the National Institute of Korean Language\footnote{\url{https://kcorpus.korean.go.kr/index/goMain.do}}. 
The collected poems are written by individuals under 10, those aged 10 to 19, 
those in their 20s and 30s, and those aged 40 and above. 
The statistics of the poems are as follows:
\begin{itemize}
    \item Individuals under 10: 19 poems
    \item Individuals aged 10 to 19: 116 poems
    \item Individuals in their 20s and 30s: 44 poems
    \item Individuals aged 40 and above: 10 poems
\end{itemize}

\paragraph{Paper Abstract}
We randomly select 100 papers related to language engineering from 
those published by the 
Korean Institute of Information Scientists 
and Engineers\footnote{\url{https://www.kiise.or.kr/academy/main/main.fa/}} 
in 2016–2018.

\subsection{LLM-Generated Text Collection}
We generate LLM-generated Korean text using 
two commercial LLMs and two open-source LLMs.
Specifically, we use the following LLMs:
1) \textbf{GPT-4o}: GPT-4o 
is a commercial LLM capable of understanding and 
processing all forms of input, including text, images, and speech. 
2) \textbf{Solar}: Solar
is a commercial LLM developed by Upstage, a Korean AI startup. 
3) \textbf{Qwen2 72B}: Qwen2 is an open-source LLM developed by 
Alibaba, capable of understanding and processing around 30 languages, including Korean. 
4) \textbf{Llama3.1 70B}: Llama3.1 is an 
open-source LLM developed by Meta, showing outstanding performance across various tasks. 
Table~\ref{tab:prompts} 
shows the prompts used for text generation.

\paragraph{Essay}
When generating essays using LLMs, we design instructions based on education levels, 
essay topics, and prompts. 
The same essay prompts used by human writers serve as inputs for LLMs. 
LLMs receive instructions to write essays following the given topic and prompt while 
maintaining a writing style suitable for the specified education level. 
This approach helps minimize the influence of writing proficiency differences 
across education levels when distinguishing between human-written and LLM-generated essays.

\paragraph{Poetry}
When generating poems with LLMs, we provide the model with a human-written poem 
along with the age group of the poet and instruct it to create a new poem that matches 
the style and content suitable for that age group. 
The model takes the full human-written poem as input and generates a new poem based on it, 
mimicking a realistic scenario where a person may draws inspiration from existing works to 
produce something original. 
Additionally, the model composes poems 
tailored to a given age group,
which helps reduce the impact of age-related differences 
in writing style when distinguishing between human-written and LLM-generated poems.

\paragraph{Paper Abstract}
A paper abstract summarizes the overall content of a study 
and highlights its key contributions. 
Therefore, we have the LLM read the entire paper excluding the abstract 
and generate a new abstract from the remaining content.

\paragraph{Data Cleaning}
We perform a manual analysis of the LLM-generated text
and remove those that fall into the following three categories: 
1) texts that do not follow the instructions and 
simply output the given prompt;
2) texts that repeatedly produce 
meaningless content~(\eg, AI assistant); 
3) text generated in languages other than Korean.

\begin{table}[hbt]
\centering\small 
\begin{tabular}{l|ccc|c}
\hline
\noalign{\hrule height 0.8pt}
& Essay & Poetry & Paper Abstract & Total  
\\ 
\hline
\# Human & 181 & 189 & 100 & 470
\\ 
\# GPT-4o & 181 & 189 & 100 & 470
\\ 
\# Solar & 140 & 189 & 100 & 429
\\ 
\# Qwen2 & 181 & 189 & 17 & 387
\\ 
\# Llama3.1 & 88 & 189 & 61 & 338
\\ 
\hline 
Total &  771 & 945 & 378 & 2,094
\\
\hline
\noalign{\hrule height 0.8pt}
\end{tabular}
\caption{Data statistics of the \texttt{KatFish} dataset.
}\label{tab:data_statistics}
\end{table}

\paragraph{Dataset Statistic}
Table~\ref{tab:data_statistics} presents the 
data statistics of the \texttt{KatFish} dataset. 
The \texttt{KatFish} includes 470 human-written Korean text and 
1,624 LLM-generated Korean text. 
Each text undergoes a careful manual review to ensure 
it does not contain any sensitive personal information.
We demonstrate that \texttt{KatFish} provides a sufficiently large benchmark 
for the task of distinguishing between human-written and LLM-generated Korean text. 
In comparison, recent studies by 
\citet{mitchell2023detectgpt} and \citet{su-etal-2023-detectllm} 
conduct experiments on similar tasks 
involving human-written and LLM-generated English texts 
using 150 to 500 examples. 

Table~\ref{tab:num_eojeol} presents the mean and standard deviation of Eojeol counts 
in texts by each author for each genre. 
An Eojeol is the smallest unit in a Korean sentence, 
separated by spaces.

\section{Detection Method: \texorpdfstring{\includegraphics[height=0.9em]{figure/icon1.png}\hspace{0.1em}}{[icon]}\texttt{KatFishNet}}
\label{sec:linguistic_analysis}

We compare and analyze the linguistic features of 
human-written Korean text and LLM-generated Korean text, 
and design \texttt{KatFishNet} based on these findings.
Specifically, we focus on spacing patterns, part-of-speech 
n-gram diversity, and comma usage patterns. 
These are closely related to writing habits, 
grammatical structures, and textual coherence. 

\begin{figure*}[hbt!]
    \centering

    \begin{subfigure}{0.32\textwidth}
        \centering
        \includegraphics[width=\linewidth]{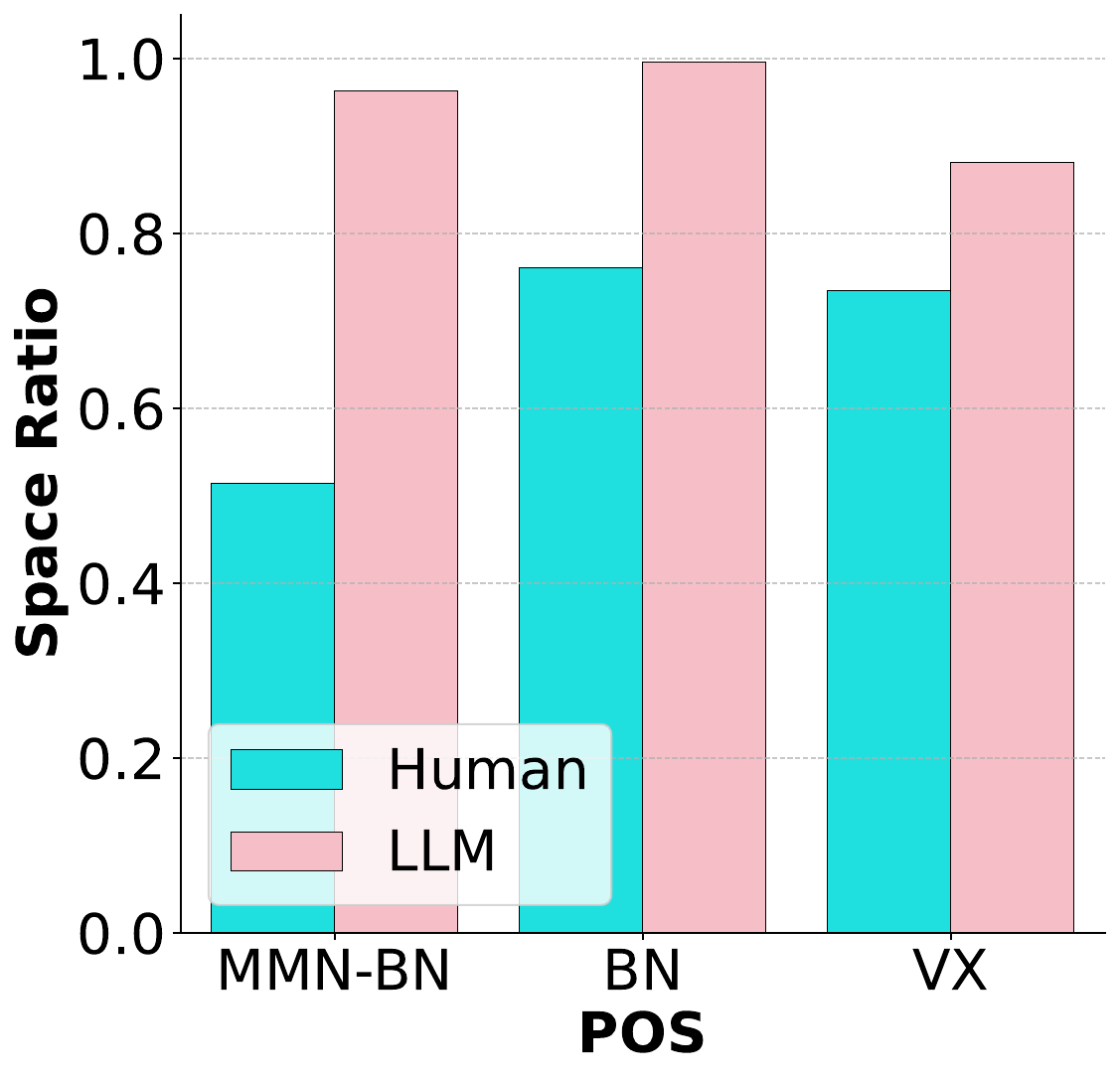}
        \caption{Essay}
    \end{subfigure}
    \begin{subfigure}{0.32\textwidth}
        \centering
        \includegraphics[width=\linewidth]{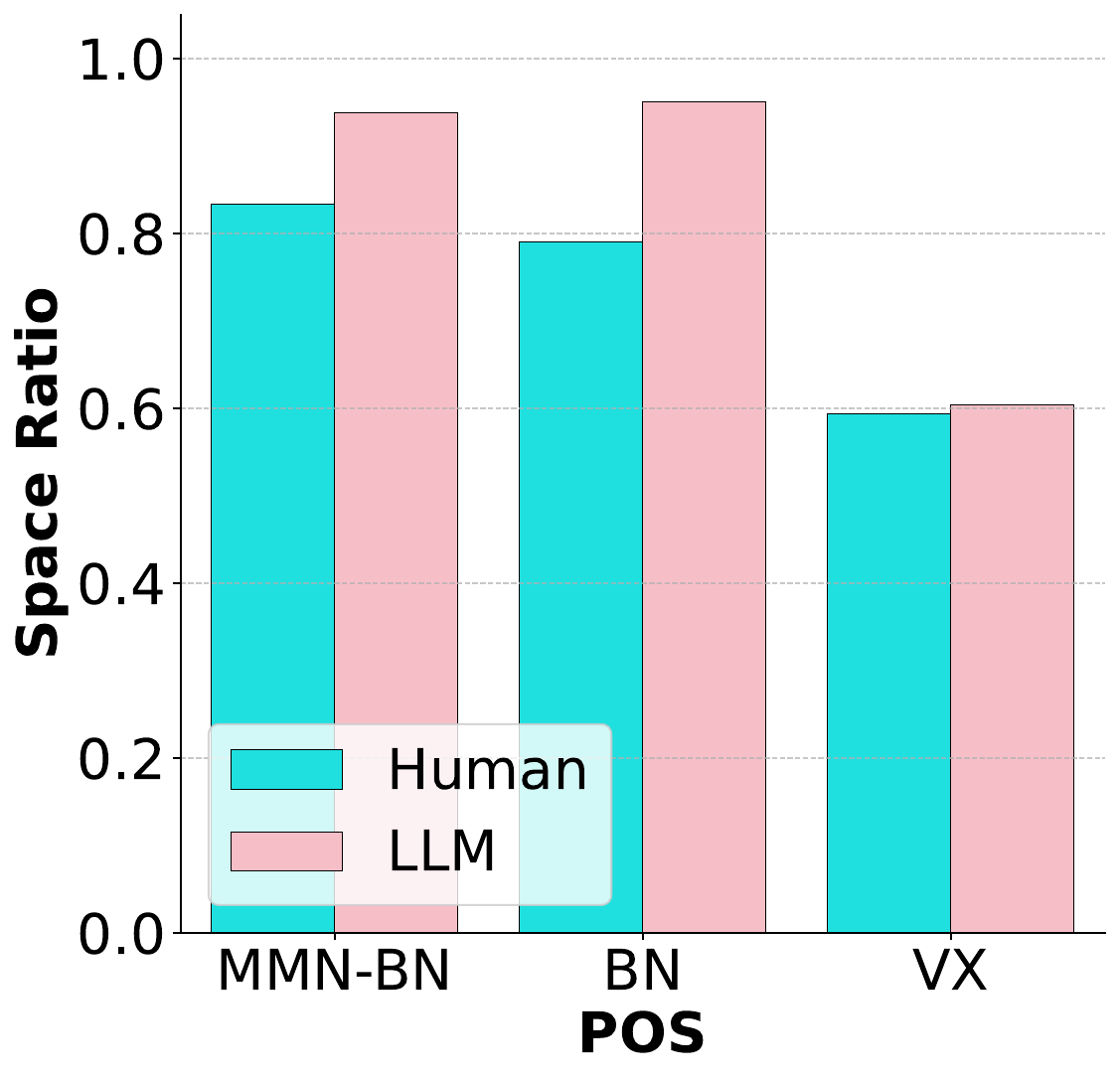}
        \caption{Poetry}
    \end{subfigure}
    \begin{subfigure}{0.32\textwidth}
        \centering
        \includegraphics[width=\linewidth]{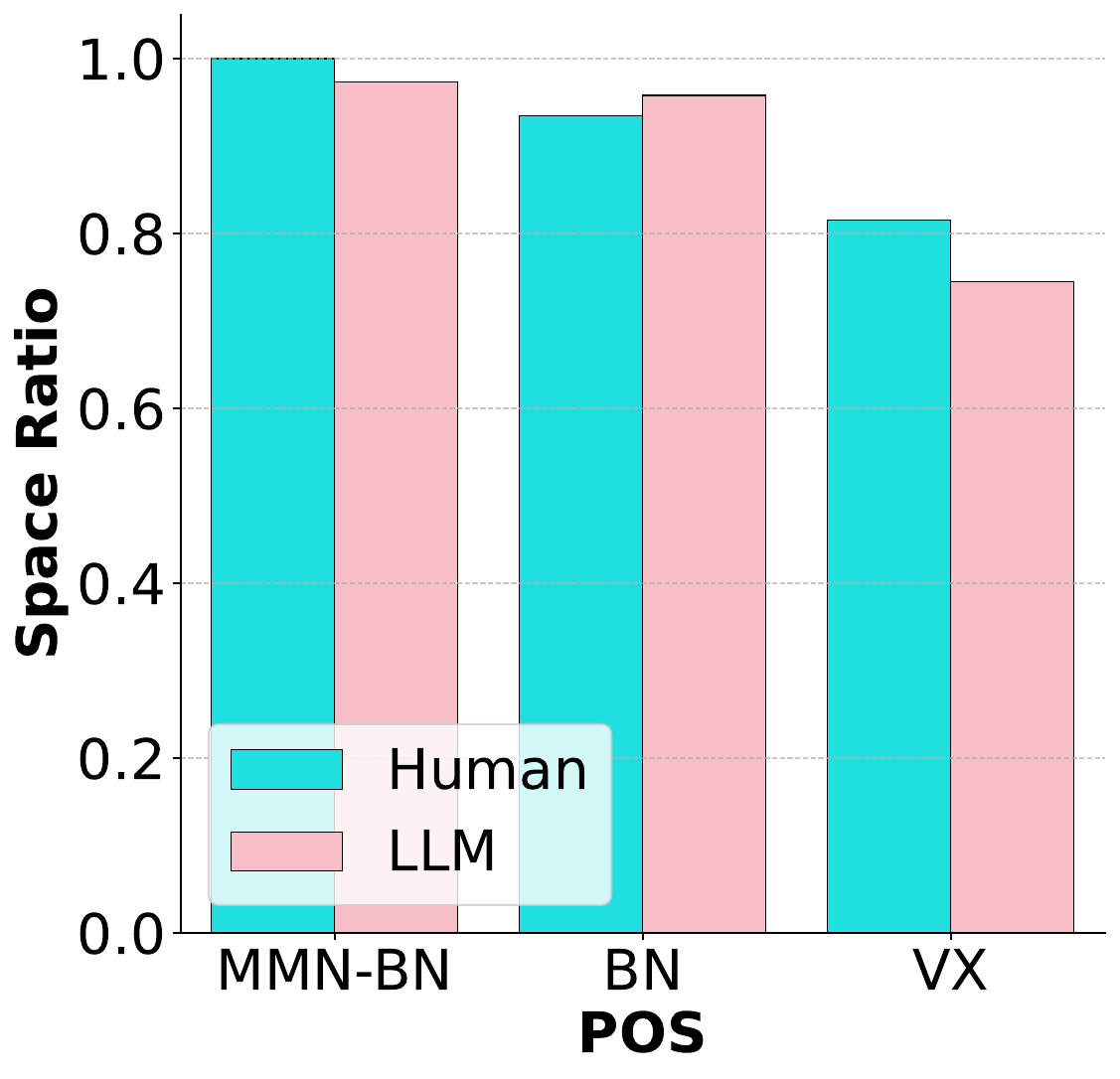}
        \caption{Paper Abstract}
    \end{subfigure}
    
    \caption{Comparison of space occurrence ratios between MMN+BN, prior to BN, and prior to VX.}
    \label{fig:space_analysis}
\end{figure*}

\subsection{Word Spacing Patterns}
\label{sec:word_spacing_patterns}
Unlike English, Korean has many exceptions and flexibilities in its spacing rules, making it one of the most variable and challenging aspects of writing. 
This makes it a valuable feature for examining stylistic and grammatical differences.
In line with the Korean orthography word spacing guidelines\footnote{\url{https://korean.go.kr/kornorms/regltn/regltnView.do?regltn_code=0001&regltn_no=182\#a182}}, 
we examine bound nouns (BN) and auxiliary predicate elements (VX), 
which are the key part-of-speech~(POS) categories related to word spacing. We also investigate \textbf{Eojeol POS Diversity} and \textbf{Unspaced VX Diversity} in Appendix \ref{sec:space_analysis_cont}.
We use Bareun POS tagger.

Bound nouns must be used in conjunction with another word but require word spacing as they function as nouns.
Their dependent nature leads to frequent mistakes, accounting for 70\% of total word spacing errors~\citep{ART001986607}.
We examine two metrics: \textbf{Numeral Determiner–Bound Noun~(MMN-BN) Space Ratio} 
and \textbf{Bound Noun~(BN) Space Ratio}. 
The MMN-BN Space Ratio measures the frequency of word spacing between 
a numeral determiner~(MMN) and a bound noun~(BN).
BN Space Ratio quantifies the frequency of word spacing before a BN. 
To focus on cases with greater variation, 
we eliminate trivial cases where spaces may be omitted.

Auxiliary predicate elements attach to the main predicate to complement its meaning. 
In principle, VXs should have a space preceding them, 
but the guidelines allow flexibility in exceptional cases. 
When used correctly, omitting the space may enhance readability.
\textbf{VX Space Ratio} measures the frequency of word spacing before a VX, 
excluding the specific case of "-아"/"-어" (ENDING) + "지" (VX), 
where spacing is strictly prohibited. This metric indicates how strictly the author adheres to the principle while also considering flexibility in applying exceptions.

Figure~\ref{fig:space_analysis} reveals that in essays and poetry, human-written text exhibits a lower space ratio across all metrics.
Notably, LLM-generated essays display a highly consistent BN Space Ratio, 
with a standard deviation of 0.02. 
While LLMs rigidly enforce spacing rules, humans often omit spaces, 
influenced by various stylistic and grammatical factors. 
These factors include prioritizing readability and convenience over adherence to principles, poetic license, and a lack of understanding of spacing rules.

The differences are the most evident in essays and the least pronounced in paper abstracts. 
This illustrates that spacing behavior is influenced by context-dependent stylistic tendencies. 
In domains where humans adhere to highly structured formats 
and conventions, word spacing patterns may be less significant. 
However, they remain useful in domains with a wider range of authors and writing styles.

\begin{mdframed}
    \noindent\textbf{Finding.}
    LLMs strictly follow spacing rules, while human writers omit spaces due to stylistic and grammatical factors.
\end{mdframed}

\subsection{Part-of-Speech N-gram Diversity}

We analyze POS n-gram diversity 
to examine structural differences between 
human-written and LLM-generated Korean text. 
Using the Kkma POS tagger~\citep{park2014konlpy},
we extract POS sequences from each text and 
compute the \textbf{POS N-gram Diversity Score} 
by dividing the number of unique POS n-grams 
by the total number of POS n-grams in the text. 
After calculating the average diversity score 
for all human-written and LLM-generated text, 
we compare the results to identify differences.
We consider n-grams ranging from unigrams~(1-gram) 
to pentagrams~(5-gram) to capture linguistic patterns 
at different levels. 
Unigrams reflect basic lexical choices, while 
higher-order n-grams capture more complex 
syntactic structures and dependencies. 
By analyzing diversity across these varying n-gram lengths, 
we aim to gain a comprehensive understanding of 
how humans and LLMs construct text differently.

\begin{figure}[h!]
    \centering
    \includegraphics[width=0.8\columnwidth]{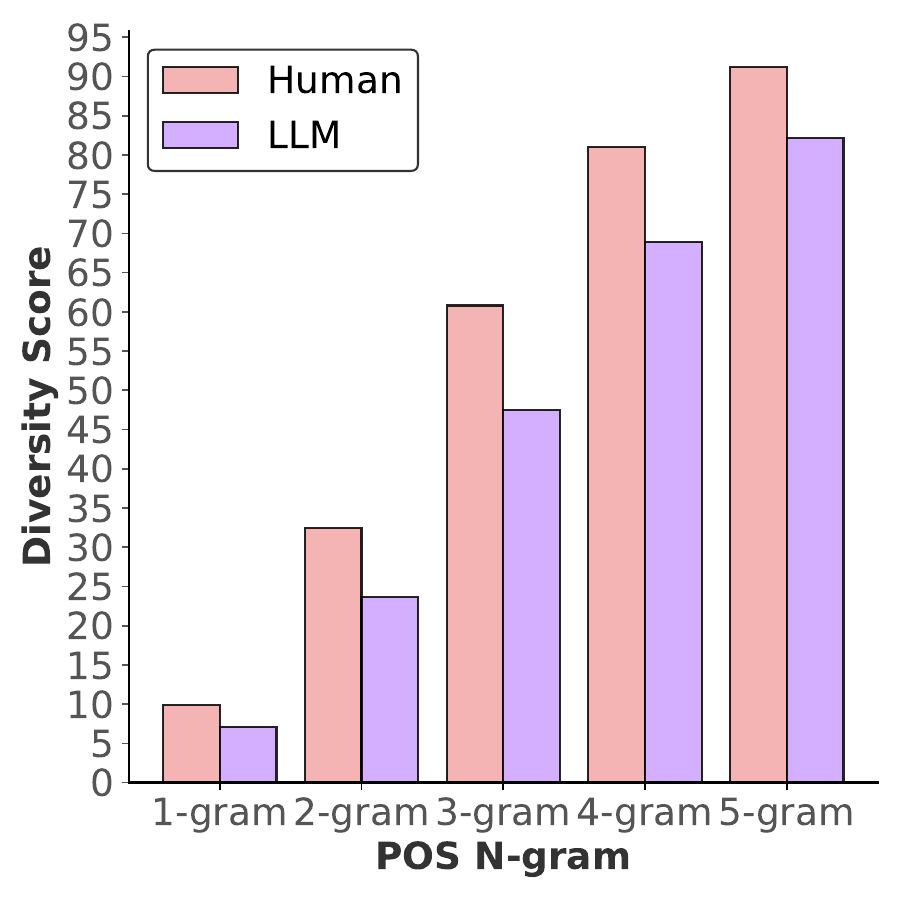}
    \caption{
    Comparison of POS n-gram diversity between 
    human-written and LLM-generated Korean essays. 
    }\label{fig:pos_ngram_diversity} 
\end{figure} 

Figure~\ref{fig:pos_ngram_diversity} 
compares the average POS n-gram diversity scores
between human-written and LLM-generated essays.
We include the analysis results for poetry and 
paper abstracts in the 
Appendix~\ref{sec:appendix_pos_ngram} due to space limitations.
Human-written essays exhibit 
a higher diversity score than LLM-generated essays.
The analysis results highlight that humans use a wider range 
of grammatical structures and construct sentences 
more flexibly than LLMs. 
Since LLMs generate text by selecting 
the most probable word combinations based on training data,
they tend to repeat commonly used structures more 
frequently.
\begin{mdframed}
    \noindent\textbf{Finding.}
    Humans tend to use a more diverse range of POS combinations in their writing compared to LLMs.
\end{mdframed}

\subsection{Comma Usage Patterns}
\label{comma_usage_patterns}
In Korean, commas help improve readability and 
clarify meaning within sentences. 
We analyze differences in comma usage to 
compare how humans and LLMs structure sentences 
and manage their flow. 
Specifically, we investigate 
the proportion of commas within sentences, 
their placement, structural changes in sentences, 
and linguistic diversity around commas. 
We compute the following five metrics:
1) \textbf{Comma Inclusion Rate}: 
The proportion of sentences containing at least one comma 
out of all sentences in a text.
2) \textbf{Average Comma Usage Rate}: 
The number of commas in a sentence divided by the 
total number of morphemes in that sentence.
3) \textbf{Average Relative Position of Comma}: 
The position of each 
comma~(counting the number of morphemes before it) 
divided by the total number of morphemes in the sentence. 
If multiple commas appear in a sentence, 
the average relative position is calculated.
4) \textbf{Average Segment Length}: 
The average length of sentence segments split by commas.
5) \textbf{POS Diversity Score Before and After Comma}: 
The diversity of part-of-speech pairs 
appearing before and after a comma. 
This score is calculated by dividing the number of unique 
POS pairs by the total number of POS pairs.
We segment each text into individual sentences, 
compute these metrics for each sentence, 
and use the average values of these sentence-level metrics 
as the representative values for the entire text. 
Appendix~\ref{sec:pos_comma_analysis} further explores the specific POS patterns surrounding commas.

\begin{table}[hbt]
\centering\small 

\begin{tabular}{ll|cc}

\hline
\noalign{\hrule height 0.8pt}
Genre & Metric & Human & LLM  

\\ 
\hline
\multirow{5}{*}{Essay} 
& Inclusion Rate~(\%) & 26.31 & \textbf{61.03 }
\\ 
& Usage Rate~(\%) & 1.13 & \textbf{2.56 }
\\ 
& Avg. Relative Position & 0.09 & \textbf{0.18} 
\\ 
& Avg. Segment Length & 4.35 & \textbf{8.56 }
\\ 
& POS Diversity Score & 24.38 & \textbf{59.39 }
\\ 
\hline

\multirow{5}{*}{Poetry} 
& Inclusion Rate~(\%) & 27.01 & \textbf{42.90 }
\\ 
& Usage Rate~(\%) & 2.61 & \textbf{4.84 }
\\ 
& Avg. Relative Position & 0.14 & \textbf{0.28 }
\\ 
& Avg. Segment Length & 1.96 & \textbf{2.13 }
\\ 
& POS Diversity Score & 23.13 & \textbf{23.86 }
\\ 
\hline

\multirow{4}{*}{Paper} 
& Inclusion Rate~(\%) & 47.48 & \textbf{65.21 }
\\ 
& Usage Rate~(\%) & 1.73 & \textbf{2.40} 
\\ 
& Avg. Relative Position & 0.20 & \textbf{0.25 }
\\ 
\multirow{1}{*}{Abstract} 
& Avg. Segment Length & 9.07 & \textbf{11.55 }
\\ 
& POS Diversity Score & 42.85 & \textbf{61.95 }
\\ 
\hline
\noalign{\hrule height 0.8pt}

\end{tabular}
\caption{Comparison of comma usage patterns between 
human-written and LLM-generated Korean text. 
}\label{tab:comma_usage}
\end{table}

Table~\ref{tab:comma_usage} presents the analysis results. 
1) LLMs include commas in more sentences and 
uses them more frequently:
LLM-generated text contains a higher proportion of 
sentences with commas compared to human-written text.
The frequency of comma usage within sentences is also 
higher in LLM-generated text, 
leading to more frequent segmentation.
2) LLMs tend to place commas later in a sentence than humans:
While both humans and LLMs often place commas near the beginning of a sentence, 
analysis shows that LLMs tend to insert them slightly later. 
This can be attributed to the fact that LLMs are trained on 
large multilingual datasets, particularly those with substantial 
amounts of English, where comma usage patterns can differ from 
those commonly found in Korean. 
Additionally, in LLM-generated text, the segments of sentences separated 
by commas tend to be relatively longer compared to human-written text.
3) LLM-generated text shows greater diversity 
in part-of-speech combinations around commas:
A higher diversity of POS combinations before and after 
commas indicates that LLMs apply a wider range of 
grammatical patterns when constructing sentences.
Poetry shows almost no difference between human and 
LLM-generated text.
This similarity likely arises because poetry 
naturally consists of shorter and simpler sentence structures.

\begin{mdframed}
    \noindent\textbf{Finding.}
    LLMs use commas more often, place them later, and show greater POS diversity than humans.
\end{mdframed}

\subsection{Design of \texttt{KatFishNet}}
\label{sec:approach}

Our detection method, 
\texttt{KatFishNet}, 
is a machine learning-based model that leverages 
Korean linguistic features.
Table~\ref{tab:linguistic_features} in Appendix~\ref{sec:feature_summary} presents the 
Korean linguistic features used by \texttt{KatFishNet}.
\texttt{KatFishNet} offers several advantages: 
First, it enables analysis of which features play 
a crucial role in detecting LLM-generated text, 
making the model highly interpretable. 
Second, compared to deep learning-based models 
such as Transformer-based approaches, 
it does not require training text embeddings or 
large-scale datasets, 
allowing for a lightweight and efficient detection system.
Lastly, traditional machine learning models like logistic regression, random forests, 
and support vector machines provide a practical and efficient solution 
by enabling training on CPUs, avoiding the costs associated with GPU resources.

We construct input feature vectors based on 
quantitative values obtained from word spacing, 
POS combinations, or punctuation analysis, 
and perform machine learning on these vectors.
In Section~\ref{sec:experimental_results}, 
we investigate which types of quantitative metrics 
are most effective in detecting LLM-generated Korean text 
by comparing the performance of models trained with 
different feature sets. 

\section{Experimental Settings}
\label{sec:experimental_setiings}

\subsection{Task Definition}
We address a binary classification task where, 
given a Korean text, 
the goal is to classify whether the text is generated by 
a human or an LLM.
We evaluate performance using the 
Area Under the Receiver Operating Characteristic curve~(AUC-ROC). 
AUC-ROC measures the area under the curve that plots the 
True Positive Rate against the False Positive Rate at 
different threshold levels. 
This metric measures overall detection performance across all potential 
classification thresholds.

\subsection{Baselines}
We select the following five types of baselines. 
\paragraph{Confidence-based Methods}
Confidence-based methods leverage 
a pre-trained language model such as GPT-2 to 
analyze text and extract distinctive features, 
such as the rank or entropy of each word based on its preceding context. 
We employ the following methods:
1) Log-Likelihood~\citep{solaiman2019release}: A language model calculates the 
log probability of each word in a given text, and the average of these values 
serves as the score. 
Higher average log probabilities indicate a greater likelihood 
that the text was generated by an LLM.
2) Entropy~\citep{gehrmann-etal-2019-gltr}: The entropy of each word 
is measured based on its preceding context, 
and the average entropy across the text is used as the score. 
LLM-generated text typically have lower entropy values.
3) Log-Rank~\citep{mitchell2023detectgpt}: The absolute rank of each word 
is determined based on the preceding context, 
and the text-level score is obtained by averaging the logarithm of these values. 
4) LRR~(Log-Likelihood Log-Rank Ratio)~\citep{su-etal-2023-detectllm}: The 
LRR is computed by dividing log-likelihood values by log-rank values. 
Generally, LLM-generated text tend to have higher LRR values 
than human-written text.

\paragraph{Perturbation-based Methods}
Perturbation-based methods evaluate changes in the log probability 
of a model when making slight modifications to the original text. 
These methods use a pre-trained language model such as T5 to 
generate multiple perturbed versions of the text. 
By calculating log probabilities for both the original and perturbed text, 
they determine whether a text is machine-generated.
We use the following methods: 
1) DetectGPT~\citep{mitchell2023detectgpt}: DetectGPT measures 
how the log probability of a model changes when making small modifications 
to the original text. 
This method is based on the idea that text generated by an LLM 
typically reaches a local optimum of the model log probability function. 
As a result, minor alterations to LLM-generated text tend to 
lower the log probability compared to the original version.
2) NPR~(Normalized Log-Rank Perturbation)~\citep{su-etal-2023-detectllm}: 
NPR examines how the log-rank score responds to slight perturbations. 
When small modifications are applied, 
the log-rank score of LLM-generated text increases more 
than that of human-written text.

\paragraph{LLM Paraphrasing}
This method 
determines whether a given text is human-written or LLM-generated 
by paraphrasing the original text using an LLM and measuring the 
similarity between the paraphrased and original versions. 
The method operates on the intuition that LLM-generated text undergo 
fewer changes because they align more closely with the generation patterns of LLMs~\citep{zhu-etal-2023-beat}. 
In other words, if the similarity between the original and paraphrased text 
is high, the original text is considered to be LLM-generated. 
We measure this similarity using BARTScore~\citep{yuan2021bartscore}.
We use Exaone~3.5~\citep{research2024exaone} 32B, 
an LLM released by the Korean company LG AI Research, 
to perform paraphrasing.

\paragraph{LLM Prompting}
We provide a Korean text to an LLM and ask it to output 1 if the text is 
LLM-generated and 0 if it is human-written. For this baseline, we use Exaone~3.5.

\paragraph{Fine-tuning}
We fine-tune the encoder of a pre-trained language model 
using the \texttt{KatFish} dataset. 
Specifically, we build on a RoBERTa-base model initially trained on the 
HC3 dataset~\citep{guo2023close}, 
which consists of English and Chinese text written by 
both humans and ChatGPT. 
By further training this model with the \texttt{KatFish} dataset, 
we enhance its ability to distinguish 
between human-written and LLM-generated Korean text.

The implementation details of \texttt{KatFishNet} and the baselines 
are provided in the Appendix~\ref{sec:appendix_implementation_details}.

\subsection{OOD Evaluation: Unseen LLMs}
\label{sec:ood_evaluation}
We perform out-of-distribution~(OOD) evaluation 
to assess how well the detection methods generalize.
Specifically, we test whether the model 
can accurately distinguish between human-written 
and LLM-generated Korean text 
even when faced with texts from an unseen LLM. 
This evaluation design is essential given the frequent emergence of LLMs with distinct 
text generation patterns. 
If a detection system relies only on data from familiar LLMs, 
it may struggle to maintain performance when confronted with a previously unseen model. 
By testing on LLMs not encountered during training, 
we can better approximate real-world conditions and gain deeper insights 
into how the detection system adapts without retraining for every new LLM. 

We split the human-written text into an 8:2 ratio, using 80\% 
of it along with text generated by 
GPT-4o—one of the most representative LLMs—to create the training dataset. 
For evaluation, we use text generated by Solar, Qwen2, and Llama3.1. 
Specifically, we construct three separate test sets by combining the text from each of these LLMs 
with the remaining 20\% of human-written text.
The detection methods requiring training include our proposed approach 
along with the fine-tuning baseline. 
These methods undergo training on the training dataset and are evaluated using the test sets. 
Despite the training process, 
the models are always tested on text generated by unseen LLMs. 
This ensures that all detection methods operate in a zero-shot classification setting.

\begin{table*}[hbt!]
\centering\small
\begin{tabular}{l|l|l|ccc|c}
\hline
\noalign{\hrule height 0.8pt}

Genre & \multicolumn{2}{c|}{Detection Methods} & $\rightarrow$ Solar & $\rightarrow$ Qwen2 & $\rightarrow$ Llama3.1 & Average
\\ 
\hline

\multirow{12}{*}{Essay} & \multirow{4}{*}{Confidence} 
& Log-Likelihood & 83.84 & 23.89 & 66.20 & 57.97
\\
& & Entropy & 31.25 & 84.53 & 44.12 & 53.30
\\ 
& & Log-Rank & 78.84 & 20.66 & 61.92 & 53.80
\\ 
& & LRR & 45.08 & 80.56 & 53.15 & 59.59 
\\ 
\cline{2-7}
& \multirow{2}{*}{Perturbation} 
& DetectGPT & 52.78 & 37.45 & 47.18 & 45.80
\\ 
& & NPR & 55.22 & 19.90 & 44.71 & 39.94 
\\
\cline{2-7}
& LLM Paraphrasing & Exaone 3.5 & 92.08 & 79.74 & 72.00 & 81.27
\\
\cline{2-7}
& LLM Prompting & Exaone 3.5 & 50.42 & 49.74 & 50.07 & 50.07
\\ 
\cline{2-7}
& Fine-tuning & RoBERTa & 66.77 & 66.65 & 64.37 & 65.93
\\ 
\cline{2-7}
& \multirow{3}{*}{\texttt{KatFishNet} (Ours)} 
& Word Spacing & 86.00 & 80.63 & 71.91 & 79.51
\\ 
& & POS Combinations & 92.26 & 83.10 & 73.63 & 82.99 
\\ 
& & Punctuation & 97.57 & 94.63 & 92.45 & \textbf{94.88} 
\\
\hline
\noalign{\hrule height 0.8pt}

\multirow{12}{*}{Poetry} & \multirow{4}{*}{Confidence} 
& Log-Likelihood & 77.06 & 47.34 & 59.99 & 61.46
\\
& & Entropy & 30.90 & 68.28 & 47.68 & 48.95
\\ 
& & Log-Rank & 75.76 & 45.67 & 60.54 & 60.65
\\ 
& & LRR & 34.40 & 55.86 & 39.79 & 43.35 
\\ 
\cline{2-7}
& \multirow{2}{*}{Perturbation} 
& DetectGPT & 67.04 & 64.00 & 67.02 & 66.02
\\ 
& & NPR & 63.75 & 41.21 & 62.92 & 55.96 
\\
\cline{2-7}
& LLM Paraphrasing & Exaone 3.5 & 71.32 & 58.79 & 61.51 & 63.87
\\
\cline{2-7}
& LLM Prompting & Exaone 3.5 & 50.53 & 50.16 & 49.42 & 50.03
\\ 
\cline{2-7}
& Fine-tuning & RoBERTa & 60.35 & 69.61 & 55.96 & 61.97
\\ 
\cline{2-7}
& \multirow{3}{*}{\texttt{KatFishNet} (Ours)} 
& Word Spacing & 71.85 & 65.56 & 43.81 & 60.40
\\ 
& & POS Combinations & 39.41 & 79.17 & 53.32 & 57.30 
\\ 
& & Punctuation & 62.65 & 93.45 & 63.22 & \textbf{73.10}
\\
\hline
\noalign{\hrule height 0.8pt}

\multirow{12}{*}{Paper Abstract} & \multirow{4}{*}{Confidence} 
& Log-Likelihood & 58.52 & 42.41 & 47.86 & 49.59
\\
& & Entropy & 36.13 & 72.64 & 51.85 & 53.54
\\ 
& & Log-Rank & 57.08 & 45.05 & 47.57 & 49.90
\\ 
& & LRR & 49.39 & 47.82 & 54.80 & 50.67 
\\ 
\cline{2-7}
& \multirow{2}{*}{Perturbation} 
& DetectGPT & 55.81 & 51.70 & 51.11 & 52.87
\\ 
& & NPR & 63.14 & 46.76 & 60.98 & 56.96 
\\
\cline{2-7}
& LLM Paraphrasing & Exaone 3.5 & 70.80 & 36.47 & 64.72 & 57.33
\\
\cline{2-7}
& LLM Prompting & Exaone 3.5 & 48.60 & 46.41 & 47.18 & 47.39
\\ 
\cline{2-7}
& Fine-tuning & RoBERTa & 50.70 & 49.73 & 50.02 & 50.15
\\ 
\cline{2-7}
& \multirow{3}{*}{\texttt{KatFishNet} (Ours)} 
& Word Spacing & 57.73 & 66.91 & 49.36 & 58.00
\\ 
& & POS Combinations & 47.47 & 70.05 & 42.47 & 53.33 
\\ 
& & Punctuation & 78.99 & 77.47 & 70.41 & \textbf{75.62} 

\\
\hline
\noalign{\hrule height 0.8pt}

\end{tabular}
\caption{
Performance of detecting LLM-generated Korean text. 
We report the average performance~(AUC-ROC) over five experiments.
We separately report the performance of the detection model for the 
task of distinguishing between human-written text and text 
generated by a specific LLM. 
For example, $\rightarrow$ Solar indicates that the detection model 
is evaluated on the task of classifying human-written text 
and text generated by Solar.
}\label{tab:experimental_results}
\end{table*}

\section{Experimental Results}
\label{sec:experimental_results}

Table~\ref{tab:experimental_results} presents the experimental results.
We analyze the results from two perspectives:
1) Which baseline method performs best?
2) Which type of linguistic features contributes most to performance?

\paragraph{Best Performing Baseline Method}
The experimental results show that among the baseline methods, 
LLM paraphrasing achieves the highest performance for essays and abstracts, 
while DetectGPT performs best for poetry. 
In terms of average performance across the three text genres, 
LLM paraphrasing outperforms the other baselines. 
This may be because LLM paraphrasing directly exploits the characteristics of LLM-generated text. 
The results provide experimental support for the hypothesis that when an LLM modifies text, 
it introduces fewer changes to LLM-generated text than to human-written text. 

\paragraph{Most Effective Linguistic Features}
We use logistic regression as the backbone model for \texttt{KatFishNet} and 
provide additional experimental results with random forest and support vector machine models 
in Appendix~\ref{sec:appendix_experimental_results}.
The results show that \texttt{KatFishNet} achieves the highest performance when leveraging comma usage patterns, 
compared to spacing patterns and POS n-gram diversity.
\texttt{KatFishNet} with comma usage patterns outperforms all other methods across all three text genres. 
It achieves a 16.74\% performance improvement over LLM paraphrasing for essays, 
a 10.72\% improvement over DetectGPT for poetry, 
and a 31.90\% improvement over LLM paraphrasing for 
paper abstracts.
Meanwhile, \texttt{KatFishNet} with POS n-gram diversity as features ranks second among all methods 
for essay, while \texttt{KatFishNet} with spacing patterns as features ranks second for abstract.
We hypothesize that comma usage patterns are more useful for detecting LLM-generated Korean text 
than spacing patterns or POS n-gram diversity, 
as LLMs tend to have more difficulty learning comma usage than word spacing or POS combinations.
Spacing follows relatively clear patterns in training data, 
and POS sequences can be learned as probabilistic patterns. 
In contrast, comma usage reflects contextual 
and stylistic factors, 
making it highly variable depending on 
the intent of the writer. 

We analyze the detection performance of \texttt{KatFishNet} in Appendix~\ref{sec:ensemble_features} 
by combining three categories of linguistic features using an ensemble approach.

\begin{mdframed}
    \noindent\textbf{Finding.}
    Comma usage patterns serve as a key feature for distinguishing 
    between human-written and LLM-generated Korean text.
\end{mdframed}

\section{Related Work}
\label{sec:related_work}

\citet{zellers2019defending} developed the GROVER dataset, 
which includes human-written and AI-generated news articles to support research on 
detecting machine-generated disinformation. 
Similarly, \citet{fagni2021tweepfake} created TweepFake, 
a dataset of tweets authored by both humans and bots, 
facilitating studies on social media content authenticity.
\citet{guo2023close} introduced the HC3 dataset, 
which contains questions and answers generated by both human experts and ChatGPT 
in English and Chinese. 
\citet{wang-etal-2024-m4} advanced this direction by introducing the M4 dataset, 
a multi-generator, multi-domain, and multilingual corpus.

\citet{wang-etal-2024-m4gt} introduced M4GT-Bench, a multilingual, multi-domain corpus 
covering various LLMs and proposes more nuanced tasks, 
such as pinpointing the specific LLM behind the generated text~(multi-way detection) 
and identifying human–machine boundary points in mixed-content texts. 
Similarly, \citet{macko-etal-2023-multitude} developed MULTITuDE, 
a large-scale multilingual dataset spanning eleven languages, 
to examine the cross-lingual and cross-generator generalization capabilities of detectors, 
comparing multilingual detectors to monolingual detectors. 
By systematically varying the training and evaluation languages, 
MULTITuDE demonstrates that English-focused models often do not seamlessly transfer to 
other languages without explicit multilingual training.

Although M4GT-Bench and MULTITuDE both address multilingual machine-generated text detection, 
neither benchmark focuses on language-specific approaches. 
In M4GT-Bench, the monolingual setup is trained exclusively on English data, 
while the multilingual setup uses a mixture of languages excluding the target language. 
As a result, there is no training that captures the morphological or 
syntactic characteristics of any single language in particular. 
MULTITuDE trains a separate model for each language and applies the same detection method 
across all of them. 
Its primary goal is to explore language transfer—that is, 
to examine how a model trained on certain languages generalizes to others—rather than 
to enhance performance for individual languages.

In contrast, our work highlights Korean-specific features—such as flexible spacing rules, 
part-of-speech n-gram patterns, and punctuation usage—in the design of the detection method. 
By focusing on language-specific characteristics, we demonstrate how linguistically 
tailored strategies can enhance the performance of LLM-generated text detection.

\section{Conclusion}
We address the challenge of distinguishing between human-written and LLM-generated Korean text 
by introducing \texttt{KatFish}, the first benchmark dataset for this task. 
Building on this foundation, we propose \texttt{KatFishNet}, 
a detection method that leverages linguistic features 
of the Korean language, 
including word spacing, POS combinations, and punctuation. 
Experimental results show that \texttt{KatFishNet}, 
particularly its use of comma usage patterns, 
sets a new state-of-the-art in detection performance.
Notably, the strong performance of \texttt{KatFishNet} demonstrates the effectiveness 
of designing language-specific detection methods.
Our research demonstrates the potential of designing detection methods based on linguistic features, 
providing a foundation for developing similar approaches for other languages in the future.

\section*{Limitations}
This study has several limitations that should be acknowledged.
First, the scope of the \texttt{KatFish} benchmark is limited to three specific genres: 
essays, poetry, and paper abstracts. 
While these genres provide a useful foundation, 
they do not encompass all text types where LLM-generated content could present risks, 
such as news articles, social media posts, and legal documents. 
Expanding the dataset to include a more diverse range of text types would enhance 
the generalizability of the findings.

Second, this study focuses on distinguishing between fully human-written and fully LLM-generated text. 
However, real-world scenarios often involve hybrid content, 
where human- and LLM-generated text are interwoven. 
Future research should investigate detection methods capable of effectively handling such mixed cases.

Finally, enhancing the performance of detection methods that rely on 
linguistic features requires advancements in Korean morphological analysis.
Current morphological analyzers still face challenges, 
which can impact the accuracy and reliability of extracted features. 
Further improvements to these tools could improve linguistic feature extraction and 
detection performance.

\section*{Ethical Considerations}
We ensure that the data collection process for \texttt{KatFish} 
respects privacy and intellectual property rights by 
using publicly available texts and generating AI content 
within ethical guidelines.

\section*{Acknowledgments}
This research was supported by the National Research Foundation of Korea~(NRF) grants 
funded by the Korean government~(MSIT)~(RS-2025-02222626 and RS-2025-00562134), 
and by the AI Graduate School Program~(RS-2020-II201361).

% Custom bibliography entries only

\bibliography{custom}

\appendix
\onecolumn

\section{Genre Selection Criteria and Rationale}
\label{sec:genre_rationale}
\paragraph{Practicality-Driven Domain Selection: Targeting Directly Problematic Use Cases}

Our selection of text types is driven by a focus on practical use cases where the use of LLMs 
for text generation is directly problematic. 
This contrasts with other genres where LLM use is either legitimate or not explicitly prohibited, 
but may still lead to indirect issues. 
We prioritize three genres that directly relate to violations of academic integrity, copyright, 
and research ethics—major issues that are of immediate concern in 
real-world applications of LLM-generated text.

\paragraph{Ensuring Broad Coverage in Writing Types}

Texts are classified into different types by their purpose, 
writing style, and usage. 
To ensure the broad applicability of our findings, 
we meticulously choose three genres—student-written argumentative essays, 
free-verse poetry, and research paper abstracts—to represent major facets of writing. 
Table~\ref{tab:genre-classification} shows how each genre in \texttt{KatFish} 
contributes to this coverage.

\begin{table*}[ht!]
\centering
\begin{tabular}{lll}
\hline
\noalign{\hrule height 0.8pt}
Category & Type & Genres \\
\hline
\multirow{3}{*}{Purpose} 
& Expressive Writing   & Poetry \\
& Informative Writing  & Paper Abstract \\
& Persuasive Writing   & Essay \\
\hline
\multirow{4}{*}{Writing Style} 
& Descriptive          & Poetry \\
& Narrative            & Poetry, Essay \\
& Expository           & Paper Abstract \\
& Persuasive           & Essay \\
\hline
\multirow{2}{*}{Usage} 
& Literary Writing     & Poetry \\
& Academic Writing     & Essay, Paper Abstract \\
\hline
\end{tabular}
\caption{Summary of genre characteristics in \texttt{KatFish}, 
illustrating their complementary roles across purpose, writing style, and usage.}
\label{tab:genre-classification}
\end{table*}

We choose free-verse poetry, a less conventional representative for literary text, 
to investigate how less conventional writing affects LLM-generated text detection. 
This allows us to gain insights that cannot be obtained from more structured forms.

\section{Essay Topics for Argumentative Writing Across Education Levels}
\label{sec:appendix_essay_topic}

Table~\ref{tab:essay_topic} 
shows the essay topics used for writing argumentative essays at each education level.

\begin{table*}[h!]
\centering\small
\begin{tabular}{l|p{12cm}}
\hline
\noalign{\hrule height 0.8pt}
Education Levels & Essay Topic 
\\ 
\hline
\multirow{4}{*}{Elementary} 
& 다문화 가족을 대하는 본인의 자세~(Your Attitude Towards Multicultural Families) \\
& 폭력 예방 방법~(Ways to Prevent Violence) \\ 
& 과학의 발전에 대한 본인의 생각~(Your Thoughts on the Advancement of Science) \\ 
& 미디어 발전과 사용방법~(Advancement of Media and its Usage Methods) \\
\hline 
\multirow{4}{*}{Middle} 
& SNS상의 문제에 대한 본인의 생각~(Your Thoughts on the Issues Related to Social Media) \\
& e스포츠에 대한 본인의 생각~(Your Thoughts on Esports) \\ 
& 전통과 악습에 대한 본인의 생각~(Your Thoughts on Tradition and Bad Practices) \\ 
& 생물학적으로 다른 남/여에 대한 본인의 생각~(Your Thoughts on Biologically Different Males and Females) \\
\hline 
\multirow{3}{*}{High} 
& 인종차별에 대한 본인의 생각~(Your Thoughts on Racism)\\
& 지적 재산권에 대한 본인의 생각~(Your Thoughts on Intellectual Property) \\ 
& 평가에 대한 본인의 생각~(Your Thoughts on the Evaluation) \\ 
\hline
\noalign{\hrule height 0.8pt}
\end{tabular}
\caption{
Essay topics used for argumentative writing at different education levels.
}\label{tab:essay_topic}
\end{table*}

\newpage

\section{Prompt for Korean Text Generation via LLMs}
\label{sec:appendix_prompt}

\begin{table*}[ht!]
\centering\small
\begin{tabular}{ll}

\hline
\noalign{\hrule height 0.8pt}
Korean & Prompt Template 
\\ 
\hline
\multirow{3}{*}{Essay} 
& 너는 과제로 주장글을 작성해야 하는 [EDUCATION LEVEL] 학생이야.  
\\
& 주어진 주제와 질문에 맞춰 에세이를 작성해줘. 
한국어로만 작성해줘. 에세이만 출력해. 
\\ 
& 주제: [TOPIC] 질문: [ESSAY PROMPT]
\\
\hline 
\multirow{3}{*}{Poetry} 
& 너는 [AGE GROUP] 시인이야. 주어진 시를 읽고 내용을 파악해줘.
\\
& 그 후 너의 스타일로 너의 나이대에 맞는 새로운 시를 작성해줘. 
\\ 
& 한국어로만 작성해줘. 새로운 시만 출력해. 주어진 시: [POEM]
\\
\hline 
\multirow{2}{*}{Paper Abstract} 
& 초록을 제외한 논문을 줄테니, 이 논문의 초록을 작성해줘.
\\
& 한국어로만 작성해줘. [PAPER]
\\ 
\hline
\noalign{\hrule height 0.8pt}

\\

\hline
\noalign{\hrule height 0.8pt}
English & Prompt Template 
\\ 
\hline
\multirow{3}{*}{Essay} 
& You are a [EDUCATION LEVEL] student who has to write an 
argumentative essay as an assignment.  
\\
& Write an essay according to the given topic and question. 
Write only in Korean. Output the essay only. 
\\ 
& Topic: [TOPIC] Question: [ESSAY PROMPT]
\\
\hline 
\multirow{3}{*}{Poetry} 
& You are a poet in your [AGE GROUP]. Read the given poem and understand its content. 
\\
& Then, write a new poem in your style that suits your age group.
\\ 
& Write only in Korean. Output the new poem only. Given poem: [POEM]
\\
\hline 
\multirow{2}{*}{Paper Abstract} 
& I'll give you a paper without the abstract. Write an abstract for this paper. 
\\
& Write only in Korean. [PAPER]
\\ 
\hline
\noalign{\hrule height 0.8pt}

\end{tabular}
\caption{
Prompt templates used for building the \texttt{KatFish} benchmark.
The upper table shows the original Korean prompt templates 
used for data generation, and the lower table displays the 
translated English prompt templates.
}\label{tab:prompts}
\end{table*}

\section{Data Statistics: Eojeol Counts}
\label{sec:appendix_data_statistics}

Table~\ref{tab:num_eojeol} 
presents the mean and standard deviation of the number of Eojeols 
in text written by each author for each text type. 
An Eojeol is the smallest unit in a Korean sentence, 
separated by spaces, and may consist of a single morpheme or a 
combination of multiple morphemes.
\begin{table*}[hbt]
\centering
\begin{tabular}{l|ccc}
\hline
\noalign{\hrule height 0.8pt}
& Essay & Poetry & Paper Abstract  
\\ 
\hline
Human & 152$_{\pm54}$ & 52$_{\pm52}$ & 91$_{\pm27}$
\\ 
GPT-4o & 268$_{\pm38}$ & 54$_{\pm24}$ & 120$_{\pm26}$
\\ 
Solar & 175$_{\pm56}$ & 40$_{\pm19}$ & 94$_{\pm55}$
\\ 
Qwen2 & 179$_{\pm44}$ & 62$_{\pm27}$ & 88$_{\pm38}$
\\ 
Llama3.1 & 211$_{\pm28}$ & 77$_{\pm30}$ & 133$_{\pm33}$
\\ 
\hline
\noalign{\hrule height 0.8pt}
\end{tabular}
\caption{The average and standard deviation of Eojeol counts 
for each text type by author. 
}\label{tab:num_eojeol}
\end{table*}

\clearpage

\newpage
\section{Word Spacing Analysis}
\label{sec:space_analysis_cont}
Alongside the three metrics related to word spacing in Section~\ref{sec:word_spacing_patterns}, 
we also analyze \textbf{Eojeol POS Diversity} and \textbf{Unspaced VX Diversity}. 
These metrics offer stylistic and grammatical distinctions in Korean text generation. 
Eojeol POS Diversity is computed by dividing the number of 
unique POS sequences by the total number of Eojeols in the text, 
where each sequence is defined at the Eojeol level.
Eojeol POS diversity captures how diverse the syntactic structures are within individual Eojeols, reflecting differences in linguistic complexity and variability between human-written and LLM-generated text. 
Unspaced VX Diversity is computed by dividing the number of unique unspaced auxiliary verb stems by the total number of unspaced auxiliary verbs. 
We exclude the case where spacing is explicitly allowed. 
This analysis provides insights into the tendency of humans and LLMs to make consistent or varied spacing choices in VX-related word spacing. 

\begin{figure*}[hbt!]
    \centering
        \includegraphics[width=0.65\textwidth]{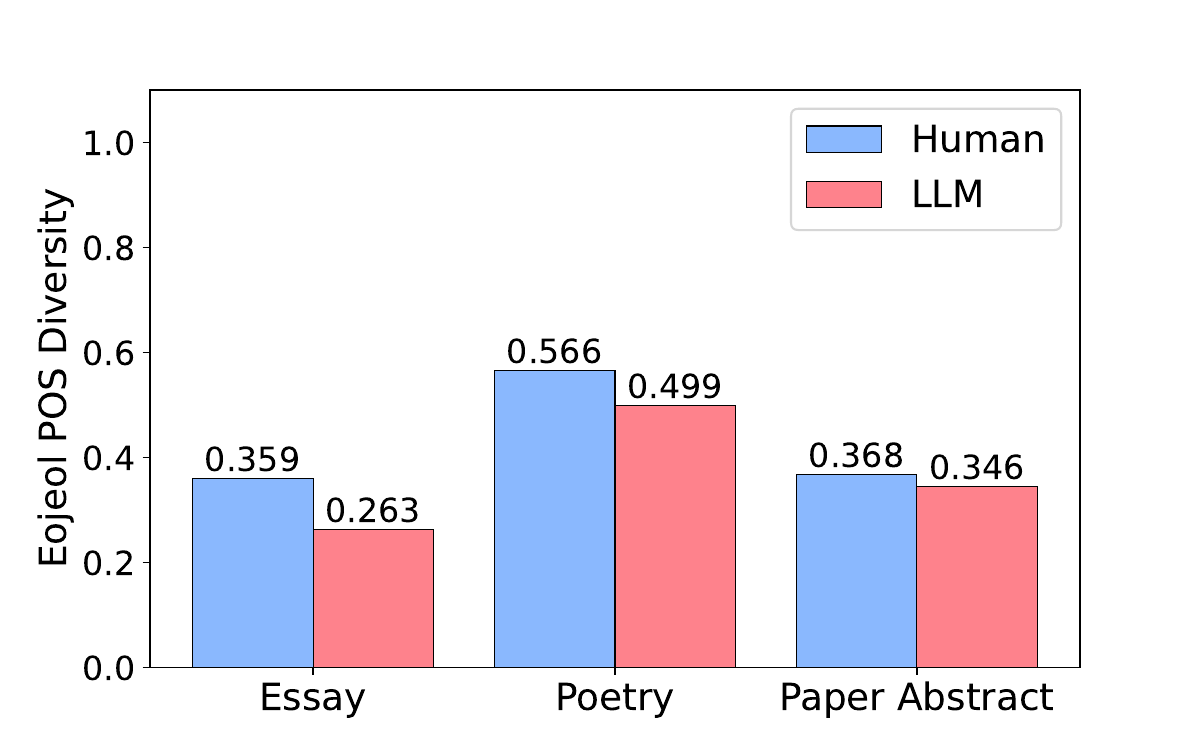}
    \caption{Comparison of Eojeol POS Diversity between human-written and LLM-generated Korean text.
    }\label{fig:eojeol_pos_diversity}
\end{figure*}

\begin{figure*}[hbt!]
    \centering
        \includegraphics[width=0.65\textwidth]{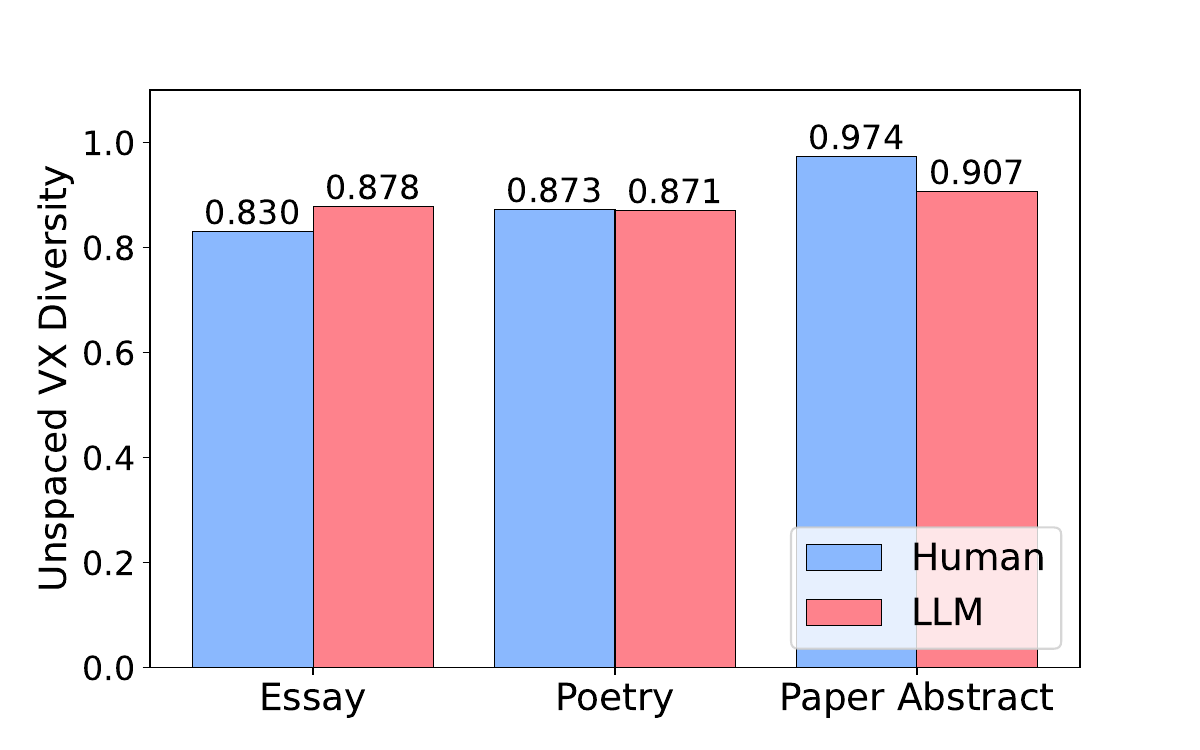}
    \caption{Comparison of Unspaced VX Diversity between human-written and LLM-generated Korean text.
    }\label{fig:unspaced_vx_diversity}
\end{figure*}

Figure~\ref{fig:eojeol_pos_diversity} shows that Eojeol VX Diversity is 
consistently higher for human-written text across all three genres, similar to the POS n-gram diversity results. 
Figure~\ref{fig:unspaced_vx_diversity} illustrates that Unspaced VX Diversity results vary by genre, with humans scoring lower for essays and LLMs scoring lower for paper abstracts. 
This shows that genre-specific stylistic tendencies influence the spacing 
behavior of both humans and LLMs, reflecting variations in writing conventions and levels of formality.

\newpage
\section{Part-of-Speech N-gram Diversity}
\label{sec:appendix_pos_ngram}

\begin{figure}[h]
    \centering

    \begin{subfigure}{0.32\textwidth}
        \centering
        \includegraphics[width=\linewidth]{figure/pos_ngram_essay.pdf}
        \caption{Essay}
    \end{subfigure}
    \begin{subfigure}{0.32\textwidth}
        \centering
        \includegraphics[width=\linewidth]{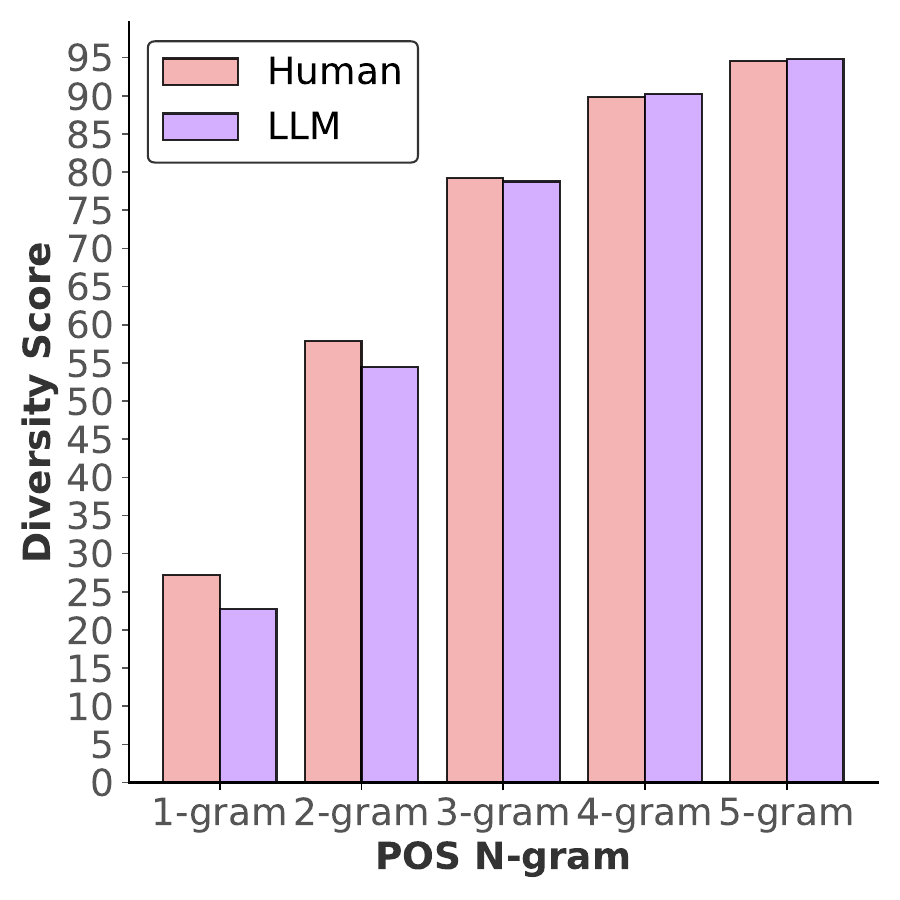}
        \caption{Poetry}
    \end{subfigure}
    \begin{subfigure}{0.32\textwidth}
        \centering
        \includegraphics[width=\linewidth]{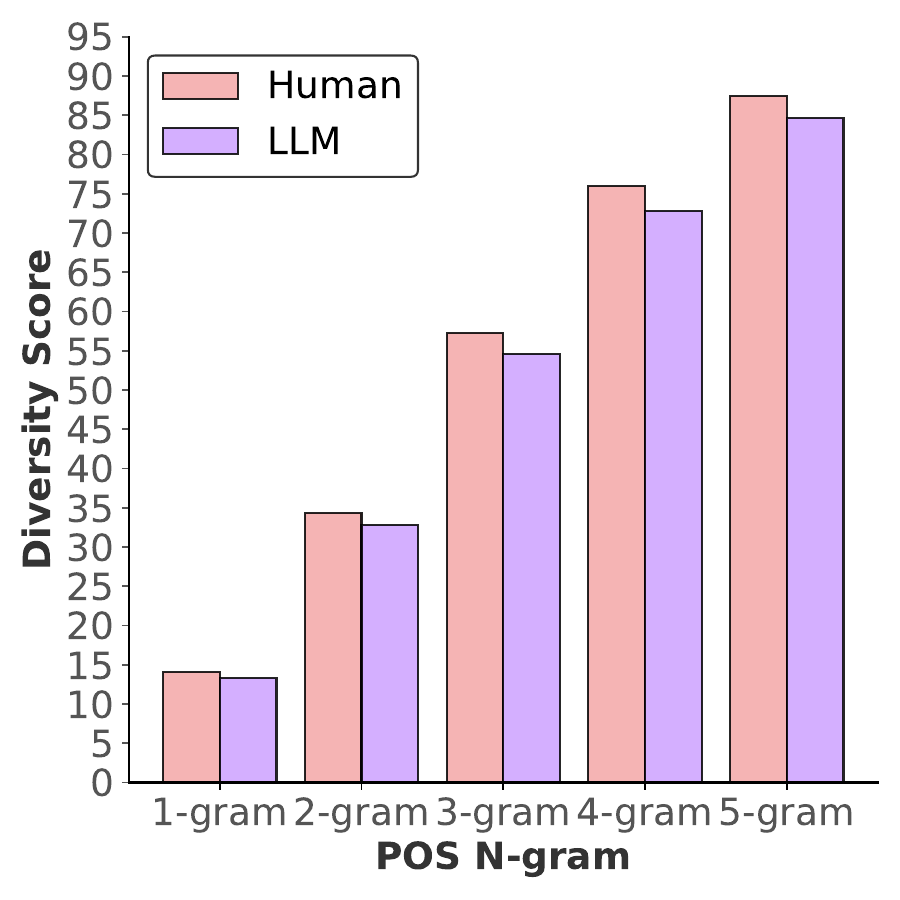}
        \caption{Paper Abstract}
    \end{subfigure}
    
    \caption{Comparison of POS n-gram diversity between human-written and LLM-generated Korean text.}\label{fig:appendix_pos_ngram}
\end{figure}

Figure~\ref{fig:appendix_pos_ngram} 
compares the average POS n-gram diversity scores
between human-written and LLM-generated Korean text.
Excluding POS 4-gram and 5-gram in poetry, 
human-written text shows a higher diversity score than 
LLM-generated text.
Poetry has shorter length and simpler structure compared 
to essays and paper abstracts, 
which can reduce the difference in 
POS n-gram diversity between human-written and 
LLM-generated texts for 4-gram and 5-gram sequences.

\section{POS-Comma Analysis}
\label{sec:pos_comma_analysis}
Exploring when humans and LLMs use commas can give valuable information on stylistic differences between human-written and LLM-generated text. Section~\ref{comma_usage_patterns} extensively discusses how humans and LLMs differ in their quantitative use of commas. In this section, we investigate the differences at the lexical level.

\subsection{Before Comma}
\begin{figure*}[hbt!]
    \centering
    \begin{minipage}{\textwidth}
        \centering
        \includegraphics[width=0.85\textwidth]{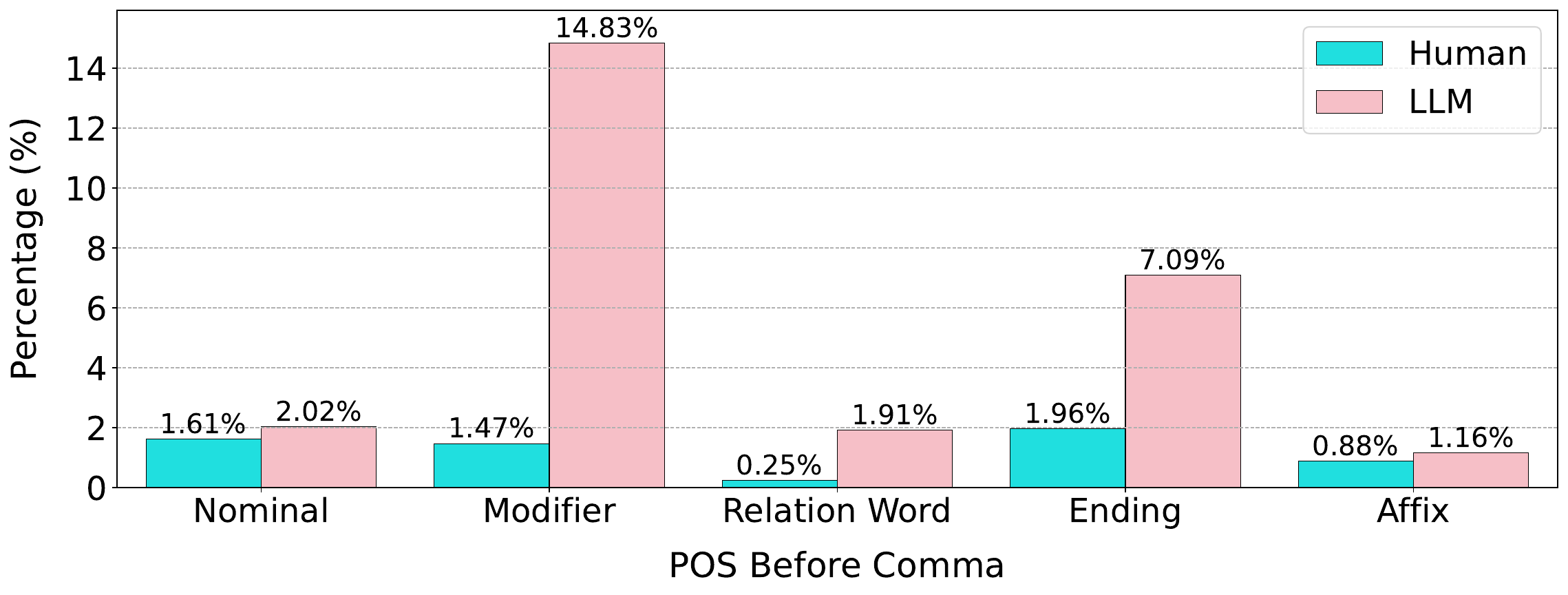}
        \subcaption{Essay}
    \end{minipage}
    \begin{minipage}{\textwidth}
        \centering
        \includegraphics[width=0.85\textwidth]{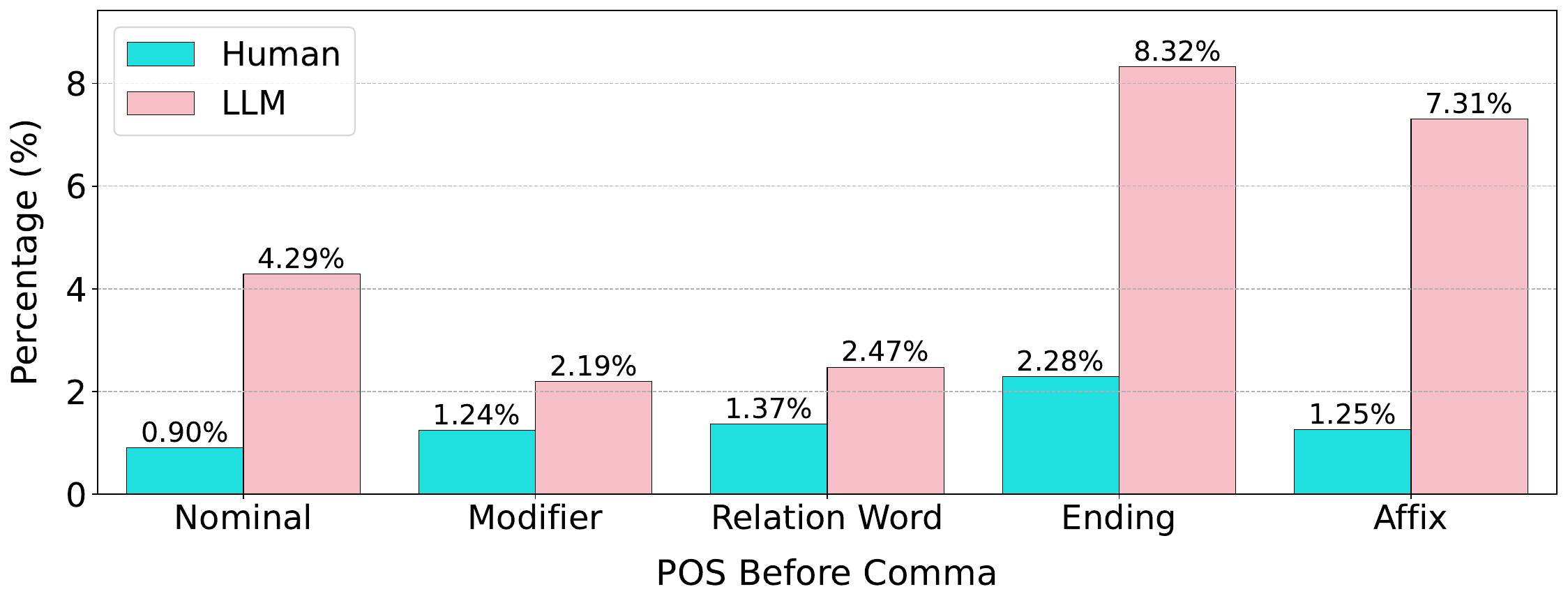}
        \subcaption{Poetry}
    \end{minipage}
    \begin{minipage}{\textwidth}
        \centering
        \includegraphics[width=0.85\textwidth]{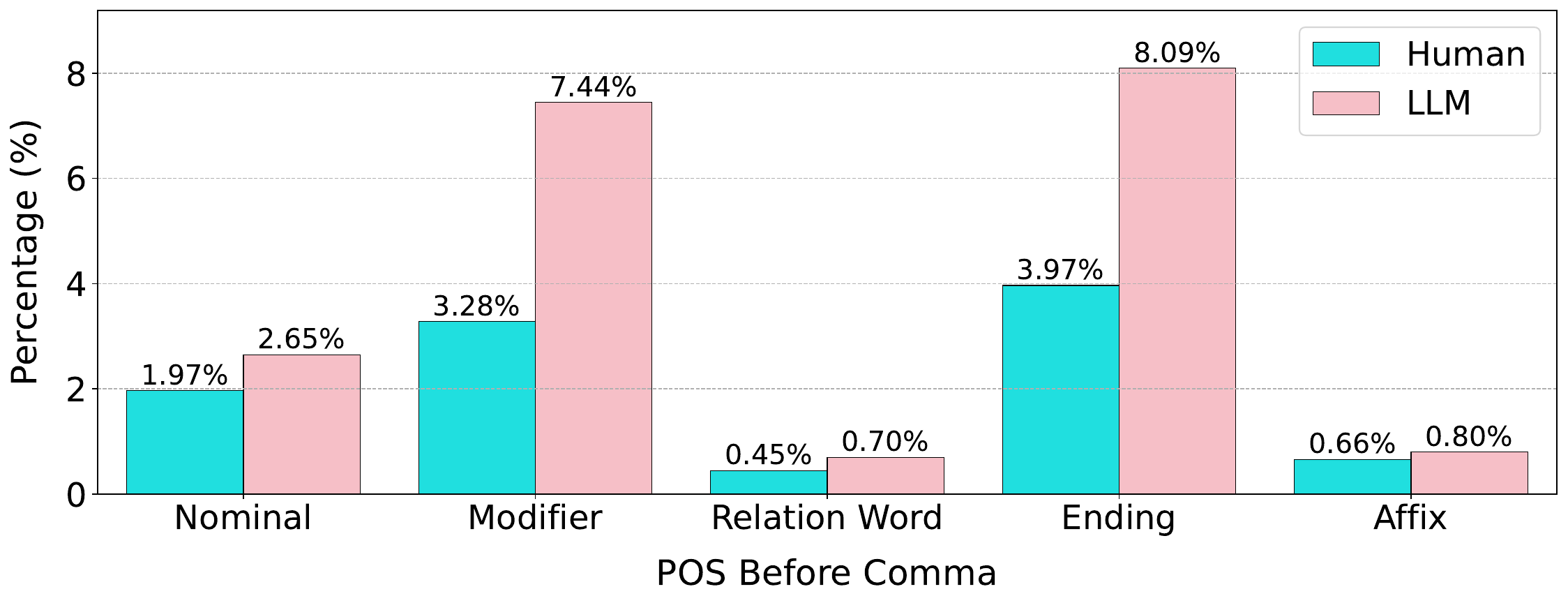}
        \subcaption{Paper Abstract}
    \end{minipage}
    \caption{
    Histogram showing the percentage of each part of speech that appears before a comma, measured by the number of times a given POS is followed by a comma relative to its total occurrences in the text.
    }\label{fig:comma_pos}
\end{figure*}

Figure~\ref{fig:comma_pos} illustrates the ratio of each POS followed by a comma, 
calculated as the number of times a given POS appears before a comma 
divided by the total occurrences of that POS in the text. 
Endings and affixes are morpheme-level tags rather than POS tags. 
However, following the tagged output of the Bareun tagger, we include them in our analysis as a individual categories. 
Predicates and interjections are excluded due to their sparsity. 

\paragraph{Use of Commas After Endings}
While the frequency of a comma after a POS is consistently higher 
for LLM-generated text, 
the difference is particularly notable in endings. 
The ratios of a connective ending followed by a comma are 
19.83\%, 15.57\%, and 28.01\% for LLM-generated essays, poetry, and paper abstracts, respectively, 
while they are 4.10\%, 4.68\%, and 13.27\% for human-written text. 
This discrepancy indicates that LLMs systematically overuse commas with connective endings, diverging from common patterns in human-written Korean text.

\paragraph{Use of Commas After Modifiers}
The Korean orthography guidelines state that it is natural 
not to use a comma after conjunctive adverbs such as 그리고~(‘and’ in function), 그러나, 그런데~(‘however’), and 그러므로~(‘therefore’), 
as the functions of conjunctive adverbs and commas overlap. 
Unlike English, which often requires a comma in such cases, Korean does not, but its use remains a matter of stylistic preference. 
The tendency of LLMs to insert unnecessary commas after these modifiers suggests an influence of English punctuation norms on their output.

\paragraph{Influence of Multilingual Training}  
The higher frequency of commas following connective endings and modifiers in LLM-generated Korean text 
may be influenced by the biases ingrained in multilingual language models. 
Prior research highlights how multilingual models exhibit preferences shaped by dominant training 
languages~\citep{wendler-etal-2024-llamas}. 
Given that LLMs are trained on multilingual data with English-dominated corpora, 
they may internalize English punctuation conventions, where commas frequently appear before conjunctions. 
This could lead LLMs to insert commas after connective endings more often than native Korean writers.

\subsection{Before and After Comma}
\begin{figure*}
    \centering
    \begin{minipage}{\textwidth}
        \centering
        \includegraphics[width=0.85\textwidth]{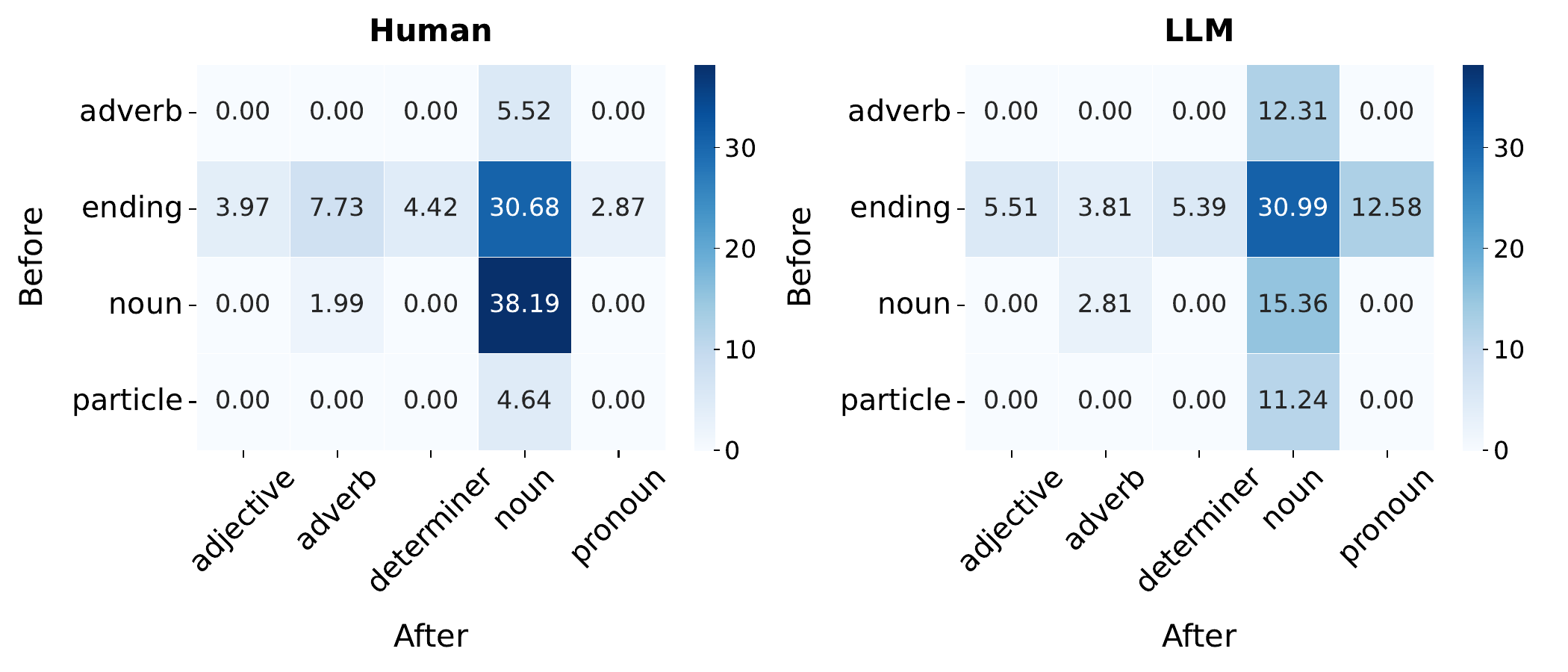}
        \subcaption{Essay}
    \end{minipage}
    \begin{minipage}{\textwidth}
        \centering
        \includegraphics[width=0.85\textwidth]{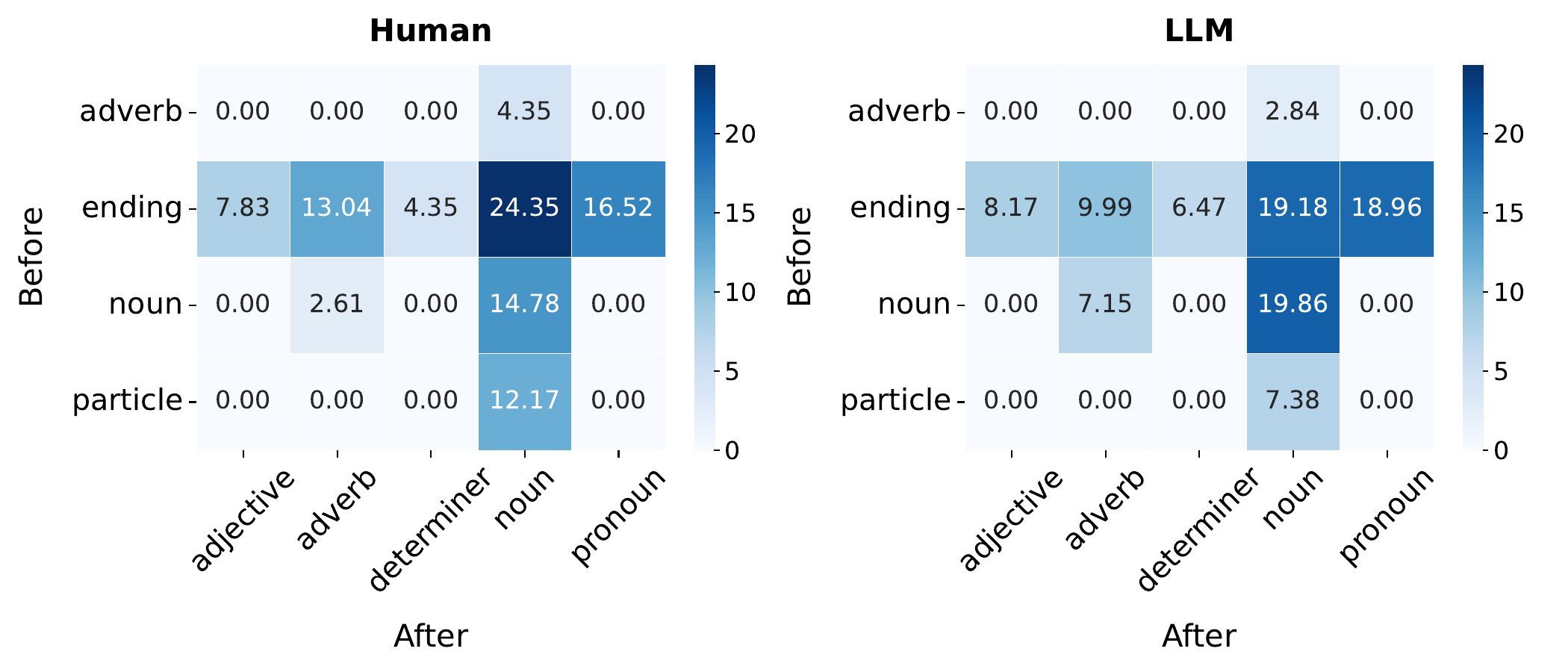}
        \subcaption{Poetry}
    \end{minipage}
    \begin{minipage}{\textwidth}
        \centering
        \includegraphics[width=0.85\textwidth]{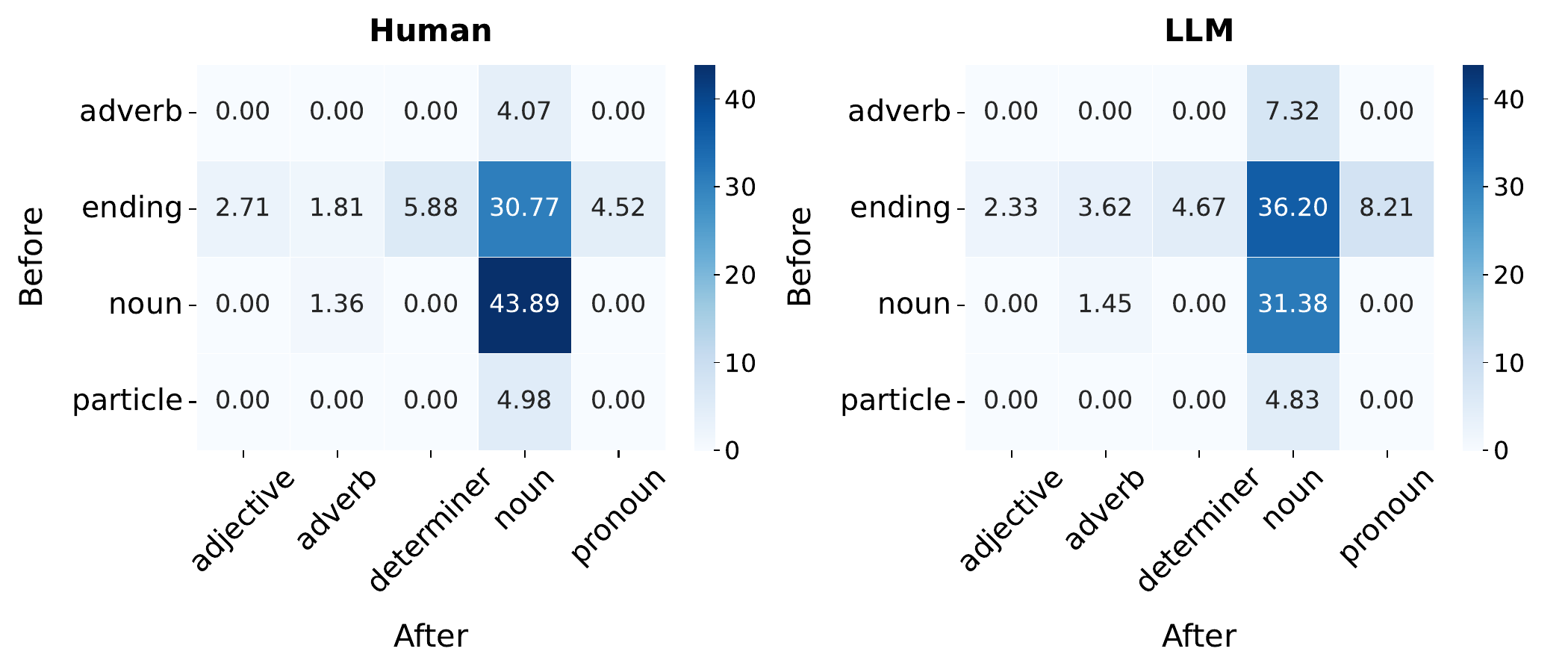}
        \subcaption{Paper Abstract}
    \end{minipage}
    \caption{
    Normalized POS pair frequencies around a comma, expressed as the percentage of each POS pair relative to the total number of comma uses.
    }\label{fig:comma_beforeafter_pos}
\end{figure*}

Figure~\ref{fig:comma_beforeafter_pos} shows the normalized POS pair frequencies, 
calculated as the proportion of each POS pair relative to the total number of comma uses. Sparse cases have been omitted. 
Both humans and LLMs primarily use commas in noun-noun and ending-noun pairs. 
Especially in essays, however, human-written texts exhibit a more concentrated pattern, 
with a few POS pairs showing high frequencies.
In contrast, LLM-generated text tends to display a more evenly distributed pattern across different POS pairs. 
This suggests that humans use commas more selectively, whereas LLMs apply them more broadly.

\clearpage

\section{Summary of Linguistic Features}
\label{sec:feature_summary}

This section presents a summary of the linguistic features employed in \texttt{KatFishNet}. 
The feature set falls into three primary categories: 
word spacing, POS combinations, and punctuation. 
Table~\ref{tab:linguistic_features} outlines each feature with a brief description. 

\begin{table}[h]
\centering
\small
\begin{tabular}{l|l|p{9cm}}
\toprule
Category & Feature & Description \\
\midrule
\multirow{1}{*}{Word Spacing} 
& MMN-BN Space Ratio & The frequency of word spacing between a MMN and a BN. \\
& BN Space Ratio & The frequency of word spacing before a BN. \\
& VX Space Ratio & The frequency of word spacing before a VX, 
excluding the specific case of "-아"/"-어" (ENDING) + "지" (VX), 
where spacing is strictly prohibited. \\
\midrule
\multirow{1}{*}{POS Combinations} 
& N-gram Diversity & The syntactic variability in a text by 
computing the ratio of unique POS n-grams to the total number of POS n-grams. \\
\midrule
\multirow{1}{*}{Punctuation} 
& Comma Inclusion Rate & The proportion of sentences containing at least one comma out of all sentences in a text.\\
& Comma Usage Rate & The number of commas in a sentence divided by the total number of morphemes in that sentence. \\
& Relative Position & The position of each comma~(counting the number of morphemes before it) divided by the total number of morphemes in the sentence. \\
& Segment Length & The average length of sentence segments split by commas. \\
& POS Diversity Score & The diversity of part-of-speech pairs appearing 
before and after a comma. \\
\bottomrule
\end{tabular}
\caption{Overview of linguistic features used by \texttt{KatFishNet} across three categories.}
\label{tab:linguistic_features}
\end{table}

\section{Implementation Details}
\label{sec:appendix_implementation_details}
When building the \texttt{KatFish} dataset, 
we access GPT-4o and Solar through their official APIs and use 
Qwen2 and Llama3.1 via Ollama\footnote{\url{https://ollama.com/search}}.
We implement the proposed detection method, 
\texttt{KatFishNet}, using machine learning models provided by 
Scikit-learn.
For the fine-tuning baseline, we use the 
chatgpt-detector-roberta\footnote{\url{https://huggingface.co/Hello-SimpleAI/chatgpt-detector-roberta}} model released by the HC3 dataset authors on HuggingFace as 
the base model and fine-tune it for five epochs using the \texttt{KatFish} dataset. 
When reproducing confidence-based and perturbation-based baselines, we use the 
implementations provided by MGTBench~\citep{he2023mgtbench}. 
For LLM paraphrasing and LLM prompting baselines, we access Exaone~3.5 through Ollama.
We conduct experiments on a server equipped with an NVIDIA RTX A6000 
with 48GB of memory.

\clearpage

\section{Performance of \texttt{KatFishNet} with Different Machine Learning Models}
\label{sec:appendix_experimental_results}

Table~\ref{tab:appendix_experimental_results} 
presents the performance of \texttt{KatFishNet} with different base machine learning models.
Below is a brief overview of the models incorporated in \texttt{KatFishNet}:
\begin{itemize}
    \item \textbf{Logistic Regression}: A statistical model used when the dependent variable 
    is categorical. 
    It applies a sigmoid function to the input values to compute probabilities, 
    and the regression coefficients help interpret how each feature influences the outcome.
    \item \textbf{Random Forest}: An ensemble method that constructs multiple decision trees 
    and combines their predictions to produce the final output. 
    Each tree is trained on randomly selected subsets of the data and features, 
    thus enhancing the diversity and generalization capability of the model.
    \item \textbf{Support Vector Machine}: A classification algorithm that seeks an 
    optimal decision boundary to separate data points. 
    It learns by maximizing the margin between the decision boundary and the 
    closest data points. 
    This approach is effective in high-dimensional spaces and can handle complex data 
    structures well.
\end{itemize}

We find that \texttt{KatFishNet}, 
which leverages comma usage patterns, 
achieves the highest performance in detecting 
LLM-generated Korean text, 
regardless of the type of machine learning model 
used as the backbone.
Looking ahead, we also plan to develop an expanded version of \texttt{KatFishNet}
based on ensemble or voting methods that integrate predictions from a diverse set 
of machine learning models. 
By understanding and leveraging the unique strengths of each model, 
we anticipate that our approach will achieve even stronger detection performance and robustness.

\begin{table*}[ht!]
\centering\small
\begin{tabular}{l|l|l|ccc|c}
\hline
\noalign{\hrule height 0.8pt}

Genre & \multicolumn{2}{c|}{\texttt{KatFishNet}} & $\rightarrow$ Solar & $\rightarrow$ Qwen2 & $\rightarrow$ Llama3.1 & Average
\\ 
\hline

\multirow{9}{*}{Essay} 
 & \multirow{3}{*}{Word Spacing} 
& Logistic Regression & 86.00 & 80.63 & 71.91 & 79.51 
\\
& & Random Forest & 81.56 & 75.53 & 68.35 & 75.14 
\\ 
& & Support Vector Machine & 82.46 & 78.29 & 69.08 & 76.61 
\\ 
\cline{2-7}

 & \multirow{3}{*}{POS Combinations} 
& Logistic Regression & 92.26 & 83.10 & 73.63 & 82.99 
\\
& & Random Forest & 91.85 & 80.08 & 75.08 & 82.33
\\ 
& & Support Vector Machine & 89.03 & 73.60 & 76.02 & 79.55 
\\ 
\cline{2-7}

 & \multirow{3}{*}{Punctuation} 
& Logistic Regression & 97.57 & 94.63 & 92.45 & 94.88 
\\
& & Random Forest & 96.07 & 95.12 & 90.29 & 93.82 
\\ 
& & Support Vector Machine & 96.26 & 95.49 & 91.33 & 94.36 
\\
\hline
\noalign{\hrule height 0.8pt}

\multirow{9}{*}{Poetry} & \multirow{3}{*}{Word Spacing} 
& Logistic Regression & 71.85 & 65.56 & 43.81 & 60.40 
\\
& & Random Forest & 59.24 & 63.04 & 52.47 & 58.25 
\\ 
& & Support Vector Machine & 67.33 & 69.79 & 47.96 & 61.69 
\\ 
\cline{2-7}

 & \multirow{3}{*}{POS Combinations} 
& Logistic Regression & 39.41 & 79.17 & 53.32 & 57.30 
\\
& & Random Forest & 45.99 & 69.27 & 55.52 & 56.92 
\\ 
& & Support Vector Machine & 43.66 & 75.88 & 56.53 & 58.69 
\\ 
\cline{2-7}

 & \multirow{3}{*}{Punctuation} 
& Logistic Regression & 62.65 & 93.45 & 63.22 & 73.10 
\\
& & Random Forest & 63.21 & 90.63 & 60.91 & 71.58 
\\ 
& & Support Vector Machine & 60.76 & 87.92 & 60.65 & 69.77 
\\
\hline
\noalign{\hrule height 0.8pt}

\multirow{9}{*}{Paper Abstract} & \multirow{3}{*}{Word Spacing} 
& Logistic Regression & 57.73 & 66.91 & 49.36 & 58.00
\\
& & Random Forest & 51.60 & 49.67 & 43.70 & 48.32 
\\ 
& & Support Vector Machine & 55.86 & 63.38 & 46.99 & 55.41 
\\ 
\cline{2-7}

 & \multirow{3}{*}{POS Combinations} 
& Logistic Regression & 47.47 & 70.05 & 42.47 & 53.33 
\\
& & Random Forest & 54.05 & 55.29 & 43.92 & 51.08 
\\ 
& & Support Vector Machine & 52.62 & 60.82 & 49.37 & 54.27 
\\ 
\cline{2-7}

 & \multirow{3}{*}{Punctuation} 
& Logistic Regression & 78.99 & 77.47 & 70.41 & 75.62 
\\
& & Random Forest & 75.51 & 77.32 & 67.61 & 73.48 
\\ 
& & Support Vector Machine & 79.51 & 76.35 & 69.59 & 75.15

\\
\hline
\noalign{\hrule height 0.8pt}

\end{tabular}
\caption{
Performance of detecting LLM-generated Korean text. 
We report the average performance~(AUC-ROC) over five experiments.
}\label{tab:appendix_experimental_results}
\end{table*}

\clearpage

\section{Ensemble of Linguistic Features}
\label{sec:ensemble_features}

We conduct an additional experiment to explore the effectiveness of combining word spacing, 
POS combinations, and punctuation features for LLM-generated Korean text detection. 
Specifically, we apply an ensemble approach by averaging the 
predicted probabilities~(ranging from 0 to 1) from separate logistic regression models. 
A value of 1 indicates that the text is predicted to be LLM-generated.
Tables~\ref{tab:ensemble_essay},~\ref{tab:ensemble_poetry}, and~\ref{tab:ensemble_abstract} show 
the experimental results.

\begin{table}[h]
\centering
\begin{tabular}{l|c|c|c|c}
\toprule
Feature Combination & $\rightarrow$ Solar & $\rightarrow$ Qwen2 & $\rightarrow$ Llama3.1 & Average \\
\midrule
Word Spacing            & 86.00  & 80.63  & 71.91  & 79.51 \\
POS Combinations        & 92.26  & 83.10  & 73.63  & 82.99 \\
Punctuation             & 97.57  & 94.63  & 92.45  & 94.88 \\
All                     & 97.67  & 93.93  & 88.05  & 93.21 \\
POS + Punctuation       & 98.97  & 95.09  & 90.58  & 94.88 \\
POS + Spacing           & 90.29  & 83.65  & 73.73  & 82.55 \\
Punctuation + Spacing   & 97.89  & 94.74  & 90.35  & 94.32 \\
\bottomrule
\end{tabular}
\caption{Performance on LLM-generated essay detection using individual 
and ensemble combinations 
of linguistic features. 
The ensemble method averages outputs from separate logistic regression classifiers.}
\label{tab:ensemble_essay}
\end{table}

\begin{table}[h]
\centering
\begin{tabular}{l|c|c|c|c}
\toprule
Feature Combination & $\rightarrow$ Solar & $\rightarrow$ Qwen2 & $\rightarrow$ Llama3.1 & Average \\
\midrule
Word Spacing            & 71.85  & 65.56  & 43.81  & 60.40 \\
POS Combinations        & 39.41  & 79.17  & 53.32  & 57.30 \\
Punctuation             & 62.65  & 93.45  & 63.22  & 73.10 \\
All                     & 63.66  & 96.04  & 59.94  & 73.21 \\
POS + Punctuation       & 55.30  & 94.83  & 63.40  & 71.17 \\
POS + Spacing           & 52.09  & 80.27  & 48.69  & 60.35 \\
Punctuation + Spacing   & 73.78  & 95.41  & 58.64  & 75.94 \\
\bottomrule
\end{tabular}
\caption{Performance on LLM-generated poetry detection using individual 
and ensemble combinations 
of linguistic features. 
The ensemble method averages outputs from separate logistic regression classifiers.}
\label{tab:ensemble_poetry}
\end{table}

\begin{table}[h]
\centering
\begin{tabular}{l|c|c|c|c}
\toprule
Feature Combination & $\rightarrow$ Solar & $\rightarrow$ Qwen2 & $\rightarrow$ Llama3.1 & Average \\
\midrule
Word Spacing            & 57.73  & 66.91  & 49.36  & 58.00 \\
POS Combinations        & 47.47  & 70.05  & 42.47  & 53.33 \\
Punctuation             & 78.99  & 77.47  & 70.41  & 75.62 \\
All                     & 68.93  & 80.00  & 57.01  & 68.64 \\
POS + Punctuation       & 71.70  & 82.82  & 60.93  & 71.81 \\
POS + Spacing           & 53.22  & 70.76  & 44.98  & 56.32 \\
Punctuation + Spacing   & 77.81  & 79.70  & 65.63  & 74.38 \\
\bottomrule
\end{tabular}
\caption{Performance on LLM-generated paper abstract detection using individual 
and ensemble combinations 
of linguistic features. 
The ensemble method averages outputs from separate logistic regression classifiers.}
\label{tab:ensemble_abstract}
\end{table}

In most cases, punctuation alone achieves detection performance that is either higher 
or comparable to that of ensemble models. 
This reinforces the strong effectiveness of punctuation-related features in 
detecting LLM-generated Korean text.
Combining POS and spacing leads to the lowest performance among the ensembles, 
suggesting that these features alone do not capture the high-impact signals 
that punctuation features provide.
Building on these findings, we plan to explore ensemble or voting approaches 
using multiple machine learning models, with a particular emphasis on punctuation features. 

\clearpage

\section{Methods for Detection Machine-Generated Text}
\label{sec:appendix_detection_methods}
This section serves as supplementary material to Section~\ref{sec:related_work}, 
providing an overview of methodologies for detecting LLM-generated text.
Methods for detecting machine-generated text can be broadly grouped into 
three main categories: 
\paragraph{Watermarking} 
Watermarking methods~\citep{kirchenbauer2023watermark,guo2024context} detect 
machine-generated text by embedding 
recognizable patterns during the text generation process. 
These methods intervene in the text generation process of LLMs 
by dividing the vocabulary into green and red lists and 
prioritizing the generation of tokens from the green list. 
If the proportion of green tokens in a generated text exceeds 
a certain threshold, the text is classified as machine-generated. 

\paragraph{White-Box Detection} 
White-box detection methods detect machine-generated text by 
analyzing the log probability~\citep{mitchell2023detectgpt} 
or log rank~\citep{su-etal-2023-detectllm} of the text. 
These methods rely on full or partial access to the text generator, 
which makes them difficult to use with commercial LLMs that restrict such access. Recently, \citet{mireshghallah-etal-2024-smaller} proposed the use of 
surrogate models~(e.g., GPT-2) to approximate log probability or 
log rank.

\paragraph{Black-Box Detection} 
Black-box detection methods require only the target text 
and do not rely on access to the text generator. 
For example, \citet{zhu-etal-2023-beat} introduced a 
paraphrasing-based detection approach using LLMs. 
Their method is based on the hypothesis that machine-generated text 
aligns more closely with the generation logic and statistical patterns 
learned by LLMs than human-written text. 
Accordingly, they proposed identifying a text as machine-generated 
if its paraphrased version closely resembles the original.

\section{Detection Results From the LLMs Used to Generate the \texttt{KatFish} Dataset}
\label{sec:appendix_llm_prompting}
We conduct an additional experiment to investigate whether the LLM that 
generated a given text can reliably distinguish between its own outputs and human-written texts.
Table~\ref{tab:appendix_experimental_results_llm_prompting} 
presents the performance of the LLM-generated Korean 
text detection task using the LLMs 
employed in the creation of the \texttt{KatFish} dataset. 
The results suggest that even LLMs struggle to identify 
their own generated text, 
highlighting the limitations of LLM prompting baseline.

\begin{table*}[ht!]
\centering\small
\begin{tabular}{l|l|ccc|c}
\hline
\noalign{\hrule height 0.8pt}

Genre & Detection LLMs & $\rightarrow$ Solar & $\rightarrow$ Qwen2 & $\rightarrow$ Llama3.1 & Average
\\ 
\hline

\multirow{4}{*}{Essay} 
& GPT-4o & 57.75 & 49.70 & 46.26 & 51.23
\\ 
& Solar & 46.50 & 48.11 & 47.38 & 47.33
\\ 
& Qwen2 & 62.04 & 61.03 & 54.79 & 59.28
\\ 
& Llama3.1 & 52.64 & 37.95 & 45.11 & 45.23 
\\
\hline
\noalign{\hrule height 0.8pt}

\multirow{4}{*}{Poetry} 
& GPT-4o & 51.70 & 52.28 & 50.06 & 51.34
\\ 
& Solar & 49.99 & 51.15 & 44.34 & 48.49
\\ 
& Qwen2 & 47.17 & 49.91 & 45.79 & 47.62
\\ 
& Llama3.1 & 64.93 & 48.58 & 50.60 & 54.70 
\\
\hline
\noalign{\hrule height 0.8pt}

\multirow{4}{*}{Paper Abstract} 
& GPT-4o & 48.20 & 47.70 & 47.77 & 47.89 
\\ 
& Solar & 51.49 & 52.32 & 53.05 & 52.28
\\ 
& Qwen2 & 50.00 & 50.50 & 49.52 & 50.00 
\\ 
& Llama3.1 & 49.80 & 49.32 & 49.68 & 49.60 
\\
\hline
\noalign{\hrule height 0.8pt}

\end{tabular}
\caption{Performance Comparison of Different LLMs in Detecting LLM-Generated Korean Text. 
We report the average performance~(AUC-ROC) over five experiments.
We separately report the performance of the detection model for the 
task of distinguishing between human-written text and text 
generated by a specific LLM.}\label{tab:appendix_experimental_results_llm_prompting}
\end{table*}

\clearpage

\section{Detection with Knowledge-Integrated Few-Shot Prompting}
\label{sec:knowledge_integrated_few_shot_prompting}

We present an additional experiment that incorporates linguistic insights into the 
prompt for LLM-generated text detection. 
The experiment focuses on comma usage patterns, 
identified as the most effective feature in our main analysis.

\subsection{Experimental Setup}

We follow the equivalent dataset and experimental settings 
described in Section~\ref{sec:ood_evaluation}. 
We use Exaone~3.5, a Korean-focused LLM for this additional experiment. 
We compare the new experimental results with the simpler prompting approach from 
Table~\ref{tab:experimental_results}.

\subsection{Prompt Design}

\paragraph{Zero-shot Learning}
\begin{verbatim}
Distinguish between Korean text written by a human and Korean text generated by an LLM.
Provide the basis for your decision under "Reasoning:".
Indicate the probability that the given text was generated by an LLM 
on a scale from 0 to 1 under "Answer:".

Korean texts written by LLMs tend to use commas more frequently 
than those written by humans.
Additionally, LLMs are more likely to place commas toward the latter part of a sentence.
They also tend to use a wider variety of part-of-speech combinations around commas.
Use these characteristics to help you distinguish the text.

Text to be classified: {t}
Reasoning:
Answer:
\end{verbatim}

\paragraph{Few-shot Learning}
\begin{verbatim}
Distinguish between Korean text written by a human and Korean text generated by an LLM.
Provide the basis for your decision under "Reasoning:".
Indicate the probability that the given text was generated by an LLM 
on a scale from 0 to 1 under "Answer:".

Korean texts written by LLMs tend to use commas more frequently 
than those written by humans.
Additionally, LLMs are more likely to place commas toward the latter part of a sentence.
They also tend to use a wider variety of part-of-speech combinations around commas.
Use these characteristics to help you distinguish the text.

Here are examples of Korean texts written by a human and by an LLM:
Human-written Korean text 1: {h1}
Human-written Korean text 2: {h2}
LLM-generated Korean text 1: {g1}
LLM-generated Korean text 2: {g2}
Text to be classified: {t}
Reasoning:
Answer:
\end{verbatim}

\subsection{Three Prompting Methods}

We compare three prompting methods:
\begin{itemize}
    \item \textbf{No Knowledge}: The default LLM prompting approach as in 
    Table~\ref{tab:experimental_results}~(label: LLM Prompting Exaone~3.5), 
    where the prompt simply asks whether the text is human-written or LLM-generated, 
    without providing specific linguistic insights.
    \item \textbf{Zero-shot Learning}: 
    The new prompt that provides linguistic insights about commas, 
    without example texts of human/LLM writing.
    \item \textbf{Few-shot Learning}: 
    The new prompt that provides linguistic insights about commas 
    and two examples each of GPT4o-generated and human-written Korean texts.
\end{itemize}

\subsection{Findings}

Table~\ref{tab:knowledge_integrated_few_shot_learning} shows the experimental results:
\begin{itemize}
    \item For poetry and paper abstracts, prompts that provide linguistic guidance about 
    comma usage improve detection performance over the baseline. 
    While Zero-shot yields limited benefit in essays, 
    the Few-shot prompt leads to a noticeable improvement.
    \item Providing both linguistic insights and representative examples further 
    enhances detection accuracy in nearly all cases. 
\end{itemize}

\begin{table}[h]
\centering
\begin{tabular}{l|l|c|c|c|c}
\toprule
\textbf{Genre} & \textbf{Prompting} & \textbf{Solar} & \textbf{Qwen2} & \textbf{Llama3.1} & \textbf{Average} \\
\midrule
\multirow{3}{*}{Essay}
  & No Knowledge & 50.42 & 49.74 & 50.07 & 50.07 \\
  & Zero-shot     & 51.89 & 47.98 & 46.39 & 48.75 \\
  & Few-shot      & 56.10 & 51.70 & 47.92 & \textbf{51.90} \\
\midrule
\multirow{3}{*}{Poetry}
  & No Knowledge & 50.53 & 50.16 & 49.42 & 50.03 \\
  & Zero-shot     & 62.51 & 66.03 & 50.06 & 59.53 \\
  & Few-shot      & 67.16 & 66.75 & 57.35 & \textbf{63.75} \\
\midrule
\multirow{3}{*}{Paper Abstract}
  & No Knowledge & 48.60 & 46.41 & 47.18 & 47.39 \\
  & Zero-shot     & 52.80 & 64.26 & 62.25 & 59.77 \\
  & Few-shot      & 53.81 & 66.61 & 59.91 & \textbf{60.11} \\
\bottomrule
\end{tabular}
\caption{Comparison of detection performance across three LLM prompting strategies.}
\label{tab:knowledge_integrated_few_shot_learning}
\end{table}

These results highlight the usefulness of incorporating linguistic insights—specifically, 
the differences in comma usage—into the prompt. 
We believe the findings signify the importance of detailed linguistic analysis, 
which can be cataloged and utilized in LLM context-learning approaches. 
We plan to explore this approach in more detail in our future work. 
For example, we may integrate chain-of-thought or other advanced prompting techniques 
with our linguistic analysis to further improve LLM-generated text detection.

\clearpage

\section{Robustness of \texttt{KatFishNet} Against Human-Style Imitation via LLM Prompting}
\label{sec:human_style_imitation}

We investigate the robustness of \texttt{KatFishNet} in scenarios where LLMs are prompted to 
imitate human writing style.
We experiment to see how effectively \texttt{KatFishNet} can detect text that has been 
deliberately revised to appear more \textbf{human-like}, 
focusing on comma-related linguistic features.

\subsection{Experimental Setup}

We follow the experimental settings described in Section~\ref{sec:ood_evaluation}. 
Specifically:
\begin{itemize}
    \item We select 20\% of the human-written texts 
    and all texts generated by Solar for our data.
    \item We then feed each Solar-generated text back into Solar, 
    instructing it to revise the text in such a way that it avoids LLM-like comma usage 
    patterns~(\eg, frequent comma placement, commas toward the end of sentences, 
    diverse POS combinations around commas). 
    Essentially, we ask the model to fix the text to make it more closely resemble human writing.
    \item We combine these revised Solar texts with the 20\% human-written texts 
    to form a new test dataset.
    \item We evaluate \texttt{KatFishNet}~(using comma-related features) on this revised dataset.
\end{itemize}

\subsection{Prompt Design}

We create prompts for essays, poetry, and paper abstracts, 
which all share the same structure except for specifying which text type is being revised. 
Below is the essay-specific prompt example:

\paragraph{Essay Prompt}
\begin{verbatim}
I'm going to give you a Korean essay written by an LLM. 
Please revise it to make it sound like it was written by a human. 
Provide the revised essay under "Revised Essay:".

Essays written by LLMs tend to use commas more frequently 
than those written by humans. 
They also tend to place commas toward the latter part of a sentence 
and use a wider variety of part-of-speech combinations around commas. 
Keep this in mind as you revise the essay to make it more human-like.

Essay to revise: {t}
Revised Essay:
\end{verbatim}

\subsection{Experimental Results}

Table~\ref{tab:robustness_prompting_results} 
compares Original~(the results from Table~\ref{tab:experimental_results}, under the ``$\rightarrow$ Solar'' column using Punctuation-based features) 
with Revised~(detection performance on the newly revised texts).

\begin{table}[h]
\centering
\begin{tabular}{l|c|c}
\toprule
\textbf{Genre} & \textbf{Original} & \textbf{Revised} \\
\midrule
Essay & 97.57 & 90.83 \\
Poetry & 62.65 & 75.91 \\
Paper Abstract & 78.99 & 80.69 \\
\bottomrule
\end{tabular}
\caption{\texttt{KatFishNet} performance on Solar-generated texts before and after 
human-style revision.}
\label{tab:robustness_prompting_results}
\end{table}

Observations: 
\begin{itemize}
    \item \textbf{Essays}: There is a performance drop from 97.57\% to 90.83\%, 
    yet the score remains above 90. 
    This suggests that while the LLM manages some level of style manipulation, 
    \texttt{KatFishNet} still maintains strong detection performance.
    \item \textbf{Poetry \& Paper Abstracts}: Interestingly, we observe a performance 
    increase~(62.65$\rightarrow$75.91 for poetry; 78.99$\rightarrow$80.69 for abstracts). 
    This suggests that instructing the LLM to alter comma usage may yield text that 
    is even more easily flagged by the punctuation-based features of \texttt{KatFishNet}. 
    This may be because the LLM performs human-like revisions based on an internal conception 
    of human-written text, which does not fully align with actual human writing patterns.
\end{itemize}

These findings indicate that comma-related linguistic features are not trivially shallow. 
While commas may seem trivial, their usage reflects deeper, inherent characteristics of LLMs. 
Without a nuanced understanding of Korean stylistic conventions, 
LLMs struggle to convincingly mimic human-like text. 
Their multilingual nature also contributes to this challenge. 
This aligns with our hypothesis that LLMs may still have difficulty fully 
capturing the stylistic intricacies of Korean writing.

\subsection{Further Justification}

In our paper, we demonstrate that comma-related features are exceptionally useful for distinguishing between LLM-generated and human-written Korean texts. In particular:

\begin{itemize}
    \item As discussed in Section~\ref{comma_usage_patterns}, 
    LLMs are trained on extensive multilingual datasets—often with a heavy emphasis on English. 
    Since English comma usage differs considerably from Korean conventions, 
    LLMs tend to adopt comma usage patterns that do not align with those commonly found 
    in native Korean writing.
    \item As discussed in Section~\ref{sec:experimental_results}, 
    comma usage reflects contextual and stylistic factors, 
    making it highly variable depending on the intent of the writer.
    Consequently, it is challenging for LLMs to precisely mimic human writers.
    \item As detailed in Appendix~\ref{sec:pos_comma_analysis}, 
    we conduct an extensive analysis of the differences in part-of-speech patterns 
    surrounding commas between human-written and LLM-generated texts.
\end{itemize}

Together, these results strongly support our hypothesis that comma-related features 
go beyond shallow surface-level analysis and capture critical stylistic 
and syntactic nuances inherent to Korean writing.

\subsection{Future Directions}

Adversarial attacks that aim to deceive LLM-generated text detectors 
represent an intriguing avenue for future exploration. 
We plan to explore prompt-engineered adversarial attacks, 
leveraging the linguistic distinctions found between human- and LLM-generated Korean texts, 
as part of our subsequent research.

\clearpage

\section{Morphological Analysis}
\label{sec:appendix_morph}

We conduct a morphological analysis of Korean text
written by humans and LLMs to 
analyze the frequency of different parts of speech.
We perform the analysis using the HanNanum POS tagger, 
provided by the KoNLPy.
Table~\ref{tab:pos_name} shows the English names of each POS 
along with their corresponding Korean names. 
In Korean, words are classified into five word types 
(nominal, predicate, modifier, interjection, relation word) 
based on their function and into nine parts of speech based on their meaning (noun, pronoun, numeral, verb, adjective, determiner, adverb, interjection, particle).
\begin{table*}[hbt!]
\centering\scriptsize
\begin{tabular}{ll|l}
\hline
\noalign{\hrule height 0.8pt}
English& Korean & Description  
\\ 
\hline

\multirow{2}{*}{Nominal} & \multirow{2}{*}{체언} & A nominal is a part of speech that includes nouns, pronouns, and numeral classifiers, 
\\
& & functioning as the subject or object in a sentence.
\\
\hline 

\multirow{2}{*}{Predicate} & \multirow{2}{*}{용언} & A predicate is a part of speech that includes verbs and adjectives, expressing actions,
\\
& & states, or descriptions while serving as the main component of 
the predicate in a sentence.
\\
\hline 

\multirow{2}{*}{Modifier} & \multirow{2}{*}{수식언} & A modifier is a part of speech that includes adverbs and determiners, providing additional 
\\
& & information about other words, such as nouns or verbs, by describing or qualifying them.
\\
\hline 

\multirow{2}{*}{Interjection} & \multirow{2}{*}{독립언} & An interjection is a part of speech that expresses sudden emotions or reactions and stands
\\
& & independently within a sentence, often without a grammatical connection to other words.
\\
\hline 

\multirow{2}{*}{Relation Word} & \multirow{2}{*}{관계언} & A relation word, or particle, is a part of speech that links nouns, pronouns, or phrases to other 
\\
& & words in a sentence, indicating grammatical relations such as subject, object, or possession.
\\
\hline 

\multirow{2}{*}{Ending} & \multirow{2}{*}{어미} & An ending is a part of speech attached to the stem of a verb or adjective, modifying its 
\\
& & meaning, tense, mood, or form, and determining the sentence’s grammatical structure.
\\
\hline 

\multirow{2}{*}{Affix} & \multirow{2}{*}{접사} & An affix is a part of speech that attaches to a root word, altering its meaning 
\\
& & or grammatical function, and includes prefixes, suffixes, infixes, and circumfixes.
\\
\hline 

\multirow{2}{*}{Symbol} & \multirow{2}{*}{기호} & A symbol is a part of speech that includes non-alphabetic characters such as
\\
& & punctuation marks, numbers, and special signs, used to convey meaning or structure in writing.
\\
\hline 

\multirow{2}{*}{Foreign Language} & \multirow{2}{*}{외국어} & Foreign language refers to words or phrases borrowed from other languages, used within a text
\\
& & to convey specific meanings, often retaining their original form and pronunciation.
\\

\hline
\noalign{\hrule height 0.8pt}
\end{tabular}
\caption{English names of each part of speech along 
with their corresponding Korean names.
}\label{tab:pos_name}
\end{table*}

\begin{figure*}
    \centering
    \begin{minipage}{\textwidth}
        \centering
        \includegraphics[width=0.8\textwidth]{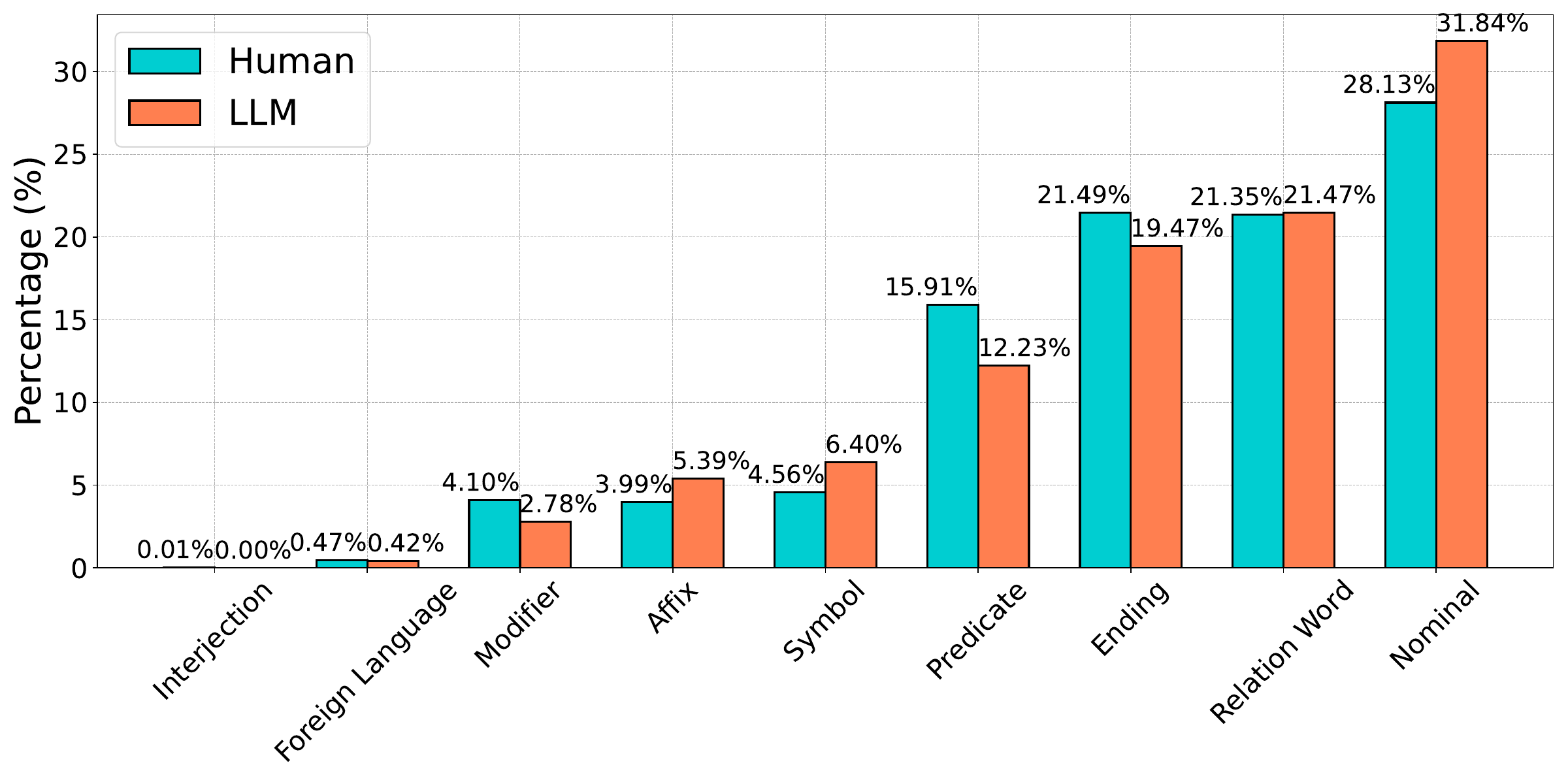}
        \subcaption{Essay}
    \end{minipage}
    \begin{minipage}{\textwidth}
        \centering
        \includegraphics[width=0.8\textwidth]{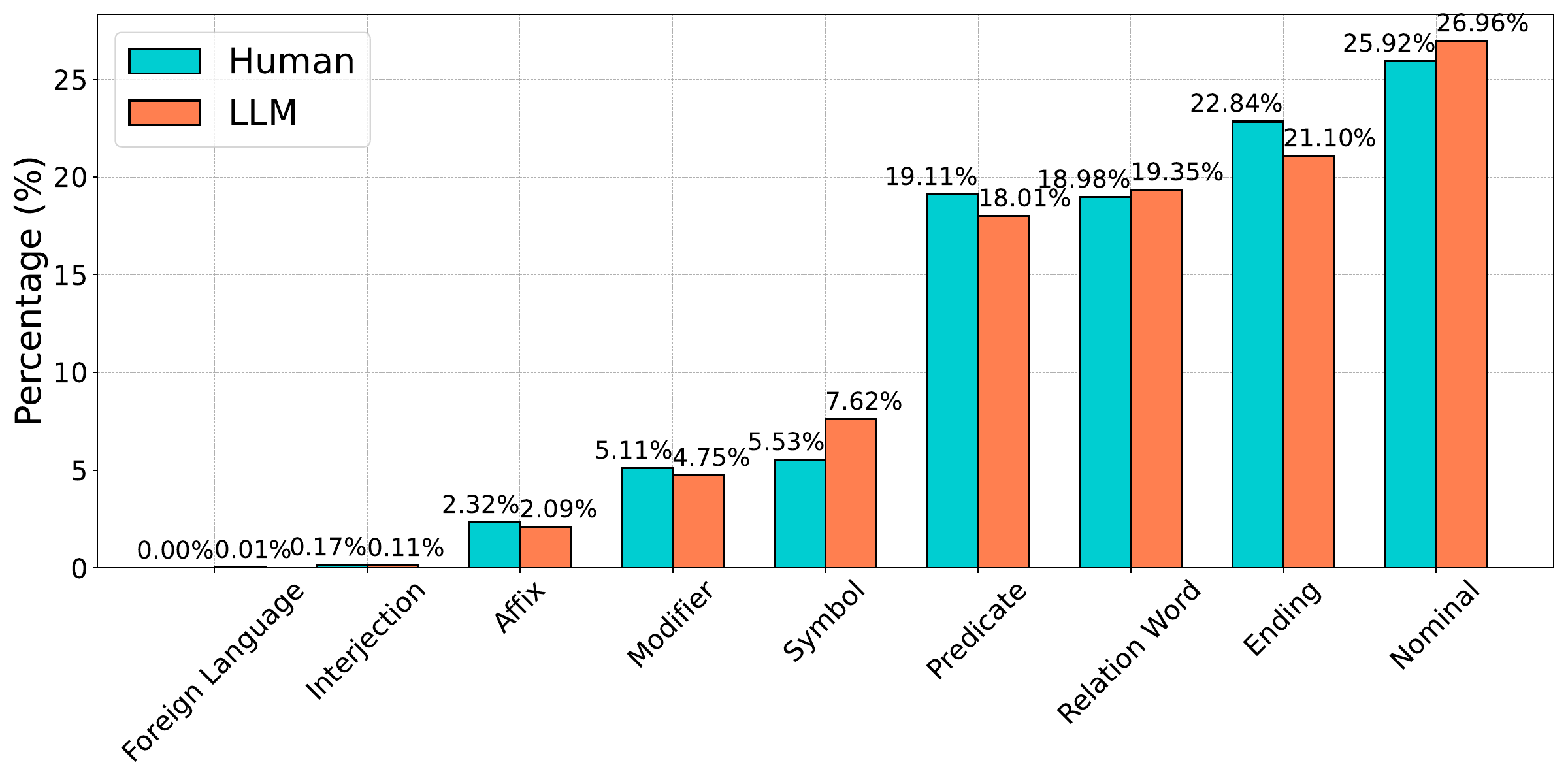}
        \subcaption{Poetry}
    \end{minipage}
    \begin{minipage}{\textwidth}
        \centering
        \includegraphics[width=0.8\textwidth]{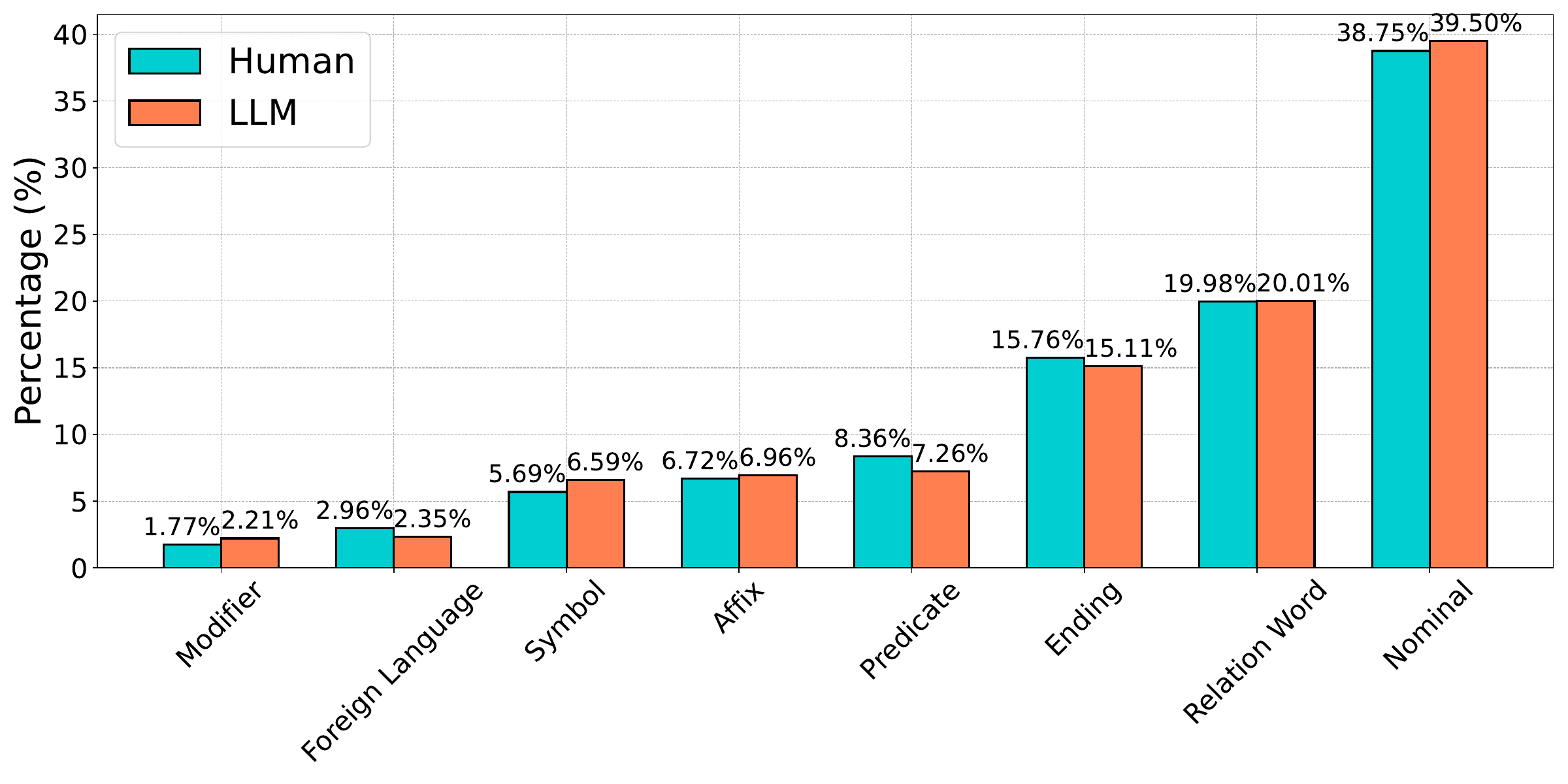}
        \subcaption{Paper Abstract}
    \end{minipage}
    \caption{
    Histogram visualizing the distribution 
    of part-of-speech usage in human-written and LLM-generated text.
    }\label{fig:pos_ana}
\end{figure*}

Figure~\ref{fig:pos_ana} 
visualizes the distribution of POS usage 
in human and LLM-written Korean text.

\subsection{Higher Usage of Endings and Predicates in Human-written text}
Endings and predicates significantly 
contribute to the diversity of sentence structure and expression. 
Humans naturally use a variety of sentence 
endings and predicates to connect sentences fluidly, 
express emotions, or reinforce arguments. 
These elements are essential for shaping the flow and rhythm of writing.

In contrast, LLM-generated text show relatively lower usage of endings 
and predicates. 
This indicates that LLMs often create simpler and more formulaic sentence structures.
In creative or emotionally rich writing, 
they tend to struggle with replicating the nuanced flow of sentences 
that human writers achieve. 
Guided by statistical patterns rather than the natural rhythm of language, 
LLMs frequently produce sentences that lack the dynamic 
quality characteristic of human writing.

\subsection{Higher Usage of Nominals in LLM-generated text}
LLMs often rely heavily on nouns, 
focusing on generating sentences that are technical or 
primarily aimed at delivering information. 
Nouns play a key role in clearly conveying topics or concepts, 
and LLMs frequently build sentences around them, 
resulting in expressions that are direct and concise.

In contrast, human writers use nouns less frequently than LLMs, 
opting for a more balanced mix of verbs, adjectives, 
and adverbs to add depth and nuance to their writing. 
Human writing typically exhibits greater flexibility and complexity 
in sentence structures, reflecting context and emotion more effectively.

\subsection{Overall Analysis}

\paragraph{Human Writing is More Dynamic in Expression and Structure}
The abundant use of endings and predicates makes human writing 
appear more natural, capturing the flow of emotions and logic. 
This is because humans tend to connect sentences 
organically and emphasize a variety of expressions.

\paragraph{LLM-generated text are Optimized for Information Delivery}
The higher frequency of nouns indicates that LLMs focus on clarity 
and brevity, often structuring sentences in 
a more repetitive manner. 

\clearpage

\section{Lexical Diversity Analysis}
\label{sec:appendix_zipf_heap}

\begin{figure}[h]
    \centering

    \begin{subfigure}{0.32\textwidth}
        \centering
        \includegraphics[width=\linewidth]{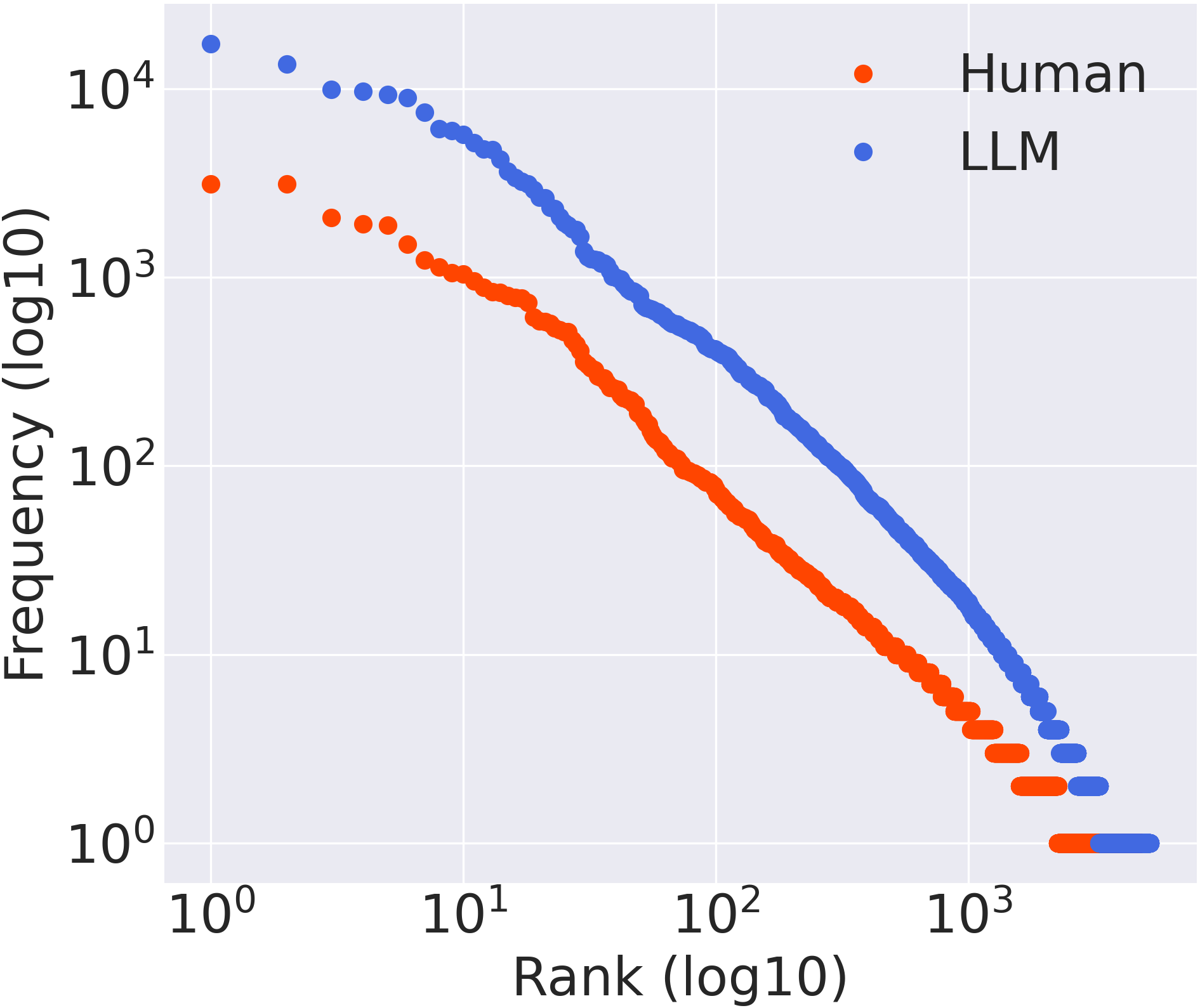}
        \caption{Zipf's Law for Essay}
    \end{subfigure}
    \begin{subfigure}{0.32\textwidth}
        \centering
        \includegraphics[width=\linewidth]{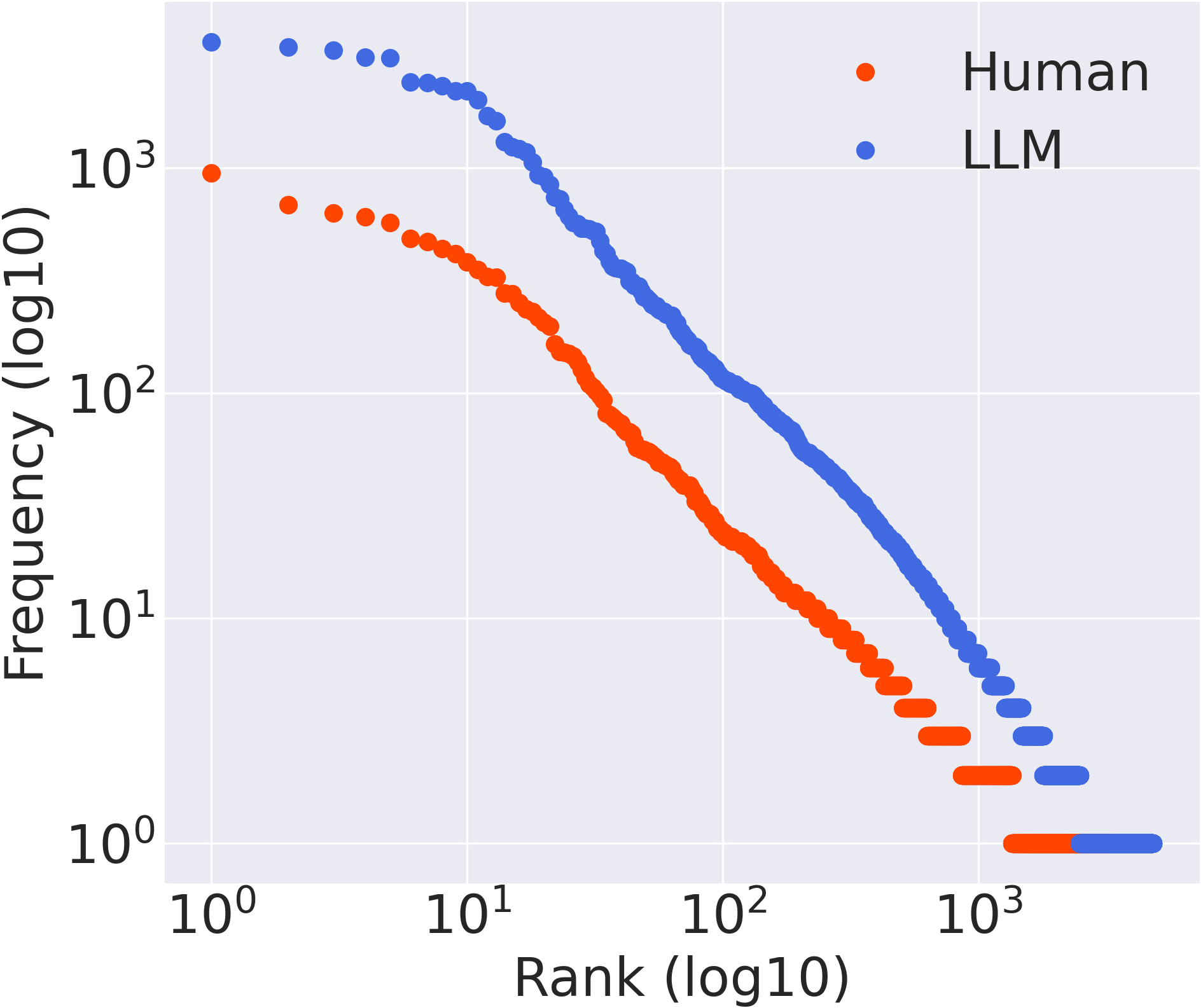}
        \caption{Zipf's Law for Poetry}
    \end{subfigure}
    \begin{subfigure}{0.32\textwidth}
        \centering
        \includegraphics[width=\linewidth]{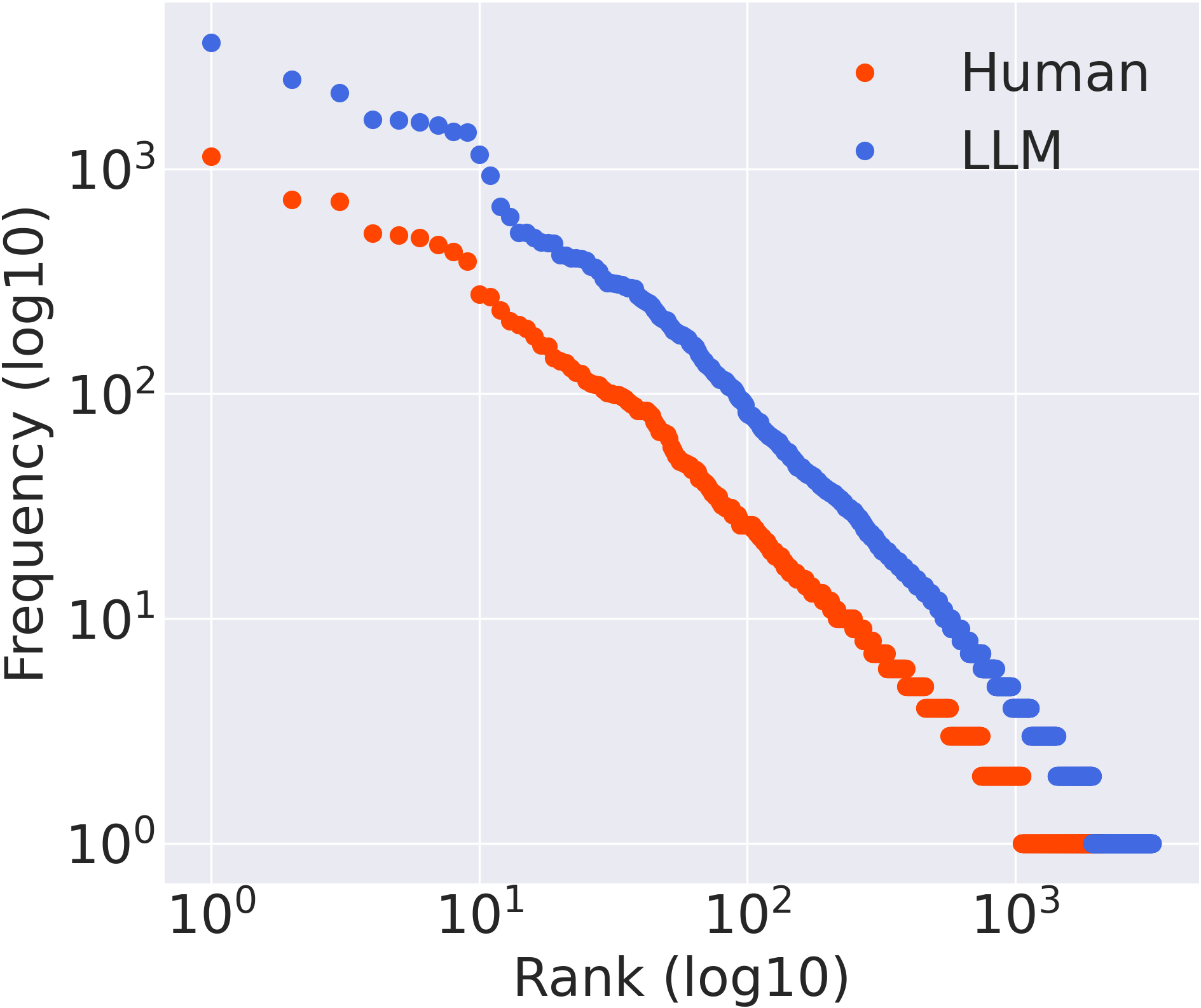}
        \caption{Zipf's Law for Paper Abstract}
    \end{subfigure}
    
    \begin{subfigure}{0.32\textwidth}
        \centering
        \includegraphics[width=\linewidth]{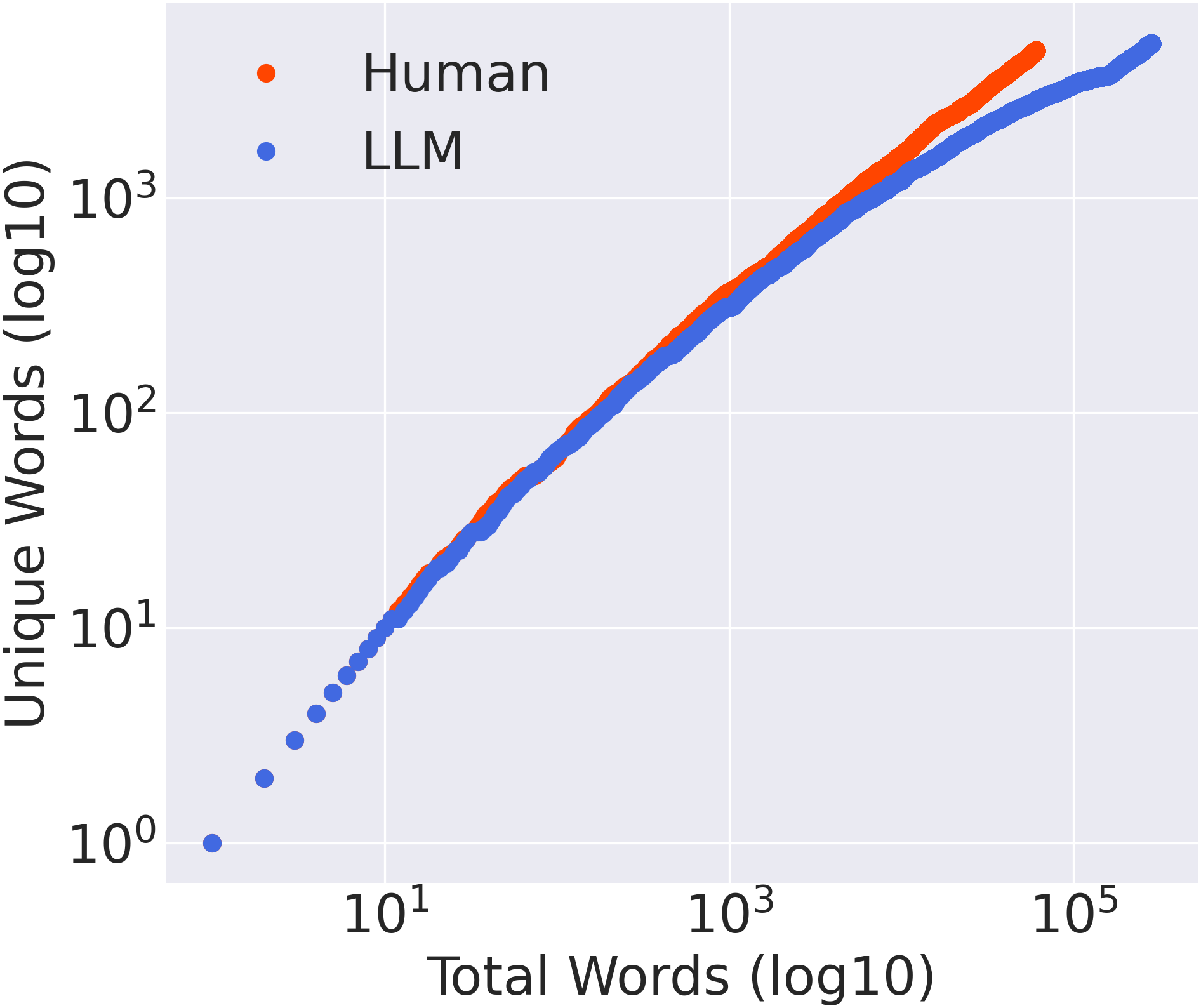}
        \caption{Heap's Law for Essay}
    \end{subfigure}
    \begin{subfigure}{0.32\textwidth}
        \centering
        \includegraphics[width=\linewidth]{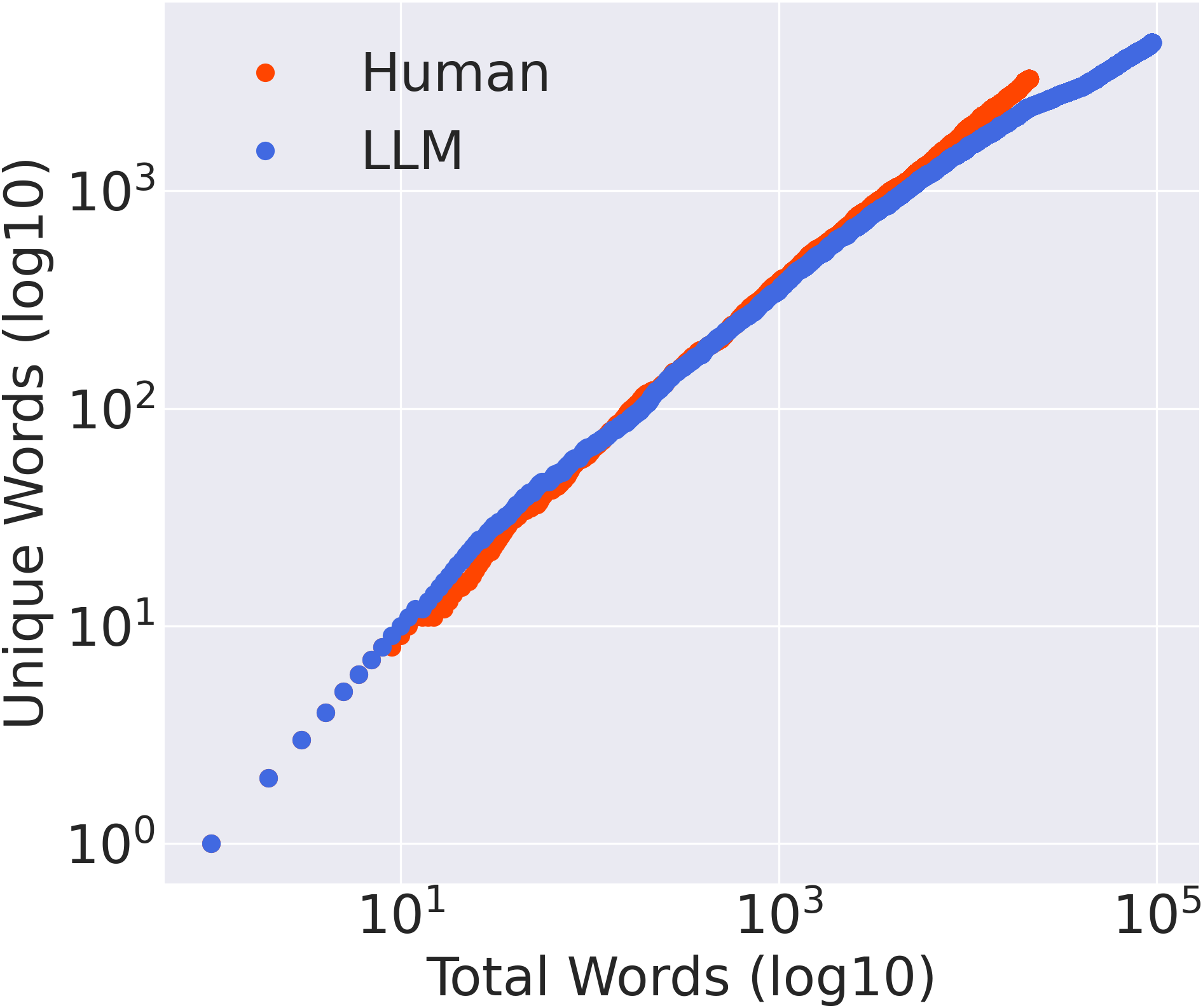}
        \caption{Heap's Law for Poetry}
    \end{subfigure}
    \begin{subfigure}{0.32\textwidth}
        \centering
        \includegraphics[width=\linewidth]{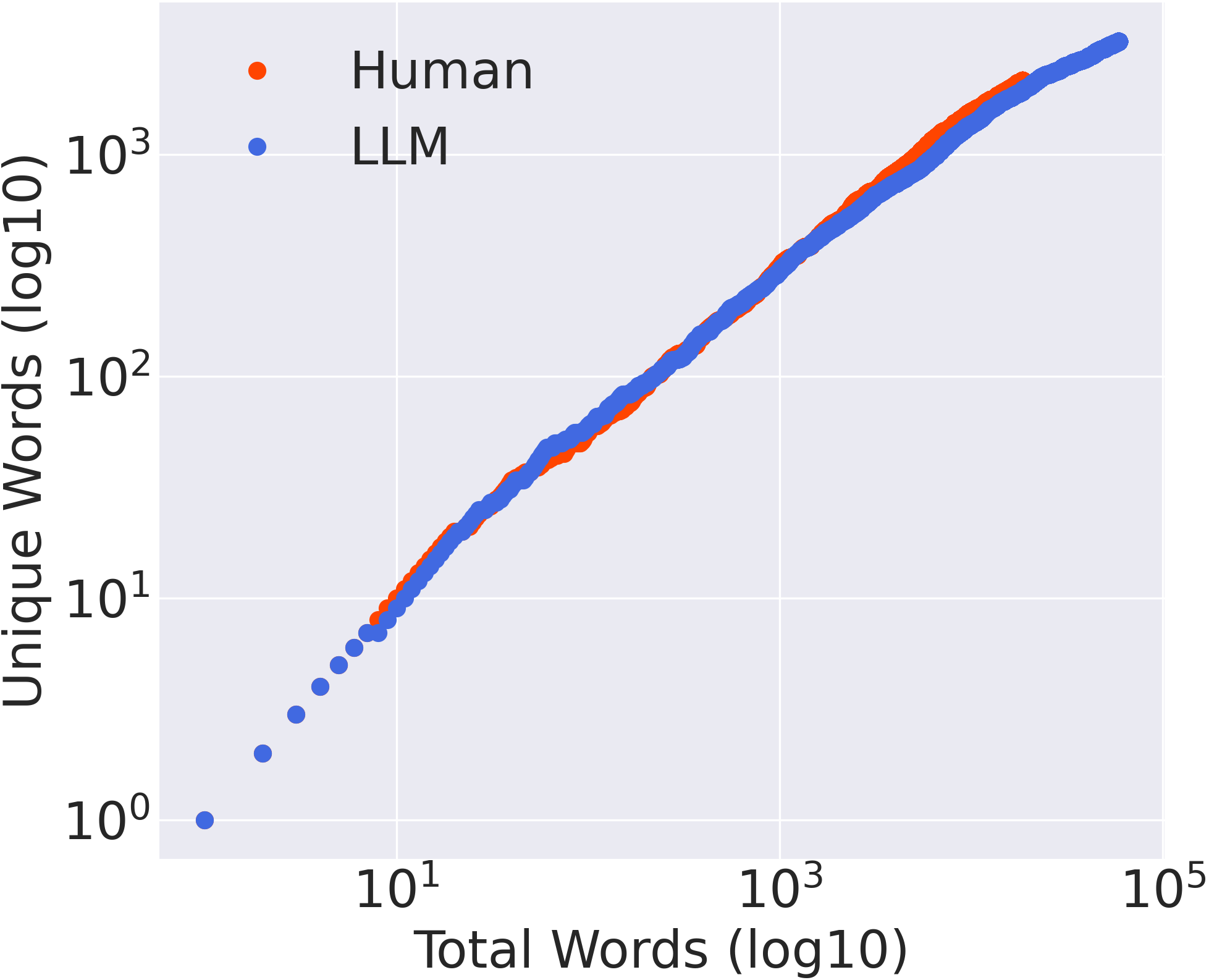}
        \caption{Heap's Law for Paper Abstract}
    \end{subfigure}
    
    \caption{Comparison of Zipf's Law and Heap's Law 
    between Korean text written by humans and those generated by LLMs.}\label{fig:appendix_zipf_heap}
\end{figure}

We use Zipf's Law~\citep{zipf2016human} and Heap's Law~\citep{heaps1978information} to compare and analyze the 
lexical diversity between human-written text and text generated by 
LLMs. 
Zipf's Law is an empirical rule 
that describes the frequency distribution of words in 
natural language texts. 
Zipf's Law illustrates the relationship between word frequency and rank,
showing that frequently used words appear much more 
often than less common words.
Heap's Law describes the relationship between the size of a 
text and the number of unique words~(vocabulary size) it contains. 
Heap's Law examines how the number of unique words increases 
with the total word count, 
indicating how often new words are introduced as the text lengthens.
Figure~\ref{fig:appendix_zipf_heap} 
shows Zipf's Law and Heap's Law for 
Korean text written by humans and LLMs. 

First, as seen in Zipf's Law, text written by humans show a 
steep decline in the frequency of commonly used words, 
indicating that a wide variety of words are used. 
In contrast, text written by LLMs show a 
higher concentration of frequently used words, 
with noticeable repetition of certain vocabulary. 
This reflects the tendency of LLMs to repeatedly 
use words that frequently appear within learned patterns.
Next, as observed in Heap's Law, 
text written by humans demonstrate a consistent increase in 
the number of unique words as the total word count rises, 
showcasing lexical diversity. 
On the other hand, text written by LLMs show a slower 
increase in unique words compared to humans, 
indicating relatively lower lexical diversity. 
This suggests that LLMs tend to use vocabulary repetitively. 
In conclusion, LLMs exhibit lower lexical diversity 
when writing texts compared to humans.

\begin{mdframed}
    \noindent\textbf{Finding.}
    LLMs tend to use a narrower range of vocabulary 
    and frequently repeat specific word patterns when writing, 
    unlike humans who typically show greater lexical diversity.
\end{mdframed}

\clearpage

\section{Expert Evaluation of Human-Written vs. LLM-Generated Korean Text}
\label{sec:human_evaluation} 

\subsection{Expert Evaluation Settings}

\paragraph{Data Preparation for Expert Evaluation}
We randomly select a total of 75 texts 
written by humans and four different LLMs.
First, we randomly select text written by humans.
For essays, we select two writings for each education level. 
For poetry, we select one poem each written by individuals 
under the age of 10 and over 40, and two poems each by individuals 
in their 10s and 20s \& 30s. 
For paper abstracts, we select three 
different papers. 
We select essays and research papers written by LLMs that are on the same topics 
as those written by humans, and for research abstracts, we select abstracts from the same papers.
Overall, we select a total of 30 essays, 30 poems, and 15 paper abstracts 
from humans and four LLMs.

\paragraph{Evaluator}
We request qualitative analysis for expert evaluation from 
three native Korean speakers specializing in Korean literature or 
Korean language education. 
Among them, one is a university student majoring in Korean literature, 
and the other two are current high school Korean language teachers.

\begin{figure*}
    \centering
    \begin{minipage}{\textwidth}
        \centering
        \includegraphics[width=0.85\textwidth]{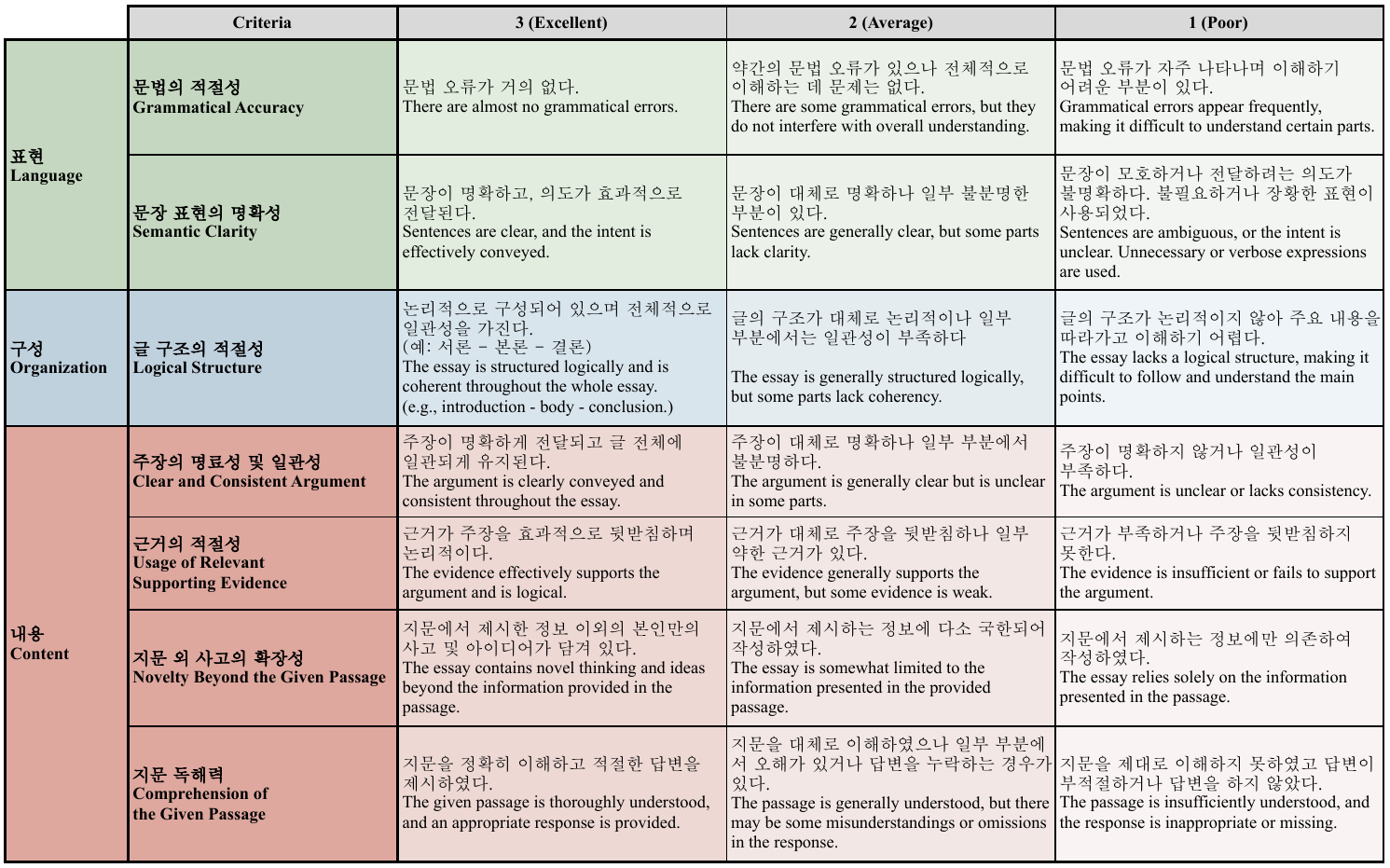}
        \subcaption{Essay}
    \end{minipage}
    \begin{minipage}{\textwidth}
        \centering
        \includegraphics[width=0.85\textwidth]{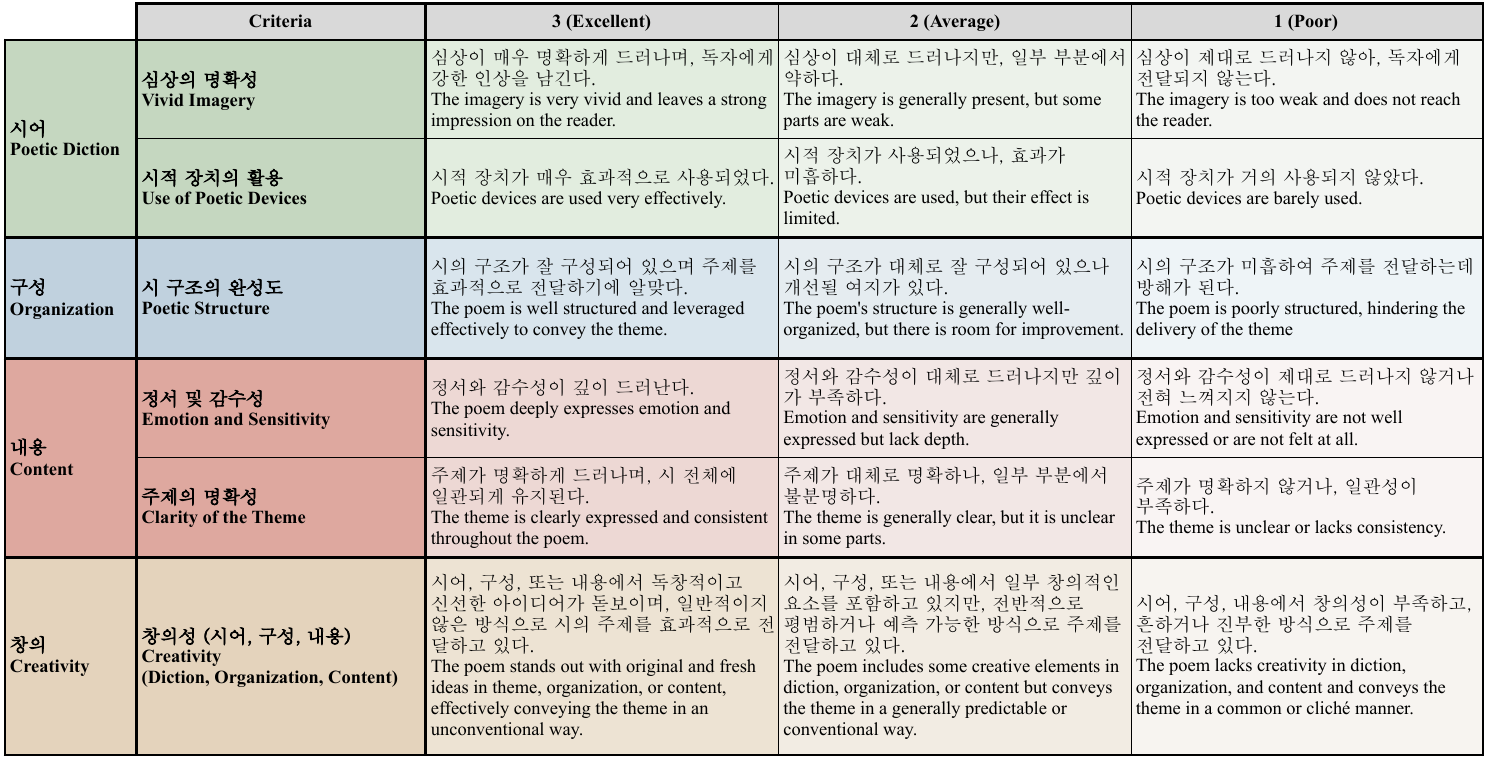}
        \subcaption{Poetry}
    \end{minipage}
    \begin{minipage}{\textwidth}
        \centering
        \includegraphics[width=0.85\textwidth]{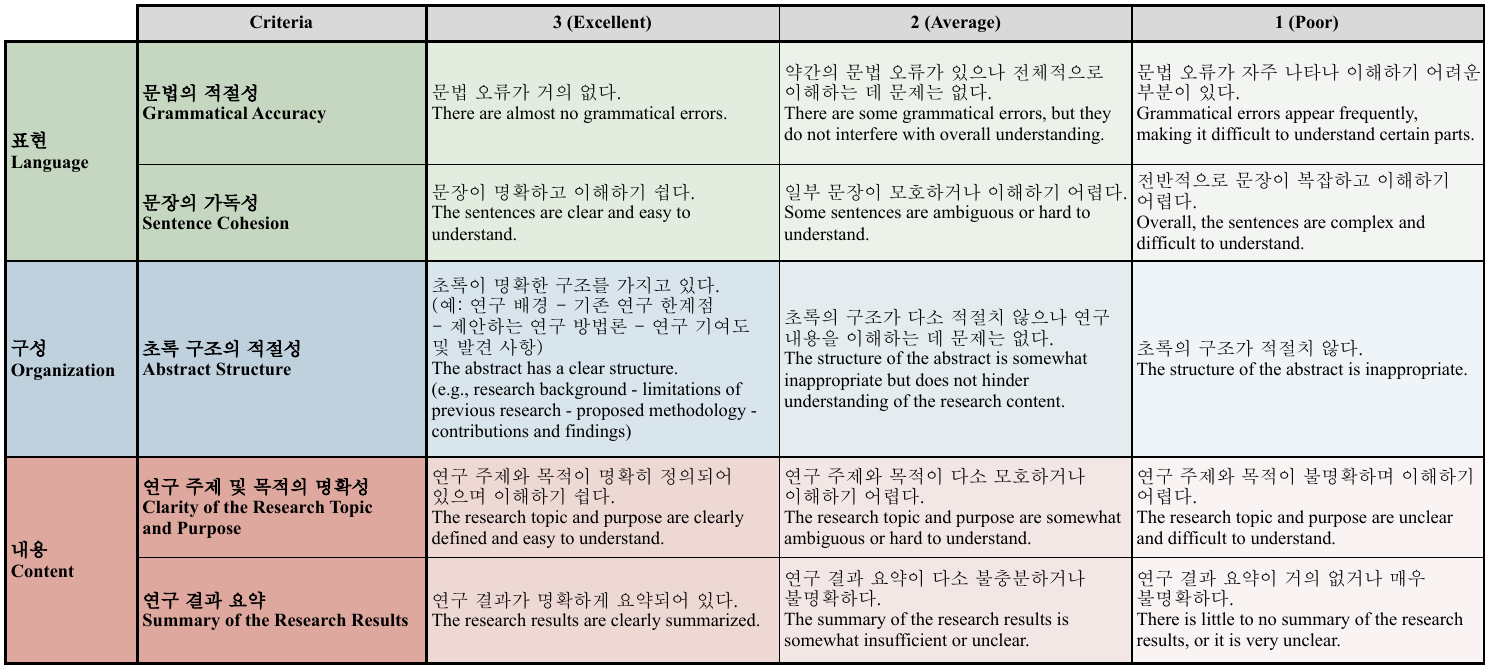}
        \subcaption{Paper Abstract}
    \end{minipage}
    \caption{Evaluation rubrics for expert evaluation.
    }\label{fig:rubric}
\end{figure*}

\paragraph{Evaluation Rubric}
We design an evaluation rubric to assess text written by humans 
and generated by LLMs from multiple perspectives. 
For essays and paper abstracts, 
we categorize the evaluation criteria into Language, 
Organization, and Content, creating specific subcategories for each. 
In poetry, we evaluate based on Poetic Diction, Organization, Content, 
and Creativity, 
also detailing subcategories for these main categories. 
Each detailed evaluation item is rated on a 3-point scale to 
standardize the assessment process across different text types. 
This structured approach allows us to systematically compare 
the qualitative aspects of human and LLM-written Korean text.
We have our rubric reviewed by an independent expert, 
a high school Korean language teacher specializing in Korean language education, 
alongside the evaluators involved in the expert evaluation.
The evaluation rubrics we designed can be found 
in Figure~\ref{fig:rubric}. 

\paragraph{Essay}
\begin{itemize}
    \item \textbf{Language} focuses on grammatical accuracy and semantic clarity, emphasizing the importance of clear communication in essays, which is crucial for conveying arguments effectively. 
    \item \textbf{Organization} evaluates the logical structure, which is essential for maintaining a coherent flow of ideas and ensuring the reader can follow the argument easily.
    \item \textbf{Content} addresses the purpose of essays, which is to present arguments. Criteria like argument clarity, use of evidence, comprehension, and extension of ideas beyond the given passage are central because essays are often judged on their analytical depth and ability to engage with the topic.
\end{itemize}

\paragraph{Poetry}
\begin{itemize}
    \item \textbf{Poetic Diction} examines imagery and poetic devices, which are fundamental in poetry for evoking emotions and creating depth.
    \item \textbf{Organization} looks at the completeness of the poetic structure, acknowledging that form and structure are as vital as content in poetry.
    \item \textbf{Content} emphasizes emotion, sensitivity, 
    and thematic clarity, highlighting the role of poetry in 
    conveying complex emotions and abstract themes.
    \item \textbf{Creativity} evaluates the originality of content, 
    diction, and organization, recognizing the reliance of poetry on 
    creative expression.
\end{itemize}

\paragraph{Paper Abstract}
\begin{itemize}
    \item \textbf{Language} focuses on grammatical accuracy and sentence cohesion.
    \item \textbf{Organization} evaluates the abstract structure, ensuring that it follows the conventional format that guides readers through the research purpose, methods, and findings.
    \item \textbf{Content} targets the clarity of the research topic, purpose, and results, ensuring that abstracts succinctly summarize the essential components of the study.
\end{itemize}

\clearpage

\begin{figure*}[h!]
    \centering
        \includegraphics[width=13cm]{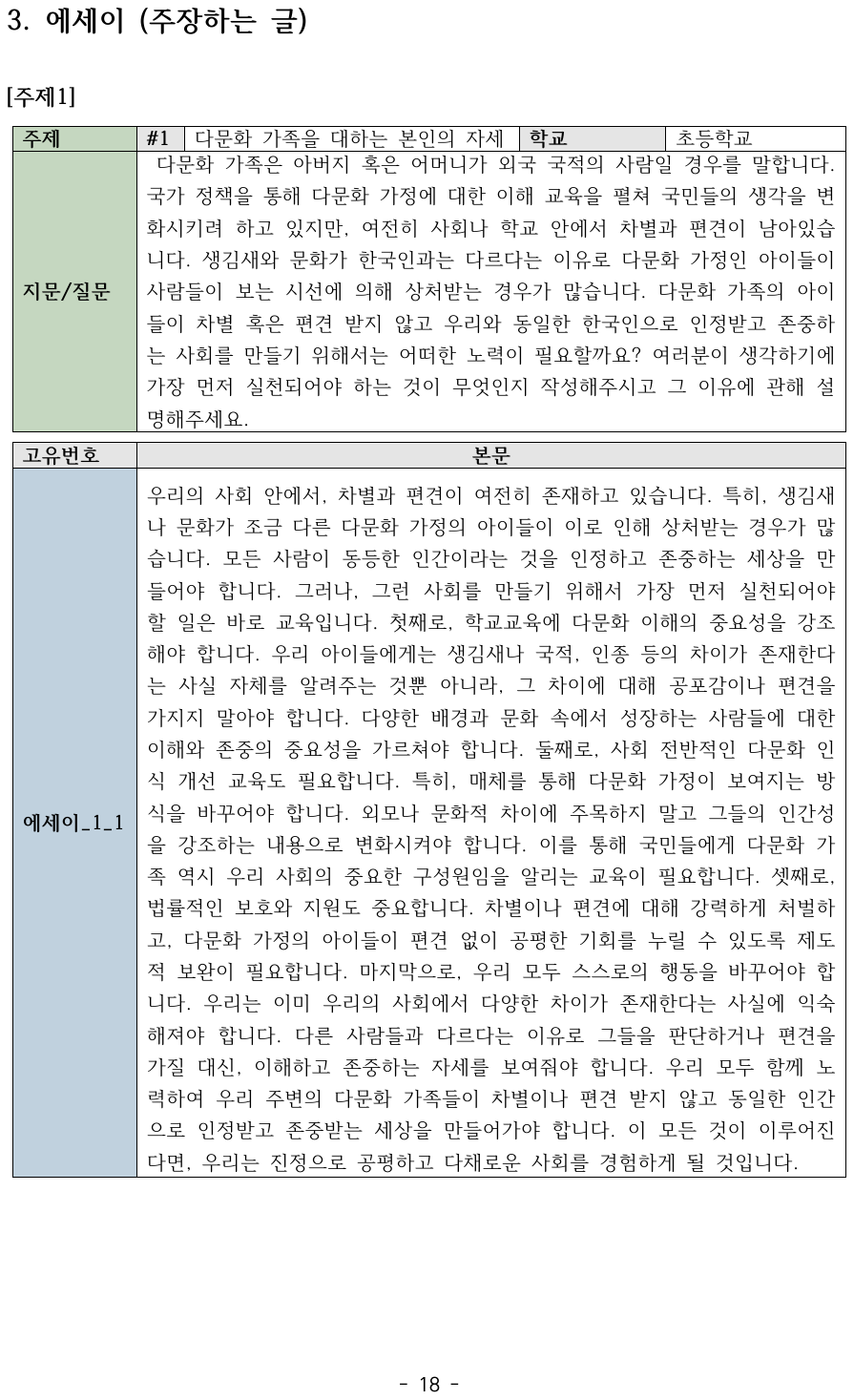}
    \caption{Example essay for expert evaluation.}\label{fig:guidelines_kor}
\end{figure*}

\begin{figure*}[h!]
    \centering
        \includegraphics[width=13cm]{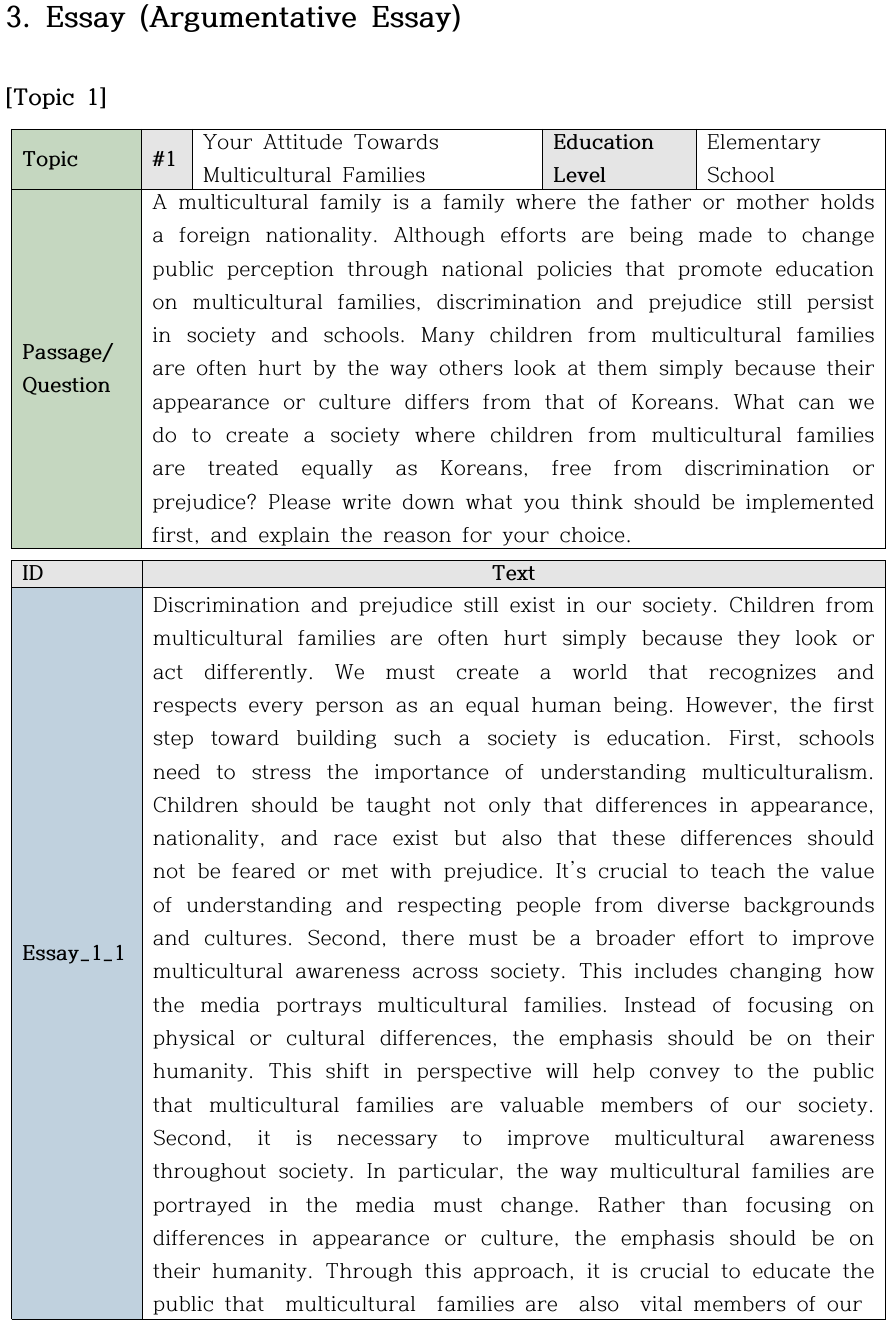}
    \caption{Example essay for expert evaluation~(English translation).}\label{fig:guidelines_eng}
\end{figure*}

\paragraph{Guidelines for Evaluation}
We provide our evaluation rubric along with the texts 
to evaluators with the following guidelines: 
1) Essay: We provide the topic of the essay, 
the education level of the author, 
and the essay prompt along with the text to be evaluated.
2) Poetry: For poetry, 
we include the age group of the author along with the 
text being evaluated.
3) Paper Abstract: We provide the title of the 
research paper and the full content of the paper 
along with the abstract to be evaluated.
These guidelines ensure that evaluators have all necessary context 
to accurately assess the texts across different categories and criteria.
Figures~\ref{fig:guidelines_kor} and~\ref{fig:guidelines_eng} 
respectively show an example essay for expert evaluation, 
and its English translation.
We ensure that evaluators do not make biased assessments 
by not providing information on whether each evaluated text 
is written by a human or generated by LLMs.

\clearpage

\subsection{Expert Evaluation Results}
\begin{figure}[hbt!]
    \centering
        \includegraphics[width=\textwidth]{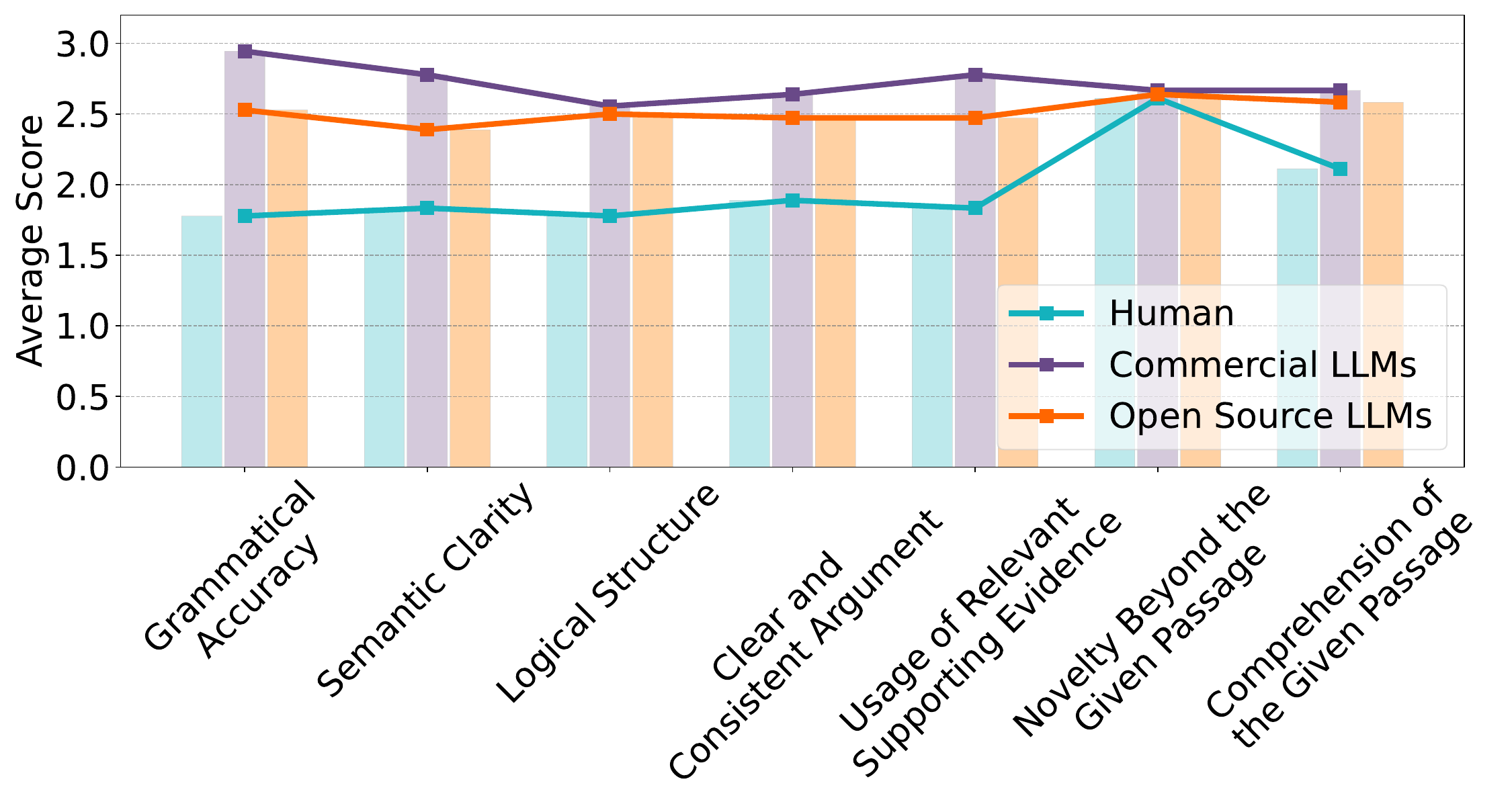}
    \caption{Results of expert evaluations for essays.}\label{fig:human_evaluation_essay}
\end{figure}

Figure~\ref{fig:human_evaluation_essay} 
presents the expert evaluation results for essays written 
by humans and those generated by LLMs. 
We compare the results for text written by humans, 
commercial LLMs~(GPT-4o and Solar), 
and open-source LLMs~(Qwen2 and Llama3.1). 
We report the average evaluation results of 
the three human evaluators.
Our analysis reveals that essays generated by LLMs 
receive better evaluations than those written by humans 
across seven metrics. 
Notably, essays by LLMs are distinctly rated higher in 
\textsc{Grammatical Accuracy} compared to those written by humans. 
This is likely because LLMs are trained on extensive data, 
enabling them to learn grammatical rules 
and thereby generate texts with fewer grammatical errors.
Overall, human-written essays score well in  
\textsc{Novelty Beyond the Given Passage} and 
\textsc{Comprehension of the Given Passage}.
One reason is that humans tend to weave in creative elements 
that go beyond the given context. 
However, human-written essays receive lower ratings in these two metrics
compared to LLM-generated essays, 
which is an interesting result showing that LLMs not only 
properly understand and write about the given passage but 
also have the capability to utilize knowledge 
beyond the information provided in the passage.

The evaluators provide descriptive comments for each sample. 
The three main characteristics identified in LLM-generated essays are the excessive use of commas,
repetition, and linguistic maturity. 
LLMs overuse commas, especially after adverbial case markers and connective endings. 
LLMs frequently repeat expressions or present paraphrased versions of the same sentence. 
Furthermore, LLM-generated text often exceed the expected writing level for the given educational stage, 
exhibiting complex vocabulary, advanced logic, and intricate sentence structures.

\clearpage

\begin{figure*}[hbt!]
    \centering
        \includegraphics[width=\textwidth]{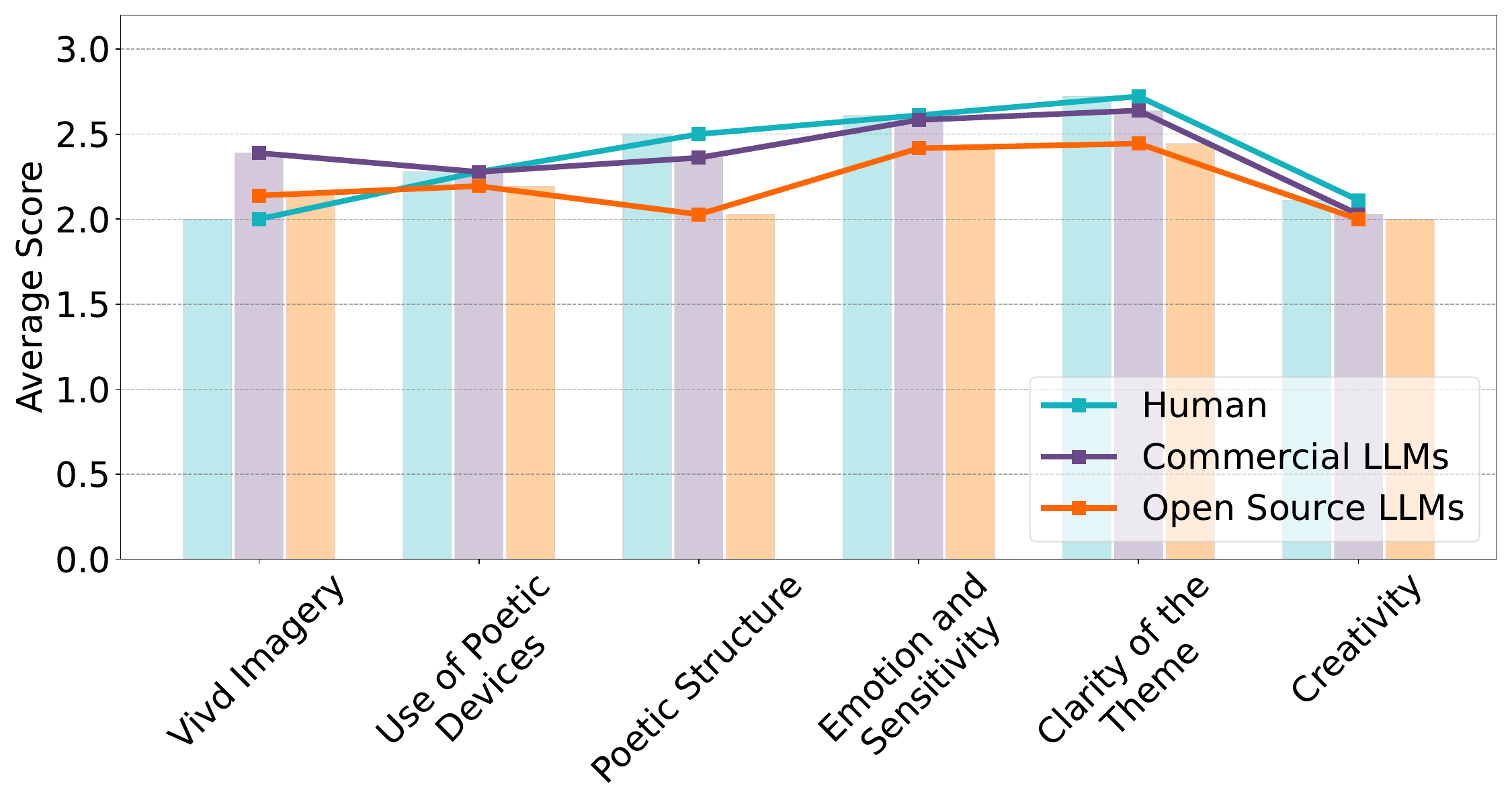}
    \caption{Results of expert evaluations for poetry.
    }\label{fig:human_evaluation_poetry}
\end{figure*}

\begin{figure*}[hbt!]
    \centering
        \includegraphics[width=\textwidth]{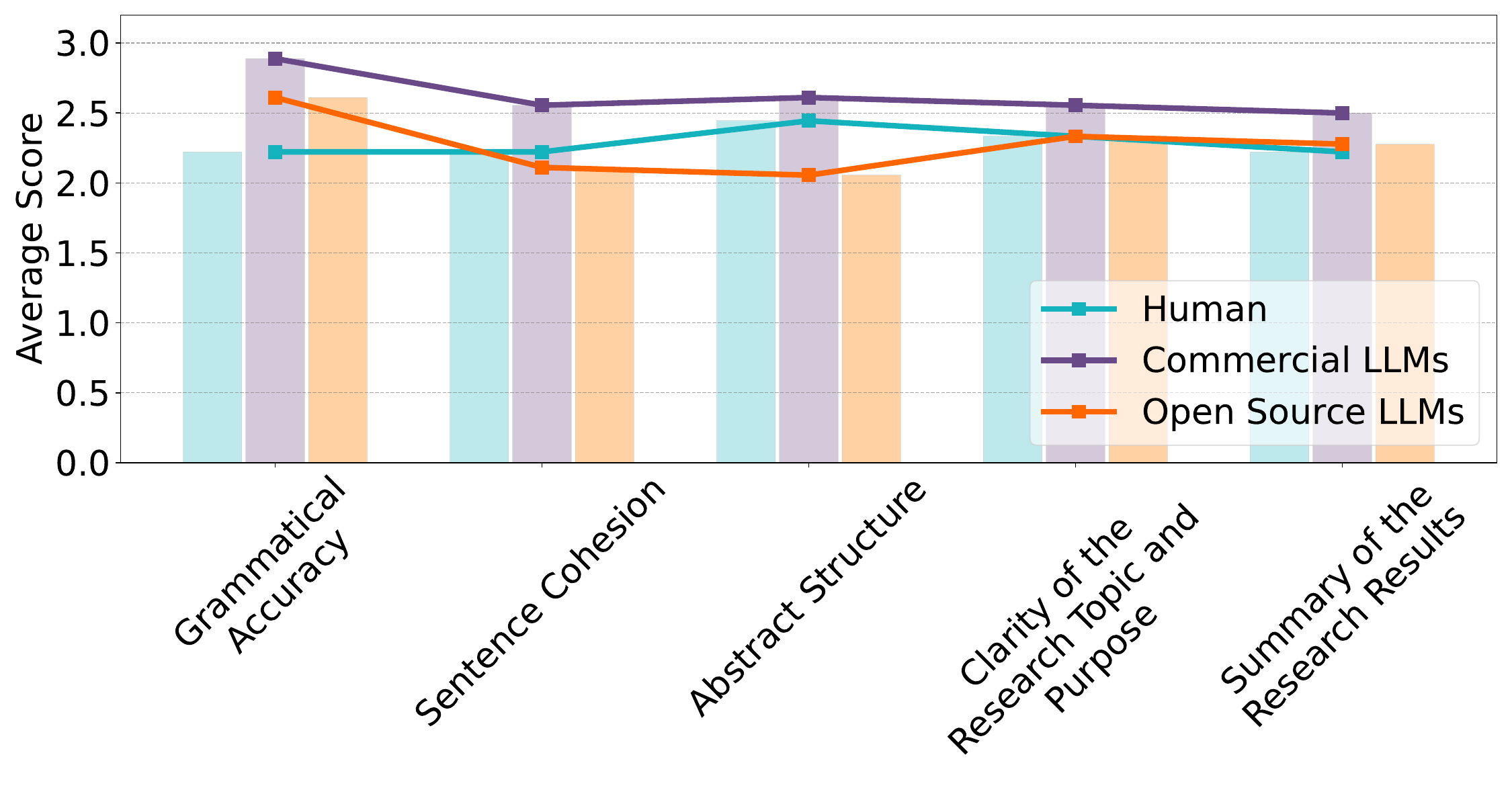}
    \caption{Results of expert evaluations for paper abstracts.
    }\label{fig:human_evaluation_paper}
\end{figure*}

Figures~\ref{fig:human_evaluation_poetry} and~\ref{fig:human_evaluation_paper} display 
the results of expert evaluations 
for poetry and paper abstracts, respectively.
Our analysis indicates that for poetry, 
commercial LLMs generally receive higher 
ratings than open-source LLMs, and the evaluations 
for poetry written by humans and commercial LLMs are quite similar. 
For paper abstracts, 
those generated by commercial LLMs receive the highest ratings, 
while abstracts written by humans and open-source LLMs 
are rated similarly. 
Overall, while LLM-generated essays consistently 
outperform human-written ones across all metrics, 
the differences in evaluations for poetry and paper abstracts are 
less pronounced. 

\end{document}